\pdfoutput=1

\documentclass[11pt]{article}

\usepackage[preprint]{acl}

\usepackage{times}
\usepackage{latexsym}

\usepackage[T1]{fontenc}

\usepackage[utf8]{inputenc}

\usepackage{microtype}

\usepackage{inconsolata}

\usepackage{graphicx}
\usepackage{booktabs}
\usepackage{colortbl}
\newcommand\sect[1]{\S\ref{#1}}
\usepackage{comment}
\usepackage{amsmath} %
\usepackage{amssymb}
\usepackage{graphicx}
\usepackage{booktabs}
\usepackage{tabularx}
\usepackage{array}
\usepackage{multirow}
\usepackage{dashrule}
\usepackage{xcolor}
\usepackage{soul}
\usepackage{enumitem}

\newlist{compactitem}{enumerate}{3} %
\setlist[compactitem,1]{label=\textbullet, nosep, leftmargin=*}
\newlist{compactitemwide}{enumerate}{3}
\setlist[compactitemwide,1]{label=\textbullet, nosep, leftmargin=2em}

\providecommand{\sectionvspace}{\vspace{-0.05cm}}

\usepackage{diagbox} %
\title{What Has Been Lost with Synthetic Evaluation?}

\author{Alexander Gill$^\dagger$ \quad Abhilasha Ravichander$^\ddag$  \quad Ana Marasovi\'c$^\dagger$ \\
$^\dagger$University of Utah\\
$^\ddag$University of Washington\\
\texttt{\small alex.gill@utah.edu \quad aravicha@cs.washington.edu \quad ana.marasovic@utah.edu}
}

\usepackage{graphicx}
\usepackage{subcaption}
\usepackage{caption}

\usepackage{listings}
\begin{document}
\maketitle

\begin{abstract}

Large language models (LLMs) are increasingly used for data generation. 
However, creating evaluation benchmarks raises the bar for this emerging paradigm. 
Benchmarks must target specific capabilities, penalize exploiting shortcuts, and be challenging. Through two case studies, we investigate whether LLMs can meet these demands by generating reasoning-over-text benchmarks and comparing them to those created through careful crowdsourcing. 
Specifically, we evaluate both the \emph{validity} and \emph{difficulty} of LLM-generated versions of two high-quality reading comprehension datasets: CondaQA, which evaluates reasoning about negation, and DROP, which targets reasoning about quantities. 
We find that prompting LLMs can produce variants of these datasets that are often valid according to the annotation guidelines, at a fraction of the cost of the original crowdsourcing effort. 
However, we show that they are \emph{less challenging for LLMs} than their human-authored counterparts, do not preserve the same model rankings, and that these differences may be imperceptible to researchers inspecting the data.
This finding sheds light on what may have been lost by generating evaluation data with LLMs, and calls for critically reassessing the immediate use of this increasingly prevalent approach to benchmark creation.

\end{abstract}

\section{Introduction}
\label{sec:intro}

\begin{table}[t]
\centering
\resizebox{\columnwidth}{!}{%
\begin{tabular}{lrrr}
\toprule
\textbf{CondaQA} & \textbf{Human$^\dagger$} & \textbf{\texttt{o3-mini}}$^\triangle$ & \textbf{\texttt{Llama-3.3}}$^\circ$ \\
\midrule
Original-Data Accuracy & 91.9  & 72.4  & 78.8  \\
\arrayrulecolor{black!20}\midrule
Full Bundle Consistency & 81.6 & 45.9  & 48.5  \\
Paraphrase Consistency & 93.6  & 69.6 & 76.5  \\
Scope Consistency & 86.5 & 55.5  & 62.8  \\
Affirmative Consistency & 88.2 & 56.9 & 67.3  \\
\arrayrulecolor{black}\toprule
\textbf{DROP} & \textbf{Human$^\ddagger$} & \textbf{o3-mini}$^\triangle$ & \textbf{\texttt{Llama-3.3}}$^\circ$ \\
\midrule
Original-Data Token F1 & 96.4 & 84.3  & 70.9  \\
\arrayrulecolor{black!20}\midrule
Consistency & N/A  &  53.5  & 35.5  \\
\arrayrulecolor{black}\bottomrule
\end{tabular}%
}
\caption{Performance on the CondaQA dev set and the DROP contrastive test set. The original accuracy is calculated on the data used to create contrastive instances. Consistency reflects the rate at which all minimally different examples within a bundle are answered correctly. \emph{These results show that these reasoning benchmarks remain challenging. %
We study whether comparable variants can be created using LLMs.} $^\dagger$Cf.\  \citet{ravichander-etal-2022-condaqa}. $^\ddagger$Cf.\ \citet{dua-etal-2019-drop}.
$^\triangle$\texttt{o3-mini-2025-01-31} 
$^\circ$\texttt{meta-llama/Llama-3.3-70B-Instruct} %
}
\label{tab:chat-benchmark-results}
\end{table}

The development of benchmarks that test frontier models in reasoning over text has been at the heart of NLP research~\cite{DBLP:journals/csur/RogersGA23}. %
Such benchmarks aim to cover a broad range of input variations and related skills, and prevent models from performing well by exploiting data shortcuts \cite{bowman-dahl-2021-will}. 
They are created to be challenging, so that even the most capable emerging models cannot solve them immediately and they remain a meaningful yardstick for measuring progress. 
Yet these goals are often hindered by crowdsourcing annotation practices that favor speed to maximize hourly pay, such as crowdworkers reusing the same pattern(s) to create dataset examples. %
An alte\-rna\-ti\-ve approach, prompting LLMs, does not suffer from the same misaligned incentives. 
LLMs are capable of creating content they will struggle to reason about \cite[\emph{the generative AI paradox}]{west2024the}.
However, a key challenge lies in getting LLMs to follow annotation instructions. 
The silver lining is that it is possible to review annotations and revise guidelines without delay because LLMs produce annotations instantly. 
Moreover, since their annotations come at little or no cost,  one can iterate as many times as they are willing to review-and-revise. 
 
This makes it tempting to substitute LLMs for human annotators, as is becoming increasingly common (\sect{sec:related_work}). 
However, before adopting this approach to creating new challenging  benchmarks, we must ask: 
\emph{When created under the same annotation criteria, how do LLM-generated benchmarks compare to their challenging human-authored counterparts?}
If the data is not comparable in quality or as effective for assessing model limitations, this should give us pause in using LLMs as annotators in this context. %

We address this question through the lens of reading comprehension, which is a common format in NLP for probing reasoning over text capabilities of frontier models \cite{DBLP:journals/csur/RogersGA23}. 
Specifically, we study compositional reasoning over numbers as in DROP \cite{dua-etal-2019-drop,gardner-etal-2020-evaluating} and reasoning about the implications of negated statements as in CondaQA \cite{ravichander-etal-2022-condaqa}. %
These reasoning types continue to challenge LLMs, as shown in Table \ref{tab:chat-benchmark-results}. 

What would LLMs-as-annotators need to be able to generate to create datasets like these?  
First, questions that are complex and require deeper understanding or inference, rather than being answerable by a phrase matching the paragraph's wording. %
Second, minimal edits to the passages or questions to test that the model has the skills related to a specific type of reasoning.  
For example, an edit that only changes what is negated in a passage enables testing whether the model has the skill of understanding negation scope. %
In turn, this skill enables reasoning about \emph{both} the original and edited passage, and answering questions about the negated content in the context of \emph{both} passages correctly. 

We extensively prompt an LLM to generate complex questions and contrastive edits, as would likely be done by dataset creators opting to use LLMs as annotators.  
Prompt engineering we conduct (\sect{sec:generations}) illustrates that using LLMs as annotators does not eliminate the need for iterative instruction refinement and manual review, much like traditional human annotation. %
That is, although an LLM creates questions and edits faster and cheaper than people, the process of obtaining them may still require multiple rounds of manual review.

Our next findings are of practical importance. First, we manually analyze generations and find that LLM-generated data is moderately to highly valid, though rarely perfect (\sect{sec:generations}).
Second, in a user study with NLP researchers, we compare human and generated edits and find that \emph{generated edits are generally preferred to human-authored ones for better meeting the annotation guidelines} (\sect{sec:pref_study}). 
We attribute this result to their apparent simplicity, which allows them to adhere more closely to the guidelines. %
However, when benchmarking various models on human and generated versions, we find that human-authored edits still pose a greater challenge and rank models differently than generated versions, especially when bundle consistency is considered (\sect{sec:difficulty}). 

These findings suggest that \emph{the core challenge is not validity, but difficulty}. 
However, optimizing prompts for difficulty would require substantial human-in-the-loop effort because LLMs cannot yet reliably validate and answer their own generations. 
Moreover, this would effectively turn evaluation-data generation into a form of adversarial dataset construction whose ``naturalness'' has been questioned \cite{bowman-dahl-2021-will}.  
Therefore, how to generate challenging instances without relying on excessive human effort and producing artificially-sounding text remains an open question. 
In light of this, using LLMs as annotators for any new benchmark does not seem advisable as of yet, especially if the validity of the generated data is the only measure of dataset quality.\footnote{Our code is available at \href{https://github.com/utahnlp/eval-synth-eval}{Github}.}

\section{Background}

\paragraph{Reading Comprehension to Probe Reasoning.}
SQuAD \cite{rajpurkar-etal-2016-squad} popularized using reading comprehension to probe models. %
\citet{weissenborn-etal-2017-making} show that SQuAD questions are often answerable with a span in the paragraph that matches the expected answer type and is lexically related to the key words in the question. %
This has led to the other datasets posing greater reasoning challenges \cite[among others]{yang-etal-2018-hotpotqa,kocisky-etal-2018-narrativeqa,DBLP:conf/iclr/YuJDF20}. %
We focus on two such datasets: DROP \cite{dua-etal-2019-drop} and CondaQA \cite{ravichander-etal-2022-condaqa}, which still challenge current models (Table \ref{tab:chat-benchmark-results}). DROP is widely used, as evident by its citation count, and CondaQA exemplifies the kind of high-quality, contrastive, and reasoning-focused dataset we aim to test LLMs' ability to generate.

\paragraph{DROP and CondaQA Questions.}

See Table \ref{tab:examples_drop_condaqa} in the Appendix for examples of human-authored questions in both datasets. 
Questions in these datasets are posed over Wikipedia passages.
In DROP, questions %
must ``require complex linguistic understanding and discrete reasoning'' \cite{dua-etal-2019-drop}. An extra condition is that the BiDAF model~\cite{seo2016bidirectional} cannot answer a question. We do not adversarially generate, as this may raise other issues \cite{bowman-dahl-2021-will}.  
In CondaQA, questions should be ``targeted towards the implications of the negated statement'' \cite{ravichander-etal-2022-condaqa}, i.e, they should be about (1) the negated statement, rather than other information in the passage, and (2) about an implication of the negated statement.

\paragraph{Contrast Sets.} Models can solve benchmarks by exploiting data shortcuts \cite{gururangan-etal-2018-annotation}. %
Thus, they are better evaluated based on \emph{consistency}\,---\,their ability to solve an entire set of slightly different examples with different ground truth, known as \emph{contrast sets} \cite{gardner-etal-2020-evaluating} or \emph{counterfactual data} \cite{DBLP:conf/iclr/KaushikHL20}. %
Contrastive edits are valid if minimal changes give a different ground truth, while maintaining fluency. %
However, edits \emph{targeting specific changes} help evaluate whether models have the necessary skills for a specific type of reasoning. 
For example, an edit that requires changing only what is negated tests models' understanding of the negation scope:

{\small
\begin{compactitem}
    \item ``Nearly all of his possessions were destroyed \emph{with the exception} of a guitar and an automobile''
    \item \underline{\emph{Edit}}: ``Nearly all of his possessions were destroyed (including a guitar), \emph{with the exception} of an automobile''
\end{compactitem}
}
\normalsize
Only with this skill can models consistently reason about negation and answer questions as ``Was Parsons able to use his guitar after the fire?'' in both contexts. %
We introduce other edit types below. %

\paragraph{CondaQA Edits.} The example above is the CondaQA \emph{scope edit}, which tests robustness to changing the part of the text that the negation cue affects. 
The other edit types are \emph{paraphrase edits},\footnote{While not contrastive to the original, paraphrase edits combined with other edit types, like scope and affirmative edits, form sets where some variants yield different answers.} 
which test a model's robustness to different ways of expressing negation, and \emph{affirmative edits}, which reverse negation to assess whether models can process negation and whether they rely on subtle artifacts when they do.\footnote{Success on negated text but failure on its affirmative version suggests artifacts were used. Failure on negated, success on affirmative text shows an inability to handle negation.}
CondaQA was designed to be contrastive, with all edit types systematically annotated. See Table \ref{tab:condqa_edits} (Appendix) for human-authored edits. 
We aim to generate all of these edit types for CondaQA dev passages. 

\paragraph{DROP Edits.} \citet{gardner-etal-2020-evaluating} report three DROP edit types. 
\emph{Compositional edits} add a reasoning step to questions (Table \ref{tab:examples_drop_condaqa}; Appendix). %
\emph{Semantic edits} invert question meanings, such as changing ``shortest'' to ``longest''. %
\emph{Temporal edits} modify event order by changing their order in the questions or the passage (Table \ref{tab:temporal_example}; Appendix), or by changing the dates associated with these events in the passage. %
Unfortunately, the DROP contrast set does not explicitly mark the edit type.\footnote{\url{https://github.com/allenai/contrast-sets/blob/main/DROP/drop_contrast_sets_test.json}} 
Thus, one author of this paper has manually determined compositional and temporal edits that change event order in passages, but could not reliably detect semantic edits.\footnote{We choose \emph{passage-level} temporal edits because they, with compositional edits, cover at least one type of both question-level and passage-level DROP edits.} %
We generate compositional edits using pairs of passages and the generated questions. %
Because the contrast set contained too few temporal edits, we generate temporal edits using the original DROP dev passages that include dates.

\section{Related Work}
\label{sec:related_work}

\begin{figure*}[t]
  \centering
  \includegraphics[width=0.9\textwidth, trim=0cm 1cm 0cm 0cm, clip]{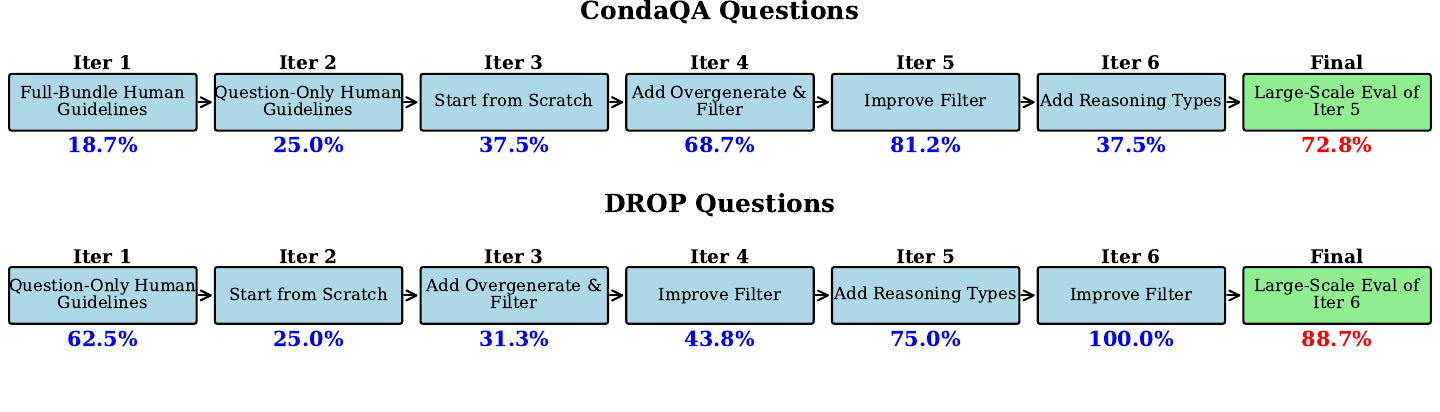} %
  \caption{The percentage of \textbf{valid questions} in a sample of 16 for intermediate evaluations, and 100 or 114 for final evaluations. 
Questions are generated with \texttt{gpt-4-turbo-2024-04-09} and assessed by one author of this paper. The full prompt code and templates for each iteration can be found in our \href{https://github.com/utahnlp/eval-synth-eval}{Github} repository.}
  \label{fig:question_prompting}
  \vspace{-0.5cm}
\end{figure*}

\paragraph{Question Generation (QG).} This is a long-standing NLP task, with the first shared task in 2010 \cite{rus-etal-2010-first} and many approaches proposed since \cite{DBLP:journals/corr/abs-1905-08949, guo2024surveyneuralquestiongeneration}. %
QG still largely focuses on generating fluent, factual questions that closely mirror the context's wording \cite{ushio-etal-2022-generative, samuel-etal-2024-llms}.  
\citet{yehudai-et-al-2024-genie} target long-form questions, but still require little composition or inference. 
Multi-hop QG, which links explicit facts across passages \cite{kulshreshtha-rumshisky-2023-reasoning}, is more relevant but differs from reasoning over \emph{implicit} meaning within a single context as in CondaQA and DROP. Other related QG work focuses on text-to-SQL \cite{wu-etal-2021-data}, pedagogical question generation \cite{pmlr-v264-tonga25a}, and paraphrasing existing mathematical questions \cite{DBLP:conf/iclr/YuJSYLZKLWL24}, none of which involve generating reasoning-intensive questions from scratch. \citet{DBLP:journals/corr/abs-2410-15512} introduce RQA, where a question is generated from a given answer, showing that LLMs struggle particularly with numerical cases.

\paragraph{Edit Generation.} A line of work has aimed to automate the creation of contrastive edits to improve their scale and diversity \cite[among others]{li-etal-2020-linguistically,ross-etal-2021-explaining,-2022-international, ross-etal-2022-tailor,dixit-etal-2022-core,chen-etal-2023-disco}; reviewed in \citet{DBLP:journals/access/StepinACP21, nguyen-etal-2024-ceval-benchmark, wang2024surveynaturallanguagecounterfactual}. %
While these approaches accept any minor change that alters the answer, we focus on editing that is done along specific dimensions such as changing what is negated or adding a set of calculations. 
 \citet{ross-etal-2022-tailor} also steer edits toward targeted attributes, but their approach requires dataset creators to adjust semantic representation-based control codes, whereas we achieve targeted edits through direct natural language instructions.

\paragraph{Paraphrase Generation.} This is another classic NLP task \cite{mckeown-1983-paraphrasing}. Our approach to generating CondaQA edits that change how negation is expressed, but keep the meaning intact, builds on work in prompting LLMs for paraphrasing \cite{cegin-etal-2023-chatgpt, witteveen-andrews-2019-paraphrasing, DBLP:journals/es/PehlivanogluGSSd24} and controllable paraphrasing \cite{hosking-lapata-2021-factorising, DBLP:conf/nips/KrishnaSKWI23}. 

\paragraph{Data Creation with LLMs.} Generating data with LLMs is increasingly common \cite{long-etal-2024-llms}, from NLP tasks such as classification \cite{moller-etal-2024-parrot, patwa-etal-2024-enhancing}, QA \cite{ye-etal-2022-zerogen,harsha-etal-2025-synthetic}, NLI \cite{liu-etal-2022-wanli,hosseini-etal-2024-synthetic}, to instruction tuning and alignment \cite{wang-etal-2023-self-instruct, DBLP:journals/corr/abs-2212-08073}.
Our work differs from this line of work in three key ways.
First, most prior efforts focus on generating \emph{training}, not \emph{evaluation}, data. Imperfect synthetic data may improve training \cite{amin2025escapingcollapsestrengthweak}, but is unacceptable for benchmarking.
Second, much of the existing work targets the easier task of generating only labels or answers \cite{maheshwari2024efficacysyntheticdatabenchmark}, whereas we generate content: questions and edits. This is more challenging to execute due to more complex prompt engineering and validity review, as well as costly answer annotation. 
Third, our goal is to critically and systematically analyze generated data. While some studies generate evaluation datasets or replicate existing crowdsourcing pipelines \cite{Gilardi_2023, qi2025large, roemmele-gordon-2024-test, wu2023llmsworkershumancomputationalalgorithms}, they do not conduct such analysis. \citet{das2024surfacetrackingartifactualityllmgenerated} compares LLM-generated data to human-authored counterparts, but focuses on training data and the downstream effects of its use. 
Our results both corroborate and go beyond \citet{maheshwari2024efficacysyntheticdatabenchmark}'s. 
Specifically, we show that synthetic data is not always a good predictor of relative performance, and that bias does not explain why synthetic benchmarks are easier.

\sectionvspace
\section{Generating DROP and CondaQA}
\label{sec:generations}
\sectionvspace

In this section, we outline our prompting workflow for generating questions (\sect{sec:q_generation}) and edits (\sect{sec:edits_generation}). We also recap some of the prompting strategies we found most useful, and the data they were used to create in Table~\ref{tab:prompting-strategies} (Appendix).

\begin{figure*}[t]
  \centering
  \includegraphics[width=0.9\textwidth, trim=0cm 1.29cm 0cm 0cm, clip]{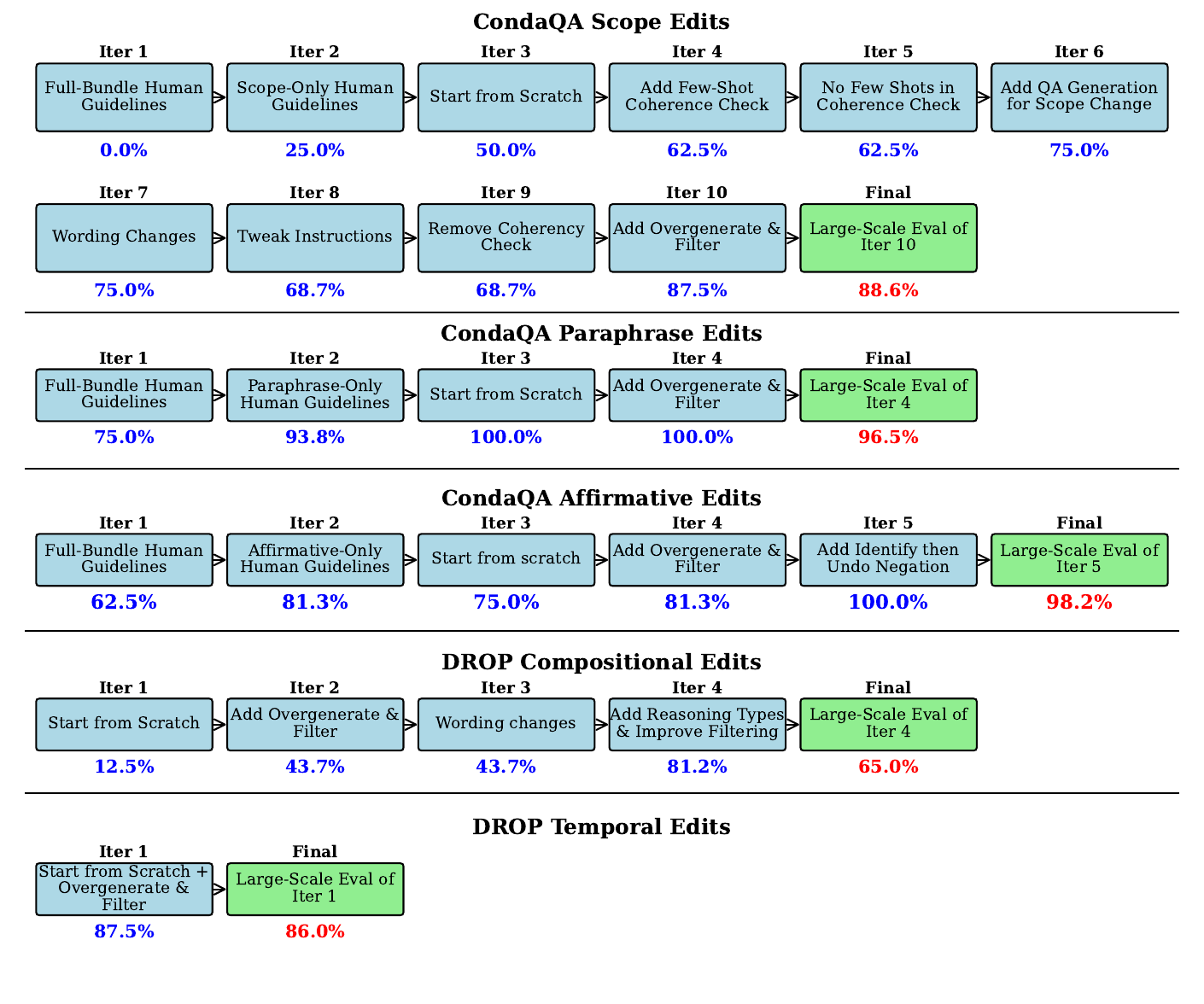}
  \caption{The percentage of \textbf{valid edits} in a sample of 16 for intermediate evaluations, and 100 or 114 for final evaluations. 
Edits are generated with \texttt{gpt-4-turbo-2024-04-09} and assessed by one author of this paper. The full prompt code and templates for each iteration can be found in our \href{https://github.com/utahnlp/eval-synth-eval}{Github} repository.}
  \label{fig:editing_prompting}
\end{figure*}

We give a Wikipedia passage from the original dataset to \texttt{gpt-4-turbo-2024-04-09} (the state-of-the-art LLM at the time) and refine prompt instruction iteratively, with one of the authors manually validating 16 generations for each promising candidate prompt. %
We consider an annotation valid if it adheres to the original annotation guidelines; see validity criteria in Table \ref{tab:validity_criteria} in the Appendix. %
If validity is below 85\%, 
we adjust the prompt and experiment using a chat interface until we reach a promising version, which we then re-evaluate on the 16 examples.
Upon achieving either 100\% validity, or if further prompt engineering suggests no prospects of progress, the same author evaluated the most valid intermediate prompt on a larger sample of 100+ instances. %
Each iteration in Figures~\ref{fig:question_prompting}--\ref{fig:editing_prompting} represents our prompt after exploring several minor variations around a central prompting idea in that iteration. 
We also show these results in  Tables~\ref{tab:validity_questions_intermediate} and \ref{tab:edit_validity_with_intermediate}; Appendix. 
In total, we evaluate 421 paragraph-question pairs and 926 edits across both datasets in this section.

\subsection{Prompting for Question Generation}
\label{sec:q_generation}

We begin with prompts based on the instructions given to annotators.\footnote{We obtain CondaQA crowdsourcing templates via personal correspondence. %
The DROP templates are available at: \url{https://github.com/dDua/mturk-drop}.} 
For CondaQA, one such prompt obtains a question and paragraph edits sequentially in a single chat (Iter 1; CondaQA) and the other collects questions or a specific edit alone (Iter 2; CondaQA).  
For DROP, the latter is the only option (Iter 1; DROP).%
\footnote{The DROP contrast set authors shared that their instructions were loosely defined since they did the edits themselves.} 
As shown in Fig.~\ref{fig:question_prompting}, 
these initial prompts result in low validity. %
\emph{This shows that \texttt{gpt-4-turbo-2024-04-09} struggles to follow instructions designed for human crowdworkers.} %
Consequently, we switch to designing prompts from scratch.

For CondaQA, we prime the model to focus on useful information by guiding it through the following process: given a passage and a cue, it identifies what is negated, de-negates the cue, considers the implications of the negation, and then generates a question. 
The prompt requires that the generated question (1) targets the negation, (2) is about an implication of negation, not factual knowledge in the passage,  (3) has a valid answer type (yes, no, don't know, or a span in the passage), and (4) avoids restating the negation. %

We try to ensure that the negation is targeted by requesting questions whose answers change depending on whether the negation cue or its de-negated version is in the passage. 

For DROP, the initial prompts ask for questions that (i) require discrete reasoning over the passage and (ii) have a valid answer span (a span in the passage or question, date, or numbers).

The validity of these initial prompts is also low: only 37.5\% for CondaQA (Iter 3) and 25\% for DROP (Iter 2). %
\emph{We improve these results with the overgenerate-then-filter protocol} \cite{yehudai-et-al-2024-genie}.  
We generate 8 questions for each passage and select the first one that satisfies filtering. 
Each filter is done with another LLM call. %

For CondaQA, we first introduce filters for (1)--(3) above, but we found that including a filter for (4) was crucial (see Iter 4 vs.\ 5 in Fig.~\ref{fig:question_prompting}). 
After this iteration, we are unable to further improve the validity of generated CondaQA questions, and ultimately, their validity in a sample of 114 passage-question pairs is 72.8\%.

For DROP, we refine the filtering in a few iterations, each of which further improves the validity.  
The first filter checks that a question requires complex reasoning and looking over more than one part of the passage to answer it (Iter 4). 
The next filter checks that the question does not combine multiple questions into one and that it has a valid answer type. 
Before introducing the final filter, we enrich the prompt with an example of 9 types of discrete reasoning questions and request generating a question for each discrete reasoning type, then ranking them according to how well they adhere to the other guidelines (see Iter 5).\footnote{Types of discrete reasoning \cite{dua-etal-2019-drop}: subtraction, comparison, selection, addition, count and sort, coreference resolution, other arithmetic, set of spans, and other.} 
Given this improvement, we try a similar addition for CondaQA (Iter 6), but it drastically reduces validity.\footnote{Dimensions of commonsense reasoning \cite{ravichander-etal-2022-condaqa}: precondition, social norms, psychology, cause and effect, mutual exclusivity, synecdoche.} 
\emph{This demonstrates that a prompt engineering strategy crucial for producing one reasoning benchmark may not be universal.}
The final filter checks that the passage is self-contained in terms of facts required to answer the generated question (e.g., it does not require external knowledge about a sport or player stats).  %
With this final filter, we reach 100\% validity on the small sample which leads to 88.7\% valid generated DROP questions in a larger sample of 100 passage-question pairs. 
This performance drop from a small- to large-scale evaluation pinpoints a limitation of this protocol, which could also happen in traditional human evaluation. 

\subsection{Prompting for Editing}
\label{sec:edits_generation}
Unlike for question generation, initial prompts for certain edits, such as paraphrasing and affirmative ones, immediately give moderate to high validity (CondaQA Iter 3, Fig.~\ref{fig:editing_prompting}). 
However, multiple iterations were required to refine the prompt for complex edits like changing the negation scope. 
In that case, we discover it is effective to have the model generate a question whose answer is the entity or event that is negated, and then edit the passage so the answer to the question changes  (Iter 6; Scope):

\begin{lstlisting}[basicstyle=\ttfamily\scriptsize, breaklines=true, frame=single]
System: Given the sentence and the negation cue, identify what is being negated.
User: Sentence: {sentence}\n\nNegation cue: {cue}
Assistant: {negation_subject}
User: Please generate a question that is answered by the subject of what is negated by \"{cue}\" in the original sentence. Just output the question and the answer in the form of \"Q: Generated question \nA: Answer to generated question\"
Assistant: Q: {generated_question} \n A: {generated_answer}
User: Now edit the original sentence so that the answer to the question is different, but make sure to keep the negation cue \"{cue}\" somewhere in the sentence.
Assistant: {edited_sentence}
User: Original Passage: {passage}\nNow rewrite the original passage with the rewritten sentence. Please make any edits necessary to make sure the whole passage is coherent. Please return just the edited passage.
Assistant: {edited_passage}
...
\end{lstlisting}

In most cases, the model needs to be primed to focus on relevant information and guided through a multi-step generation process. %
For instance, in affirmative edits, the initial prompt (Iter 3) involves removing the negation cue and then polishing the text. %
A more effective alternative (Iter 5) is to explicitly ask what is being negated, ask the model to de-negate the negation (e.g., change ``rarely'' to ``often'') 
while keeping the original negated content in focus, and then integrate the revised content into the passage. 
We often introduce overgenerating-and-filtering with other refinements, which makes its role in editing less immediately apparent. %
However, it is crucial (except for paraphrasing), as evident by its use in all final prompts.

Finally, while we achieve high validity for paraphrasing and affirmative edits, validity remains moderate for scope and temporal edits and poor for compositional edits.\footnote{To produce compositional edits, we do not edit the original questions because we simulate creating a dataset with complex questions and edits from scratch.} %
In those cases, as with question generation, human verification would be necessary to ensure clean contrast sets.

\section{Whose Annotations are Preferred?}
\label{sec:pref_study}

We next study whether NLP researchers prefer human or generated questions and edits based on how closely they follow the original guidelines for producing questions and edits with desired properties.
There is a concern about potential dataset contamination when doing such studies with synthetic data, as models may simply memorize and reproduce the human-authored dataset from their training data. To mitigate against the effects of straightforward reproduction, we analyze the generated data to eliminate the possibility that generations are paraphrased instances of their human-authored counterparts; see Appendix~\ref{sec:memorization}.

\paragraph{Setup.} We recruited 16 annotators who are NLP researchers and doctoral students to play the role of ``dataset creators''.\footnote{We obtained approval from our institution's IRB.}  
These annotators were selected to represent the typical demographic of researchers who develop evaluations for LLMs. 
We recruited them through social media platforms and personal networks and compensated them at a rate of \$16 per hour. 
In total, 63 paragraph-question pairs and 125 edits across two datasets were evaluated in this study, at a total annotation cost of \$256.  

Screenshots of this study are shown in Figures~\ref{fig:pref_study_instructions}--\ref{fig:question_pref_choices} for CondaQA and Figures~\ref{fig:drop_pref_1}--\ref{fig:drop_pref_9} for DROP (Appendix).%
\footnote{HTML templates for the study are available at our \href{https://github.com/utahnlp/eval-synth-eval}{Github}.}
Annotators were asked to imagine themselves as dataset creators who were designing a benchmark to probe LLMs' capabilities to reason and were provided background about the format of the datasets (without mentioning their names). 
They were then asked to choose edits and questions that better meet the dataset specification. 
They were also asked if the question or edit was invalid. 
For CondaQA, annotators are shown one question pair and three edit pairs corresponding to the three edit types. 
For DROP, annotators are shown one question pair and one compositional-edit pair.\footnote{We excluded temporal edits because we could identify only $\approx$20 human-authored ones.}
Since the compositional edits in \sect{sec:generations} were made on generated questions, we created compositional edits for human-authored questions here to enable a valid comparison. 
The order of human and generated annotations for both studies is randomized. 
Other details are in Appendix \sect{sec:pref_study_details}.

\begin{table}[t]
    \centering
    \resizebox{\columnwidth}{!}{%
    \begin{tabular}{llrr}
    \toprule
   & \multirow{2}{*}{\textbf{CondaQA}} & \% Valid & \% Valid \\
   & & (Annotators) & (Author) \\
    \midrule
     \multirow{4}{*}{\rotatebox{90}{\small\emph{Gen.}}}
   & Paraphrase & 77.4 & 90.3 \\
        & Scope & 64.5 & 87.1 \\
        & Affirmative & 83.9 & 93.5  \\
        & Questions &  64.5 & 77.4 \\
        \midrule
       \multirow{4}{*}{\rotatebox{90}{\small\emph{Orig.}}} & Paraphrase & 64.5 & - \\
        &Scope & 61.3 & - \\
       & Affirmative & 90.3& - \\
        & Questions & 58.1 & - \\
    \midrule 
       & \textbf{DROP} &  & \\ %
    \midrule
     \multirow{2}{*}{\rotatebox{90}{\small\emph{Gen.}}}
   & Compositional & 78.1 & 53.1 \\
        & Questions & 81.2    & 84.4\\
        \midrule
     \multirow{2}{*}{\rotatebox{90}{\small\emph{Orig.}}}
   & Compositional & 84.4 &  - \\ 
        & Questions & 78.1   & - \\
    \bottomrule
    \end{tabular}
    }
    \caption{The validity assessments for questions and edits from the human preferences study. Each row header states the dataset and edit type evaluated. The first column shows the percentage of total edits or questions evaluated which were marked as valid by annotators. All edits and questions evaluated by annotators were also evaluated by an author of this paper. Those results are shown in the second column.} %
    \label{tab:validity_from_pref}
\end{table}

\begin{table}[t]
    \centering
    \resizebox{\columnwidth}{!}{%
    \begin{tabular}{lrrrr}
    \toprule
        & \textbf{Win} & \textbf{Lose} & \multicolumn{2}{c}{\textbf{\% Tie}} \\
        &            &                 & As good & As bad \\
    \midrule
\textbf{CondaQA}\\
 \arrayrulecolor{black!20}\midrule
Paraphrase &  32.3  &  19.3 &  35.5  &  12.9 \\
(74.2\%)   &   39.1 &  13.0 &  43.5 &   4.3 \\
 \arrayrulecolor{black!20}\midrule
Scope &  38.7  &  35.5 &  12.9  &  12.9 \\
 (61.3\%) &  57.9  &  21.0 &  21.0  &  0.0 \\
 \arrayrulecolor{black!20}\midrule
Affirmative &  20.0  &  19.3 &  48.4  &  3.2 \\
 (77.4\%) &  37.5  &  4.2 &  54.2  &  4.2 \\
 \arrayrulecolor{black!20}\midrule
Questions &  35.5  &  32.3 &  16.1  &  16.1 \\
(48.4\%)  &  53.3  &  20.0 &  6.7  &  20.0 \\
 \arrayrulecolor{black}\midrule
\textbf{DROP}\\
  \arrayrulecolor{black!20}\midrule
Compositional & 18.8 & 34.4 &  43.8  & 3.1 \\
(53.1\%)   & 29.4 & 17.6 &  52.9  & 0.0 \\
 \arrayrulecolor{black!20}\midrule
Questions & 25.0& 18.8 &  53.1  & 3.1 \\
 (75.0\%) & 25.0 & 8.3 &  63.6  & 0.0 \\
 \arrayrulecolor{black}\bottomrule
    \end{tabular}
    }
    \caption{
    Human preference according to how well generations adhere to the definition of the CondaQA/DROP question/edit. Win/lose/tie refers to LLM-generated edits over human-authored. The percentages in the first column indicate the fraction of generations of that type for which the annotator and the author agree on validity. Every second row presents preferences considering only those unanimously valid generations where both the author and the annotator agree.
    }
    \label{tab:preferences_edits}
\end{table}

\paragraph{Results.} Tables~\ref{tab:validity_from_pref}--\ref{tab:preferences_edits} show the results. 
We observe that annotators assess both generated and human edits and questions as valid at a much lower rate than one author of this paper, with the exception of DROP compositional edits. 
They also rate the original CondaQA paraphrase/scope edits and questions, along with the DROP questions, as less valid than the generated ones. Moreover, annotators are often shown to prefer model-generated questions and edits to human-written questions and edits from the original dataset. We include example agreement and disagreement cases between annotators and one author of this paper for generated instances of CondaQA and DROP questions and edits in Tables~\ref{tab:compositional-agreements-drop}--\ref{tab:affirmative-agreements-condaqa} in the Appendix. %

The annotators' higher validity and preference for generations likely stems from the fact that the generated edits tend to be simpler and, due to the way the prompts were designed, follow the guidelines for each type of edit more closely. %
In contrast, original annotators often interpret the question and edit definitions in slightly different ways and sometimes introduce extra flair to the annotations. 
While these human annotations are, in our opinion, still valid, they do not always strictly follow the edit guidelines. %
This is aligned with observations from prior work \cite{ruggeri2024letguidelinesguideyou}. 
However, when juxtaposed with the generated edits and evaluated by NLP researchers based solely on the edit definitions, the generated annotations' stricter compliance with the guidelines highlights small flaws in the human edits that would not necessarily be evident if one only saw those edits in isolation.

\section{Which Benchmark is More Difficult?}
\label{sec:difficulty}
\begin{table*}[t]
\centering
\resizebox{\textwidth}{!}{%
\begin{tabular}{lrrrrrrrrrrrrrr}
\toprule
 & \multicolumn{2}{c}{\textbf{GPT-4-Turbo}$^\triangle$} 
 & \multicolumn{2}{c}{\textbf{GPT-4o}$^\circ$} 
 & \multicolumn{2}{c}{\textbf{o3-mini}$^\dagger$} 
 & \multicolumn{2}{c}{\textbf{Claude-Opus-4}$^\diamond$}
 & \multicolumn{2}{c}{\textbf{Gemini-2.5-Flash}$^\bigcirc$}
 & \multicolumn{2}{c}{\textbf{Llama-3.3-70b}$^\ddagger$ } 
 & \multicolumn{2}{c}{\textbf{Qwen2.5-72b}$^\star$ } \\
\cmidrule(lr){2-3}\cmidrule(lr){4-5}\cmidrule(lr){6-7}\cmidrule(lr){8-9}\cmidrule(lr){10-11}\cmidrule(lr){12-13}\cmidrule(lr){14-15}
& \emph{Orig.} & \emph{Gen.} & \emph{Orig.} & \emph{Gen.} & \emph{Orig.} & \emph{Gen.} & \emph{Orig.} & \emph{Gen.} & \emph{Orig.} & \emph{Gen.} & \emph{Orig.} & \emph{Gen.}& \emph{Orig.} & \emph{Gen.}\\ %
\midrule
\textbf{CondaQA}\\
\arrayrulecolor{black!20}\midrule
Original-Data Accuracy  
 & $73.9_{2.6}$ & \cellcolor{blue!25}$\mathbf{79.7_{1.2}}$ 
 & $70.5_{0.9}$ & \cellcolor{blue!25}$\mathbf{78.3_{1.4}}$ 
 & $68.5_{1.9}$ & \cellcolor{blue!25}$\mathbf{83.4_{1.4}}$
 & $67.5_{2.5}$ & \cellcolor{blue!25}$\mathbf{80.3_{0.9}}$
 & $70.2_{0.9}$ &\cellcolor{blue!25}$\mathbf{81.7_{1.4}}$
 & $81.4_{0.0}$ & $81.4_{0.0}$ 
 & $69.5_{0.0}$ & \cellcolor{blue!25}$\mathbf{78.3_{0.8}}$\\
\arrayrulecolor{black!20}\midrule
Full Bundle Consistency 
& $40.5_{2.7}$ & $\mathbf{44.5_{4.1}}$ 
& $32.6_{1.6}$ & \cellcolor{blue!25}$\mathbf{50.9_{5.2}}$ 
& $40.5_{2.1}$ & \cellcolor{blue!25}$\mathbf{47.3_{3.0}}$
& $41.9_{2.8}$ & $\mathbf{44.1_{5.0}}$
& \cellcolor{yellow!25}$\mathbf{54.0_{3.4}}$ & $39.5_{2.6}$
&\cellcolor{yellow!25}$\mathbf{53.0_{3.0}}$ & $40.9_{2.3}$ 
& $40.9_{2.1}$ & $\mathbf{44.5_{4.7}}$\\
Paraphrase Consistency  
& $69.5_{1.5}$ & \cellcolor{blue!25}$\mathbf{73.9_{1.0}}$ 
& $66.5_{1.6}$ & \cellcolor{blue!25}$\mathbf{72.5_{1.0}}$ 
& $65.5_{3.1}$ & \cellcolor{blue!25}$\mathbf{78.6_{1.8}}$
& $64.7_{1.6}$ & \cellcolor{blue!25}$\mathbf{72.1_{3.0}}$
& $69.1_{1.8}$ & $\mathbf{72.5_{2.7}}$
& $\mathbf{78.9_{1.6}}$ & $77.1_{1.5}$ 
& $70.2_{1.0}$ & $\mathbf{74.6_{1.5}}$\\
Scope Consistency       
& $53.3_{2.4}$ & \cellcolor{blue!25}$\mathbf{67.8_{1.7}}$ 
& $44.2_{0.9}$ & \cellcolor{blue!25}$\mathbf{64.5_{3.1}}$ 
& $49.2_{5.6}$ & \cellcolor{blue!25}$\mathbf{65.3_{2.0}}$ 
& $49.2_{3.2}$ & \cellcolor{blue!25}$\mathbf{62.0_{4.0}}$
& $59.6_{2.4}$ & $\mathbf{64.5_{1.1}}$
& $\mathbf{60.0_{2.7}}$ & $58.8_{1.7}$
& $49.2_{2.4}$ & \cellcolor{blue!25}$\mathbf{59.6_{3.4}}$\\
Affirmative Consistency 
& $55.9_{4.0}$ & $\mathbf{56.2_{2.1}}$ 
& $46.9_{3.2}$ & \cellcolor{blue!25}$\mathbf{63.8_{1.6}}$ 
& $51.8_{2.3}$ & \cellcolor{blue!25}$\mathbf{61.9_{3.4}}$ 
& $47.8_{2.7}$ & \cellcolor{blue!25}$\mathbf{56.2_{3.4}}$
& \cellcolor{yellow!25}$\mathbf{59.6_{3.4}}$ & $53.1_{3.2}$
& \cellcolor{yellow!25}$\mathbf{68.6_{1.8}}$ & $60.0_{0.9}$ 
& $54.7_{1.7}$ & $\mathbf{58.1_{2.1}}$\\
\arrayrulecolor{black}\midrule
\textbf{DROP}\\
\arrayrulecolor{black!20}\midrule
Original-Data Token F1  
& $61.7_{2.1}$ & \cellcolor{blue!25}$\mathbf{83.0_{3.1}}$ 
& $62.8_{2.6}$ & \cellcolor{blue!25}$\mathbf{83.9_{1.7}}$ 
& $75.2_{4.1}$& \cellcolor{blue!25}$\mathbf{92.7_{1.3}}$
& $73.2_{0.8}$& \cellcolor{blue!25}$\mathbf{90.3_{1.4}}$
& $71.9_{1.8}$ & \cellcolor{blue!25}$\mathbf{90.9_{0.0}}$
& $52.8_{2.0}$& \cellcolor{blue!25}$\mathbf{75.1_{2.8}}$
& $57.1_{1.8}$ & \cellcolor{blue!25}$\mathbf{77.9_{2.0}}$\\
\arrayrulecolor{black!20}\midrule
Compositional Consistency 
& $30.4_{3.6}$ & \cellcolor{blue!25}$\mathbf{58.8_{2.7}}$
& $29.2_3.0$ & \cellcolor{blue!25}$\mathbf{63.6_{0.9}}$ 
& $57.2_{3.0}$ & \cellcolor{blue!25}$\mathbf{80.8_{3.0}}$
& $55.6_{2.6}$ & \cellcolor{blue!25}$\mathbf{82.0_{2.4}}$
& $58.0_{2.4}$ &\cellcolor{blue!25}$\mathbf{86.4_{1.7}} $
& $25.2_{1.1}$ & \cellcolor{blue!25}$\mathbf{35.2_{3.3}}$
& $23.2_{2.3}$ & \cellcolor{blue!25}$\mathbf{48.4_{3.6}}$\\
\arrayrulecolor{black}\bottomrule
\end{tabular}%
}
\caption{Comparison of difficulty of the original CondaQA/DROP samples and their generated version measured by the performance with human-authored questions and edits against generated questions and edits. One author of this paper selected instances with valid generated questions and edits for this comparison. Human annotators provided answers for all of the valid generated questions and edits, and only those with at least 2/3 agreement were used for evaluation. Cells highlighted in \sethlcolor{blue!25}\hl{blue} indicate cases where scores on generated data were found to be significantly higher than the scores on the original data, and cells highlighted in  \sethlcolor{yellow!25}\hl{yellow} indicated cases where scores on original data were found to be significantly higher ($p < .05$).
$^\triangle$\texttt{gpt-4-turbo-2024-04-09} 
$^\circ$\texttt{gpt-4o-2024-11-20}
$^\diamond$\texttt{claude-opus-4-1-20250805}
$^\bigcirc$\texttt{gemini-2.5-flash}
$^\dagger$\texttt{o3-mini-2025-01-31}
$^\ddagger$\texttt{meta-llama/Llama-3.3-70B-Instruct}
$^\star$\texttt{Qwen/Qwen2.5-72B-Instruct}
}
\label{tab:chatdataset-difficulty}
\end{table*}

So far, we have observed that an LLM can often generate valid complex questions and contrastive edits along specific dimensions. 
Our goal was to assess the automation of creating \emph{challenging} reasoning benchmarks. %
However, validity according to the instructions does not necessarily imply difficulty, as hardness often emerges organically from creative and diverse annotations. %
To this end, we compare model performance on datasets containing only generated annotations against those with only original human annotations. %

\paragraph{Setup.} Generated questions may be ambiguous.  To filter unambiguous questions for benchmarking, we collect answers from three annotators per question and keep those for which two annotators agree. 
As NLP expertise is not required here, we recruit 59 annotators through \href{https://www.prolific.com/}{Prolific}, all screened using examples with known answers. 
Screenshots of the answering interfaces are shown in Fig.~\ref{fig:answering_screenshot} for CondaQA and Fig.~\ref{fig:drop_answer_1}--\ref{fig:drop_answer_5} for DROP. 
In total, we collected 1242 answers across two datasets at a cost of \$1744. 

We use all CondaQA-generated bundles that one author finds valid across the question and all edit types. 
For benchmarking on original data, we use CondaQA dev questions corresponding to the same passages that have answers in all four passages of their bundle.\footnote{If multiple such questions exist, we select one at random.} 
For DROP, we generate questions and compositional edits from passages in the contrast set with human compositional edits, and use question-edit pairs judged valid by one author. %
For benchmarking on original DROP, we first take compositional edits from the human-written dataset for all passages used to create the generated set. %
We then take the original test set questions from which those human-written compositional edits were created. 
We report question counts after ambiguity filtering in Table \ref{tab:retained_subsets} in the Appendix.\footnote{We manually check if 2 annotators refer to the same span with minor boundary disagreement. If so, the question is considered unambiguous, and one author sets the boundaries.} We provide an illustration of our full synthetic benchmark creation process in Figure \ref{fig:process_illustration} in the Appendix.

LLMs can be biased when they are used both as a data generator and a task solver \cite{maheshwari2024efficacysyntheticdatabenchmark}, which causes inflated performance.  
Thus, we evaluate LLMs from 6 families on the original and generated data: \texttt{GPT-4-Turbo}/\texttt{GPT-4o}\footnote{\url{https://openai.com/index/hello-gpt-4o/}}, \texttt{o3-mini}\footnote{\url{https://openai.com/index/openai-o3-mini/}}, \texttt{Claude-Opus-4.1}\footnote{\url{https://www.anthropic.com/claude/opus}}, \texttt{Gemini-2.5-Flash}\footnote{\url{https://deepmind.google/models/gemini/flash/}}, \texttt{Llama-3.3-70B}\footnote{\url{https://huggingface.co/meta-llama/Llama-3.3-70B-Instruct}},  \texttt{Qwen2.5-72b} \cite{qwen2.5}. %

We evaluate CondaQA zero-shot and DROP in a 3-shot setting, following prior work. 
CondaQA is evaluated using accuracy, as most answers are ``yes'', ``no'', or ``don't know''. 
For DROP, we use token-level F1, the harmonic mean of precision and recall over overlapping tokens between the prediction and the gold answer. 
Both datasets are also evaluated using \emph{consistency}, which is 1 if all instances in a bundle are answered correctly. 
For DROP, an instance is considered correct for consistency if its token-level F1 exceeds 0.8. 
CondaQA bundles consist of a question paired with four passages (``Full Bundle Consistency'' in Table \ref{tab:chatdataset-difficulty}) or a single passage edited in a specific way. 
DROP bundles consist of a question and its compositional edit in the context of the same passage. We perform 5 runs of the original and generated datasets on each of the models tested. The reported data is the mean value across those 5 runs, and the standard deviation of the mean is shown in subscript. Statistical significance testing is performed to measure the significance of the differences between model performances on the generated and original datasets. We use a mixed effects model to account for variance in the scores across runs, and a one-tailed t-test with $p < .05$ as a threshold for significance.

\paragraph{Results.} They are reported in Table \ref{tab:chatdataset-difficulty}, and show that LLMs' accuracy/token-F1 and consistency tend to be higher in model-generated datasets than in human-authored counterparts. Four out of six model families show higher accuracy/token-F1 and consistency across the board on the generated dataset, with most scores showing a significant increase in model performance. \texttt{Llama-3.3-70B} and \texttt{Gemini-2.5-Flash} on the CondaQA dataset have higher scores on the human-authored datasets, with full bundle consistency and affirmative edit consistency scores on the original dataset being significantly higher than their generated counterparts. Both models have much lower accuracy in answering generated affirmative edits as compared to human-authored edits. Looking into the evaluation logs, we find no discernible pattern as to why generated affirmative edits are harder for these models. However, these are the exceptions to the general trend. Looking at DROP scores, the generated dataset and compositional edits are shown to be significantly easier than the original DROP dataset across all models.
That is, generated versions are generally easier. %
We acknowledge that the results for the original DROP sample seem low. This is not due to a bug; we provide a detailed discussion in Appendix~\ref{sec:dif-details}.

Moreover, LLM-authored datasets do not preserve the relative ranking of models in CondaQA.   %
That is, synthetic evaluations currently cannot reliably serve as a substitute for human-authored ones, even when used solely for model comparison.%

Note that this occurs despite model-generated instances often being rated as more valid by humans. 
This finding suggests that even though generated datasets may closely follow data specifications, and appear to be high-quality to humans, they may feature systematic patterns or lack the creativity of human-authored datasets\,---\,thus making them more easily solvable by models. For example, we observe that the human-authored questions in CondaQA tend to be considerably longer (Figure \ref{fig:length_distribution}).%

\section{Conclusion}

We ask how benchmarks with LLM-generated content compare to human-authored versions. We summarize our main findings and conclusions here, and provide a more detailed discussion in Appendix~\ref{sec:discussion}.

We find that while generated content is often valid (\sect{sec:generations},\sect{sec:pref_study}), it tends to produce easier benchmarks (\sect{sec:difficulty}). Further, we find that these synthetically-generated datasets do not necessarily preserve model ordering%
, which could consequently lead practitioners to make different decisions around model development and deployment. Finally, we also find that these quality differences between synthetic and human-authored data, may not be perceptible to researchers inspecting the data. Thus, while LLMs are promising for data creation where complexity is less critical, reliable human annotators remain vital for new benchmarks assessing real-world generalization, corner cases, and testing models on nuanced, complex scenarios.

\section{Limitations}
Despite the contributions of our work, there are a few limitations to consider. 

First, we focus only on two reasoning-over-text benchmarks, but seven distinct generation objectives: implications, paraphrasing, changing negation scope, identifying and undoing negation, discrete reasoning, adding intermediate reasoning steps, and altering the temporal order of events. 
We were constrained by: 
(i) needing to become deeply familiar with intricate annotation guidelines, (ii) identifying key prompting ideas that ended up being dataset-specific (e.g., reasoning step types for DROP and changing the negation scope through a QA instance generation), (iii) reading passages in these datasets over a thousand times ourselves, (iv) recruiting expert annotators who were either pursuing or had completed a doctoral degree in NLP, (v) designing annotation templates, and (vi) the annotation cost of thousands of dollars. 

Second, despite extensive prompting, we may have missed strategies that yield more valid synthetic annotations. 
Likewise, another LLM might produce more valid generations than \texttt{gpt-4-turbo-2024-4-09}. 
However, greater validity does not imply greater difficulty, as our comparison of human and generated benchmarks shows. 
Critically, we first had to discover that valid generations (beyond labels) can still be notably easier. 
Now that this is known, improving difficulty would require further rounds of prompt review, costly answer annotation, and manual validity review.
A related open question is whether our structured prompts\,---\,designed to decompose the task into subtasks with subtask-specific verifiers\,---\,improved validity at the cost of difficulty. 
These prompts were crucial for producing valid generations, yet may have also made the task easier for LLMs. This further highlights the challenge of balancing validity with difficulty.

Finally, in this work, we look at two key metrics for benchmark quality: validity and complexity. However, machine-generated and human-generated benchmarks could differ on several other tertiary aspects, which are nevertheless important for building high-quality evaluations, such as diversity, coverage, and representativeness. We leave it to future work to compare these data-creation paradigms on these additional axes.

\section*{Acknowledgments}

We thank Jacob K.\ Johnson for the initial codebase and Dheeru Dua for sharing information about (contrastive) DROP. 
We are also grateful to the annotators for their contributions and to the UtahNLP lab for valuable feedback. 
Abhilasha Ravichander has been supported by NSF DMS-2134012 and ONR N00014-24-1-2207 while working on this project.

\clearpage
\appendix

\section*{Appendix Overview}
\label{sec:appendix}

\begin{figure}[!t] %
    \centering
    \includegraphics[width=1.0\columnwidth]{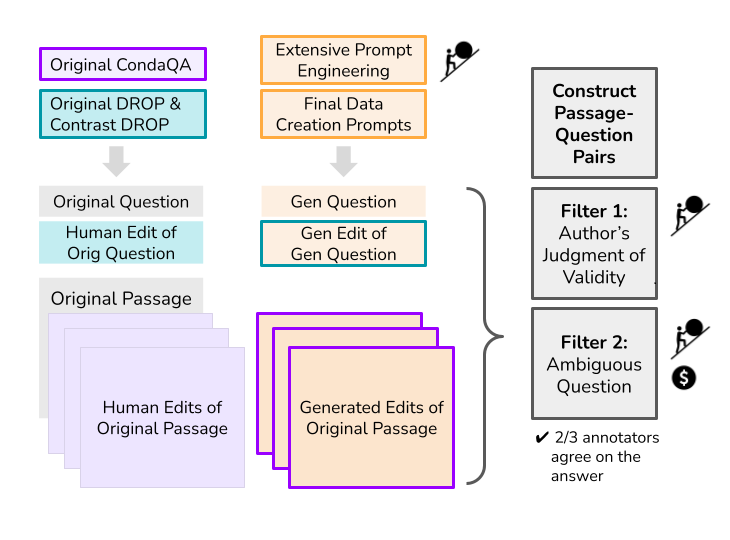} %
    \caption{Illustration of our synthetic dataset creation process. Purple borders around a block indicates relevance to CondaQA, teal borders indicate relevance to DROP. All other blocks are relevant for both datasets.}
    \label{fig:process_illustration}
\end{figure}

\begin{figure}[!t] %
    \centering
    \includegraphics[width=1.0\columnwidth]{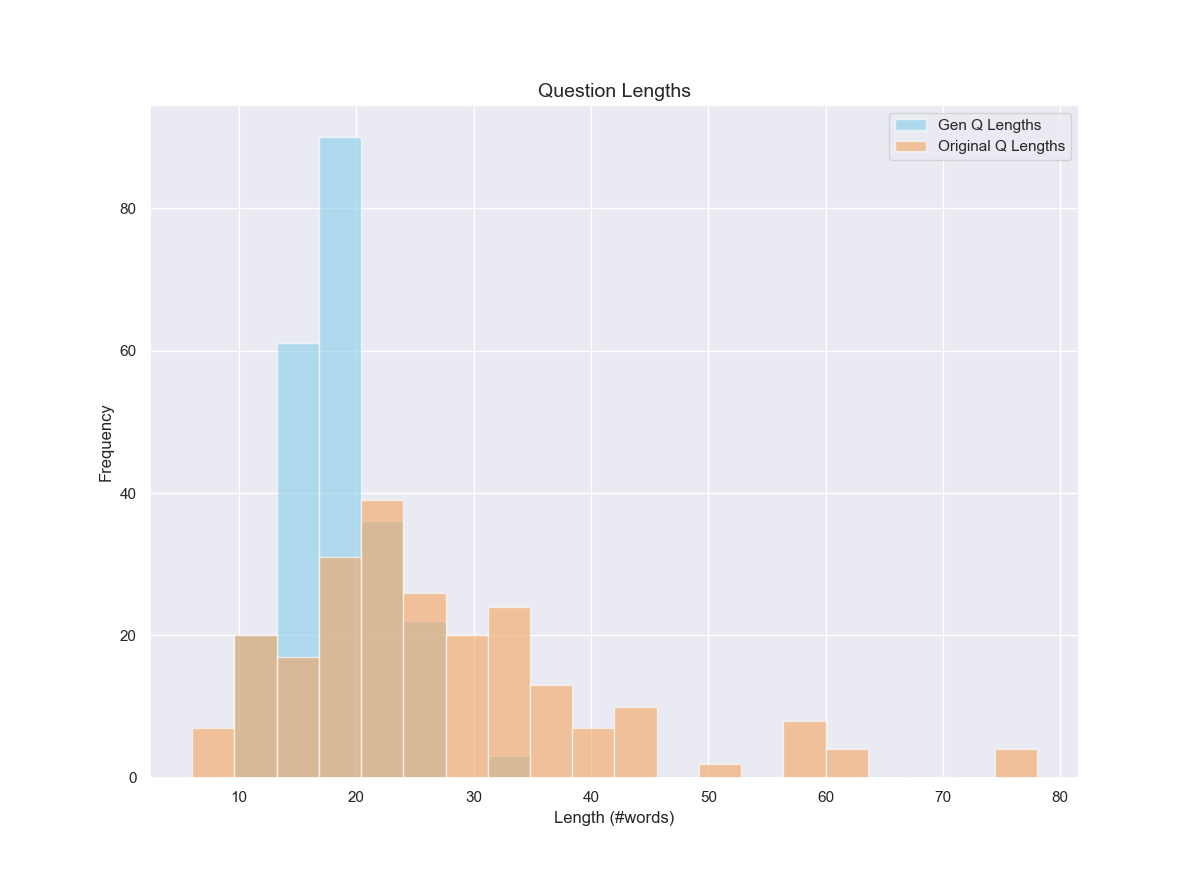} %
    \caption{Distribution of lengths of model-generated questions vs human-authored questions. We find that on average generated questions are 18.35 words, and the human-authored questions were 27.17 words long. We also find that 26.3\% of generated questions are more than 20 words long, but 67.68\% of human-authored questions were more than 20 words long.}
    \label{fig:length_distribution}
\end{figure}

\begin{figure*}[htbp]
    \centering
    \caption{\textbf{CondaQA} answer crowrdsourcing interface used in  \sect{sec:difficulty}.}
    \includegraphics[width=\textwidth]{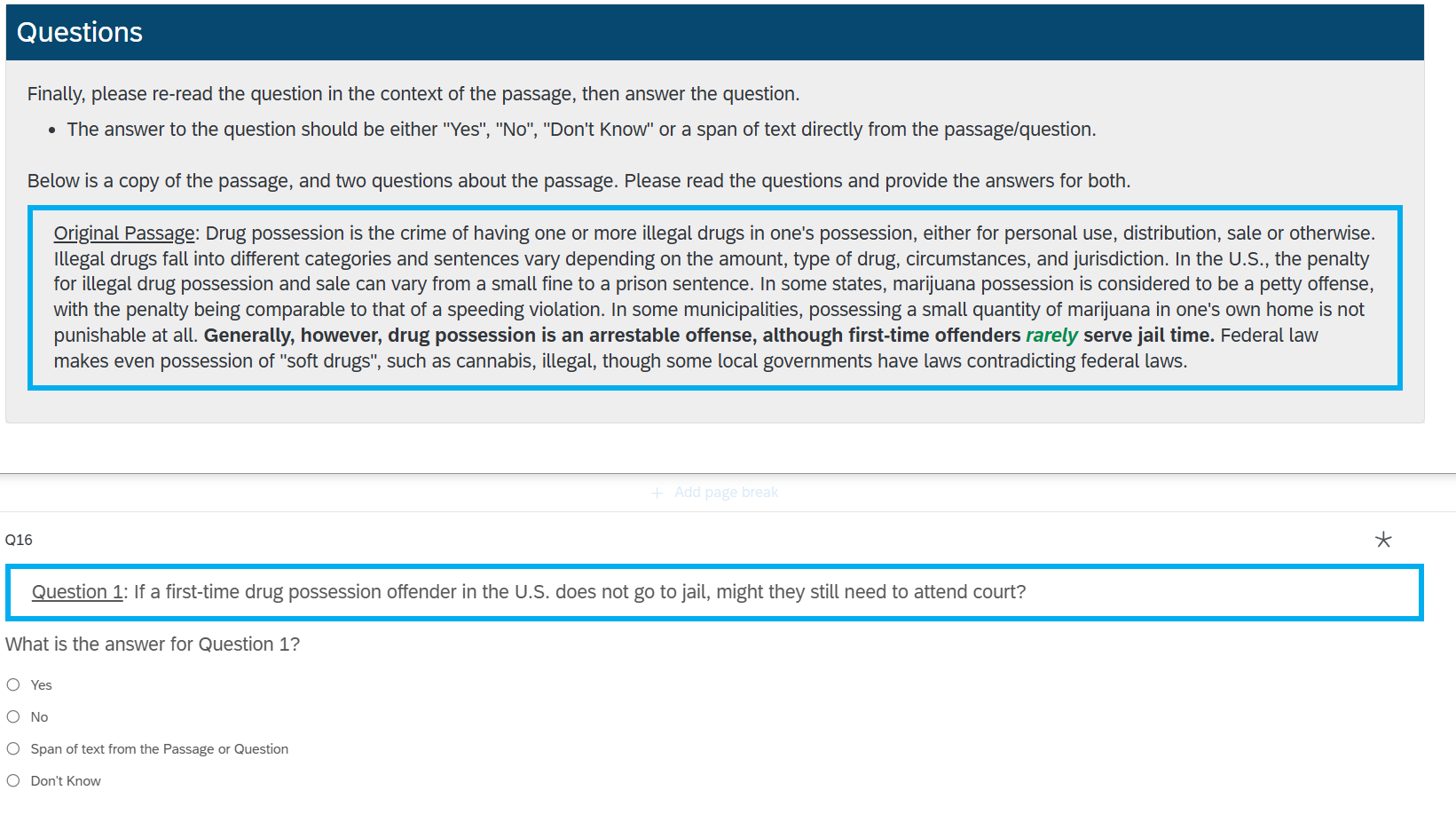}
    \label{fig:answering_screenshot}
\end{figure*}

\begin{figure*}[htbp]
  \centering
  \caption{The beginning of the instruction for the \textbf{DROP} answer crowdsourcing (\sect{sec:difficulty}). The instructions continues in the next figure.}
  \includegraphics[height=0.35\textheight,keepaspectratio]{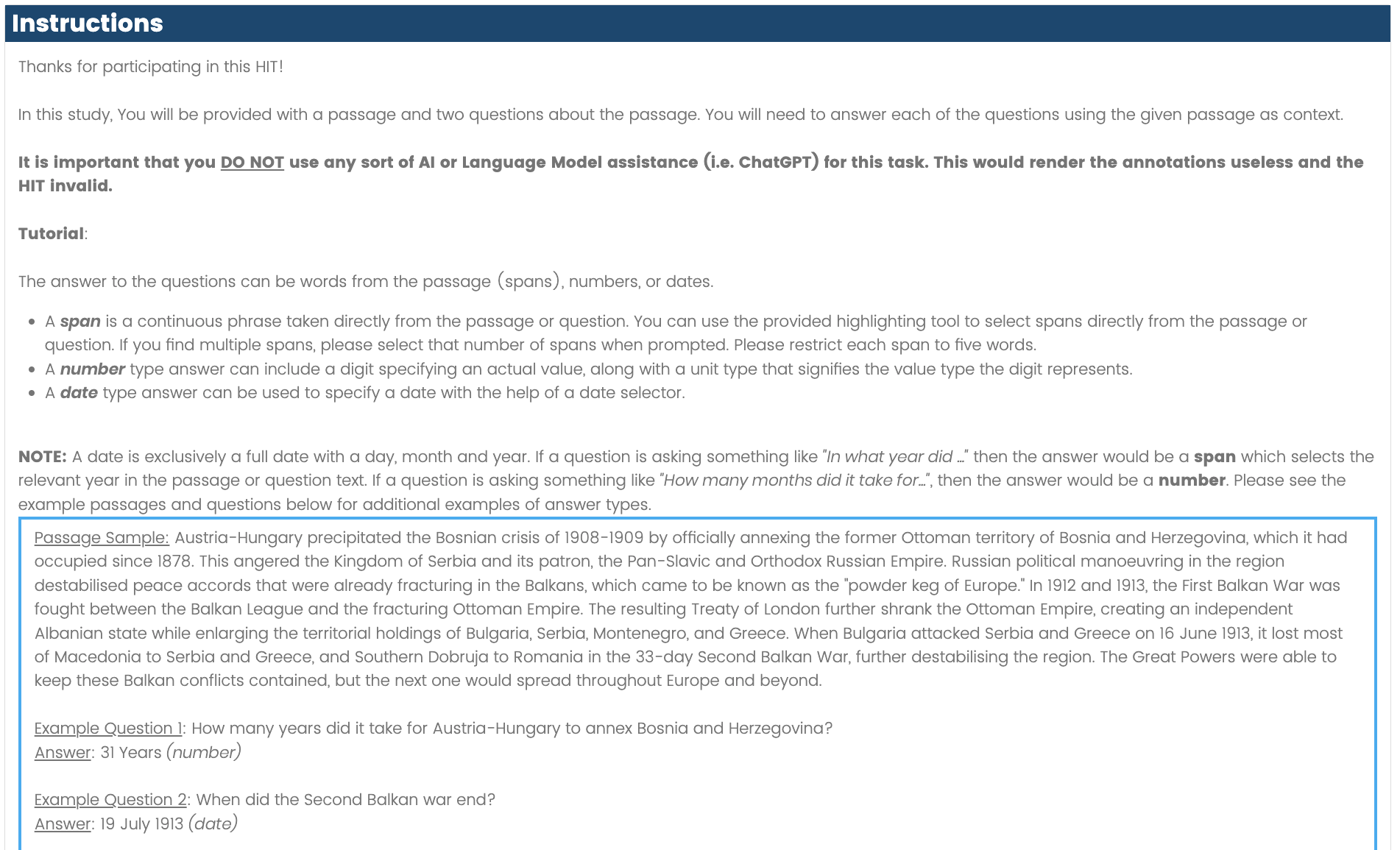}
  \label{fig:drop_answer_1}
\end{figure*}

\begin{figure*}[htbp]
  \centering
  \caption{The rest of the instruction for the \textbf{DROP} answer crowdsourcing (\sect{sec:difficulty}).}
  \includegraphics[height=0.35\textheight,keepaspectratio]{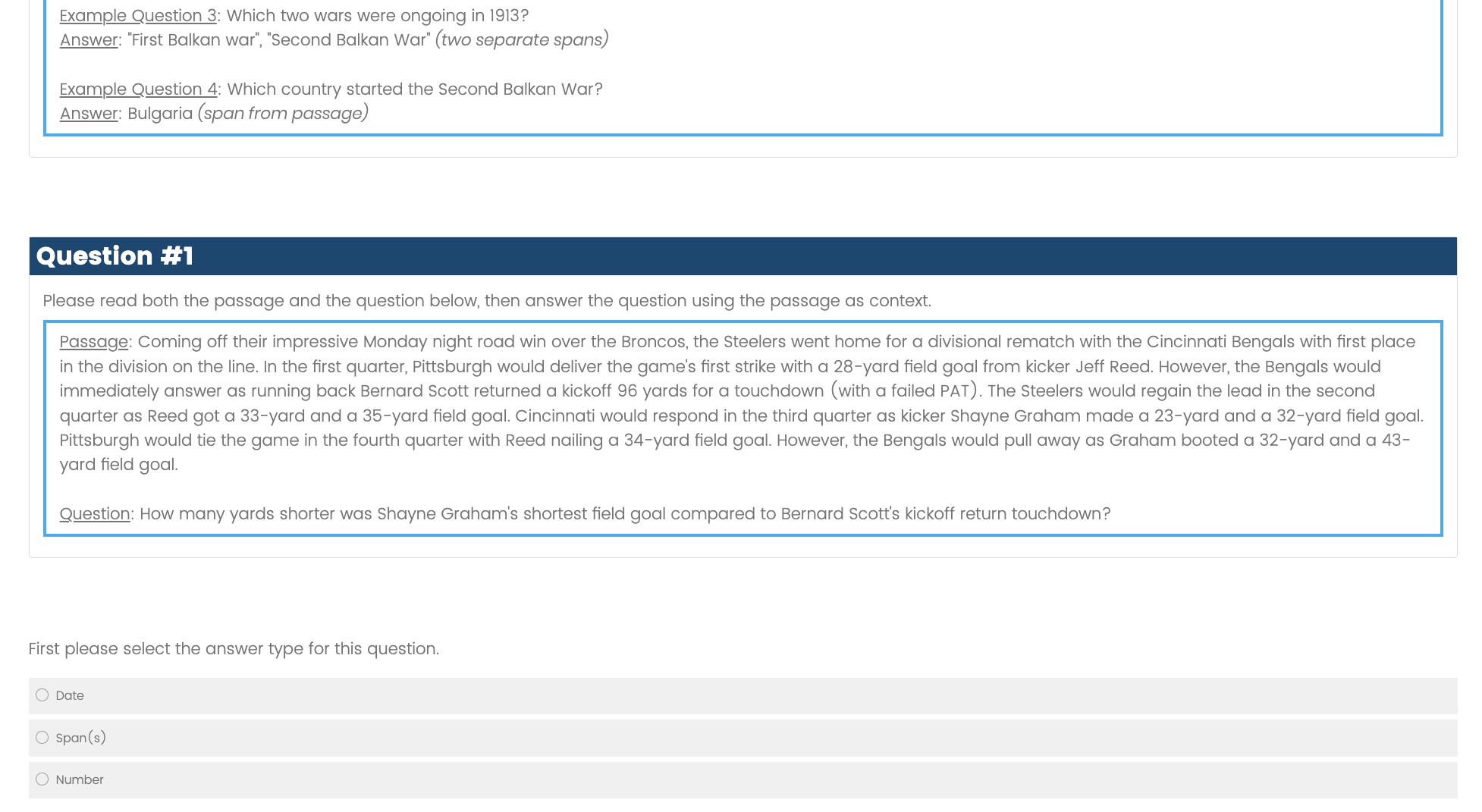}
  \label{fig:drop_answer_2}
\end{figure*}

\begin{figure*}[htbp]
  \centering
  \caption{Specifying the \textbf{DROP} \emph{numerical} answer (\sect{sec:difficulty}).}
  \includegraphics[height=0.25\textheight,keepaspectratio]{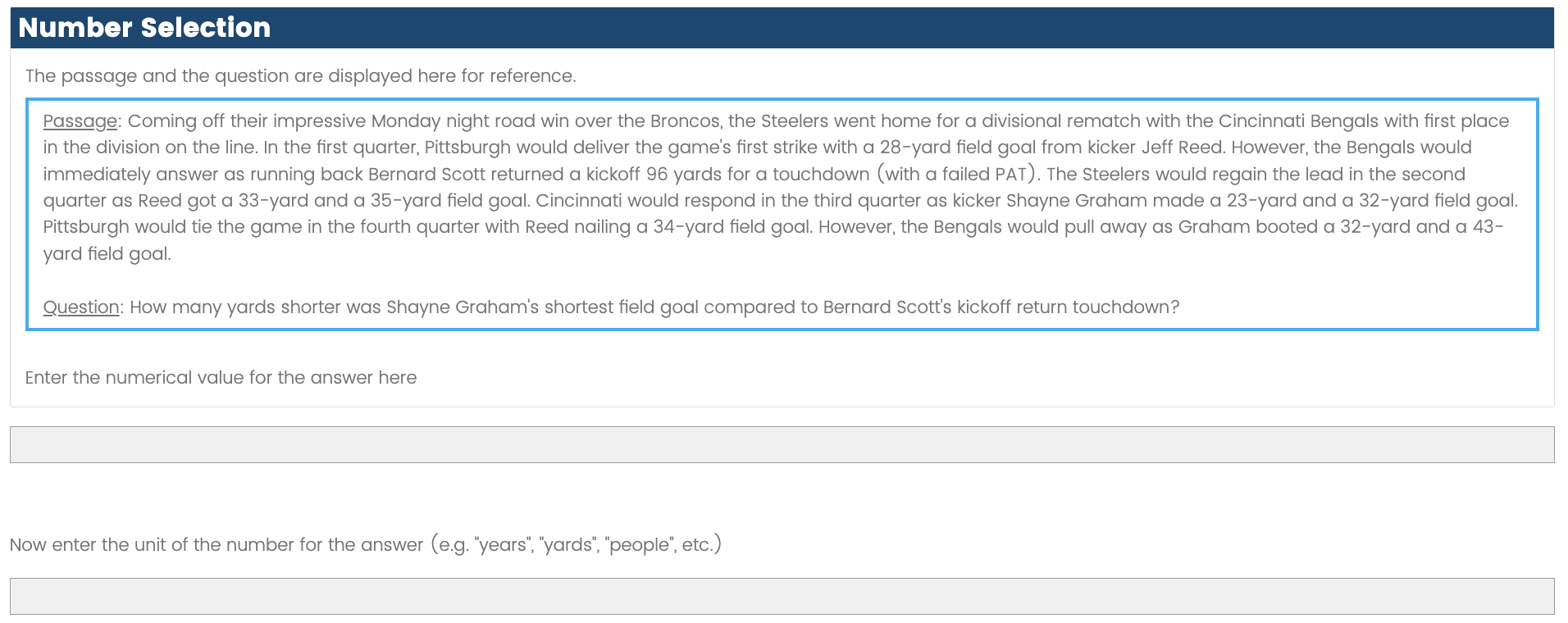}
  \label{fig:drop_answer_3}
\end{figure*}

\begin{figure*}[htbp]
  \centering
  \caption{Specifying the \textbf{DROP} \emph{span} answer (\sect{sec:difficulty}).}
  \includegraphics[height=0.4\textheight,keepaspectratio]{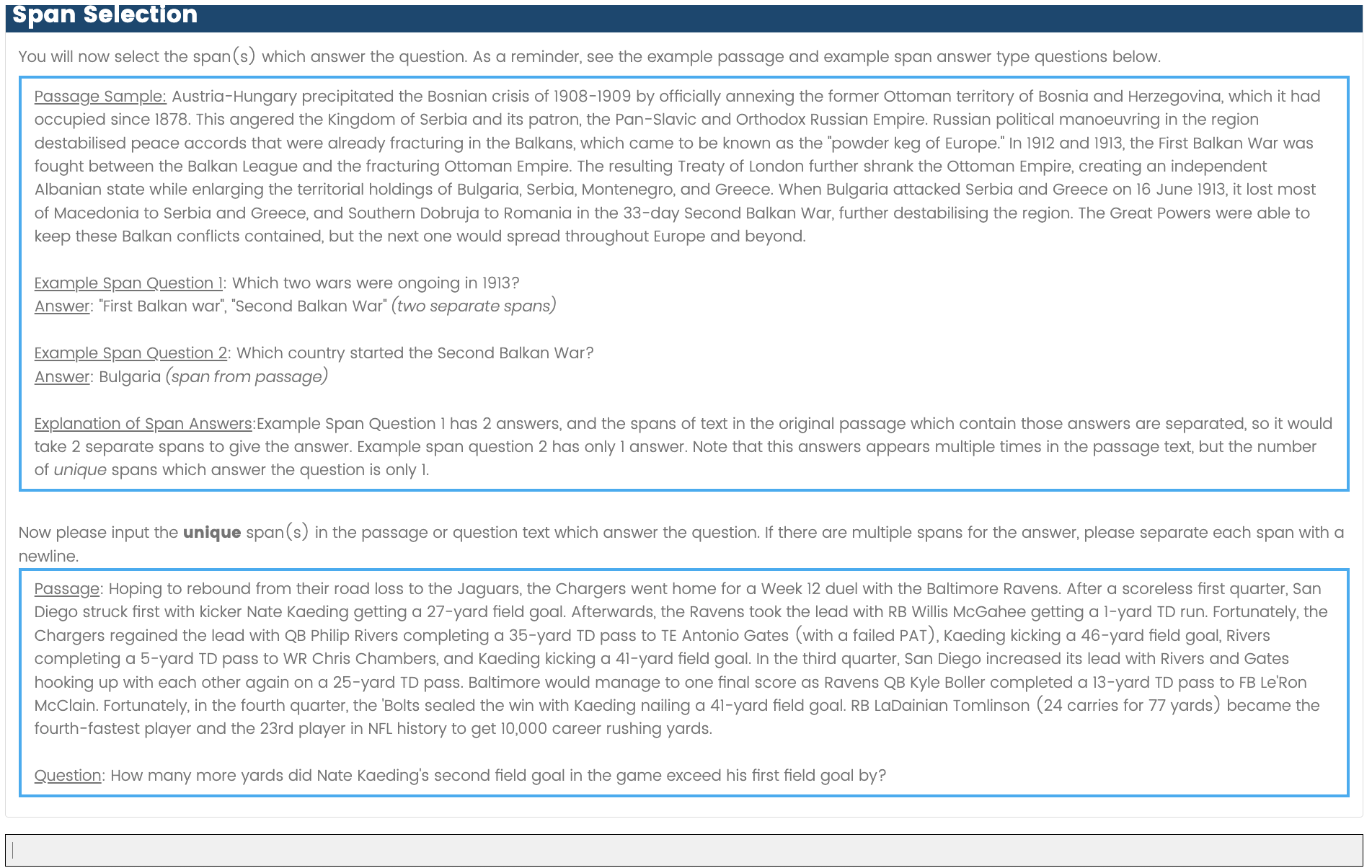}
  \label{fig:drop_answer_4}
\end{figure*}

\begin{figure*}[htbp]
  \centering
  \caption{Specifying the \textbf{DROP} \emph{date} answer (\sect{sec:difficulty}).}
  \includegraphics[height=0.4\textheight,keepaspectratio]{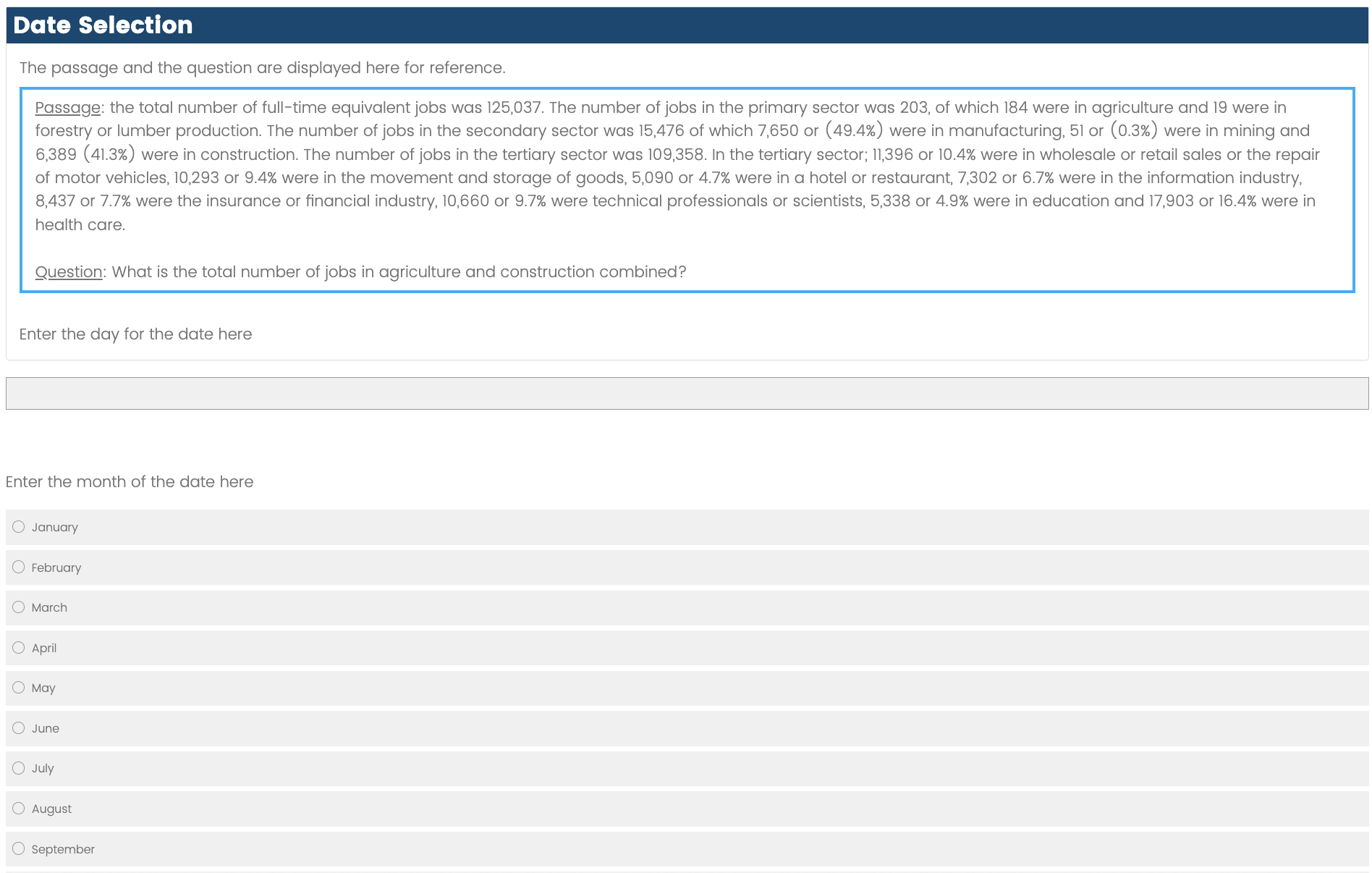}
  \label{fig:drop_answer_5}
\end{figure*}

In this Appendix, we provide:
\begin{compactitem}
    \item A description of prompting ideas that we found to be useful for generating valid benchmarks in Appendix \ref{sec:promptingideas}. 
    \item A contamination analysis of generated content in Appendix \ref{sec:memorization}.
    \item Details of the preference study in Appendix \ref{sec:pref_study_details}.
    \item A discussion of discrepancies between reported DROP results and our small-sample DROP  results in Appendix \ref{sec:dif-details}.
    \item A more detailed discussion of findings and takeaways in Appendix \ref{sec:discussion}.
    \item A version of Table \ref{tab:chat-benchmark-results} and Table \ref{tab:chatdataset-difficulty}, where instead of prompting the model to answer questions in the context of all related passages in a single chat, we use separate chats for each passage-question pair. See: Table \ref{tab:benchmark-results}. 
    \item Examples of human-authored questions and edits in Tables  \ref{tab:temporal_example}, \ref{tab:examples_drop_condaqa}, and  \ref{tab:condqa_edits}.
    \item Tables~\ref{tab:validity_questions_intermediate}--\ref{tab:edit_validity_with_intermediate}: Tabular versions of Fig.~\ref{fig:question_prompting}--\ref{fig:editing_prompting} (\sect{sec:generations}). 
    \item The validity criteria for each annotation item in Table \ref{tab:validity_criteria}.
    \item Screenshots of the preference study (\sect{sec:pref_study}) in Fig.~\ref{fig:pref_study_instructions}--\ref{fig:question_pref_choices} for CondaQA and Fig.~\ref{fig:drop_pref_1}--\ref{fig:drop_pref_9} for DROP. 
    \item Screenshots of answer crowdsourcing (\sect{sec:difficulty}) in Fig.~\ref{fig:answering_screenshot} for CondaQA and Fig.~\ref{fig:drop_answer_1}--\ref{fig:drop_answer_5} for DROP.
    \item Table \ref{tab:retained_subsets}: Question counts after ambiguity filtering in \sect{sec:difficulty}. 
    \item Figure \ref{fig:process_illustration}: Illustration of our synthetic benchmark creating process. Question and edit generation discussed in \sect{sec:generations}, label creation discussed in \sect{sec:difficulty}.
    \item Figure \ref{fig:length_distribution}: Distribution of lengths of model- vs. human-authored questions. Discussed in \sect{sec:difficulty}.
    \item Examples of the LLM-generated and human-authored questions in the two datasets: DROP (Table \ref{tab:dataset-questions_drop}) and CondaQA (Table \ref{tab:dataset-questions_condaqa}).
    \item Examples agreement and disagreement cases between annotators and one author of this paper for generated instances of CondaQA and DROP questions and edits in Tables \ref{tab:compositional-agreements-drop}-\ref{tab:affirmative-agreements-condaqa}.
\end{compactitem}

\section{Overview of Key Prompting Ideas}
\label{sec:promptingideas}

While our main goal was not to develop a novel data generation strategy, our prompting yielded several practical insights: 
\begin{compactitem}
\item Human annotation guidelines make poor prompts. 
\item Improving filtering is promising. 
\item Prompting strategies that succeed for one dataset may fail to work on another. 
\item Decomposing the annotation into sub-tasks was crucial.
\end{compactitem}

We outline some of the prompting strategies we found most useful, and the data they were used to create in Table~\ref{tab:prompting-strategies}.

\label{sec:examples_ogqa}

\section{Contamination Analysis}
\label{sec:memorization}

Our goal is to compare benchmarks with LLM-generated content to their counterparts with human-authored content.  
But what if generations are actually memorized human instances from the original dataset\,---\,a problem known as \emph{dataset contamination}. 
This could falsely lead us to conclude that synthetically-generated data is comparable to human-authored data, but these conclusions would then not generalize to new, unseeen datasets. 

To assess this, we run a qualitative analysis of potential contamination in the generated questions by comparing them to the original human-authored datasets. 
We randomly sampled 50 generated CondaQA questions and 50 for DROP. 
We use the generation created by the corresponding best prompts. 
For each, we \emph{manually} compared the generated question to all human-written questions associated with the same passage in the original dataset, and marked cases where the generated question was a paraphrase of an original. 
For the CondaQA dataset, we found that none of the 50 sampled questions were paraphrases of any of the questions from the original dataset. 
For DROP, we found that 9/50 of the sampled generated questions were paraphrases of questions from the original dataset.\footnote{One note with DROP is that many human-annotated questions were created for each passage, up to 70 for some.}  
From these results, we conclude that \emph{the original datasets are unlikely to be a significant source of contamination for the model-generated data.}

\section{Details of the User Study in \sect{sec:pref_study}}
\label{sec:pref_study_details}

\begin{figure*}[t]
    \centering
    \caption{The instructions for the \textbf{CondaQA} preference study (\sect{sec:pref_study}). }
    \includegraphics[width=\textwidth]{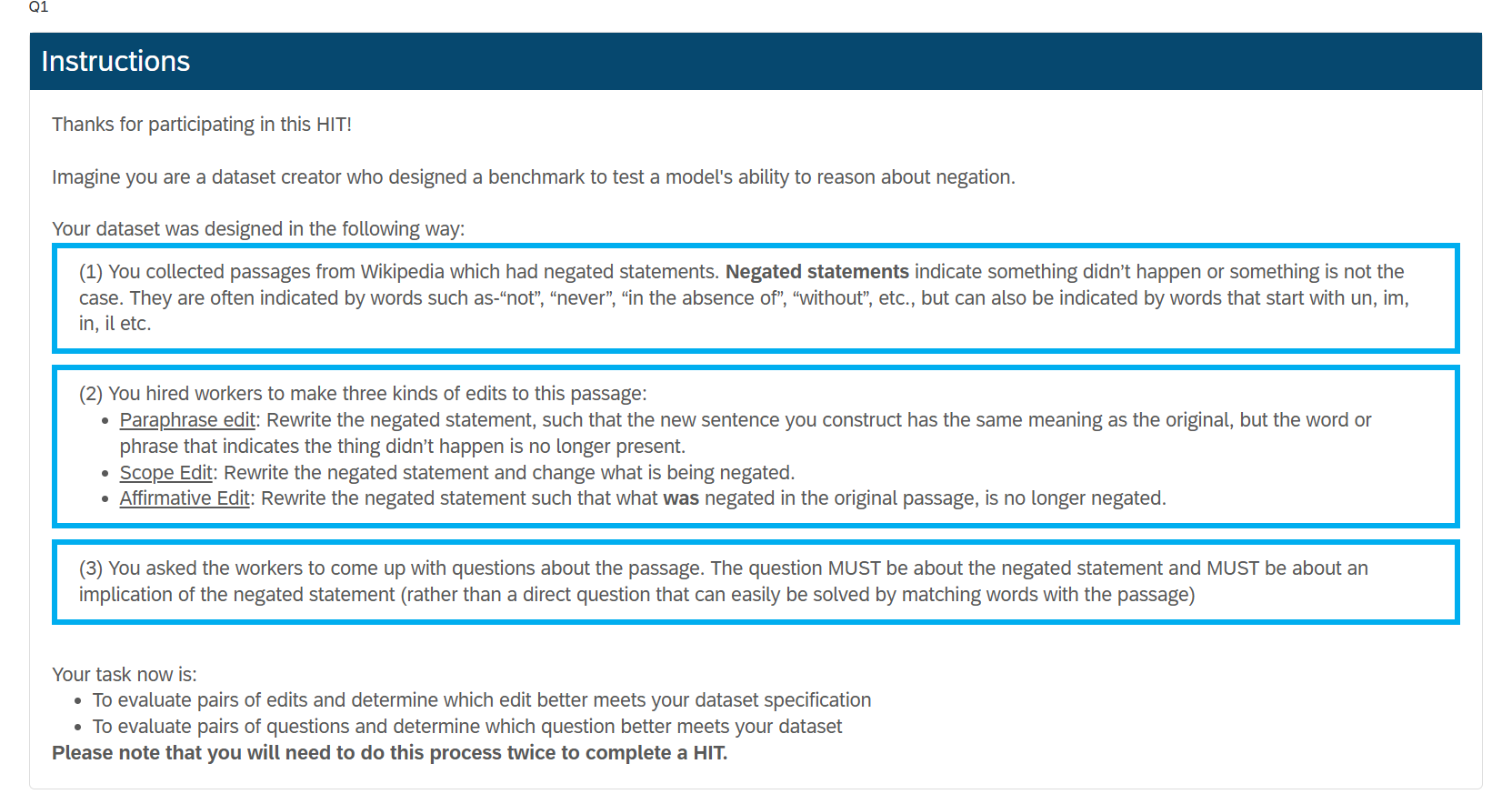}
    \label{fig:pref_study_instructions}
\end{figure*}

\begin{figure*}[htbp]
    \centering
    \caption{This screenshot shows what comes after the instructions in the \textbf{CondaQA} preference study (\sect{sec:pref_study}): the original passage.}
    \includegraphics[width=\textwidth]{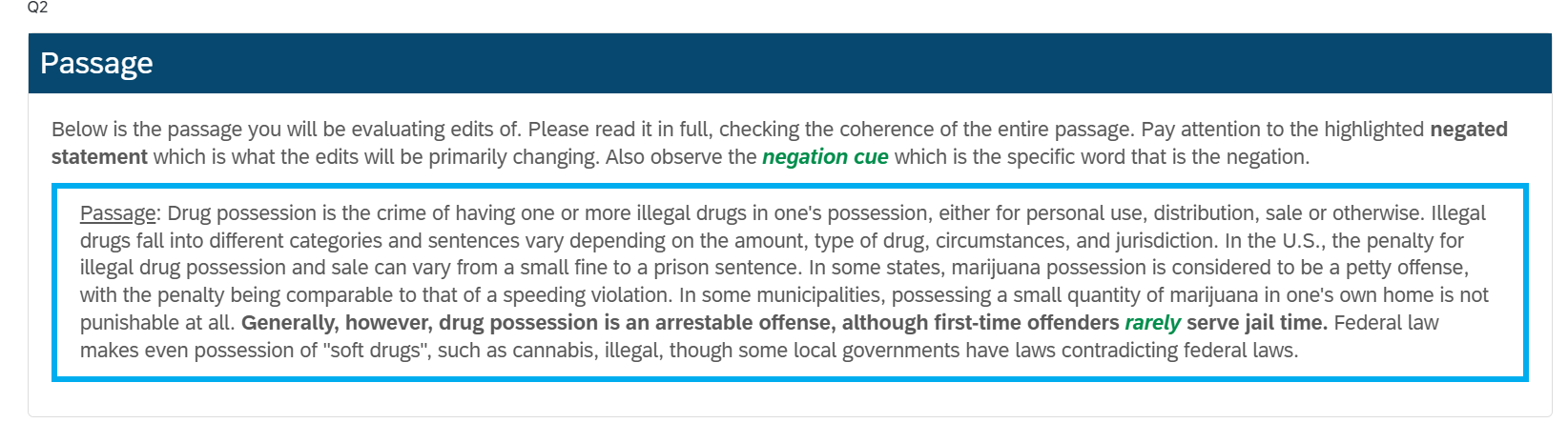}
    \label{fig:pref_study_orig_p}
\end{figure*}

\begin{figure*}[htbp]
    \centering
    \caption{After the original passage (Fig.\ \ref{fig:pref_study_orig_p}), annotators get a definition of a paraphrase edit and two paraphrase edits. }
    \includegraphics[width=\textwidth]{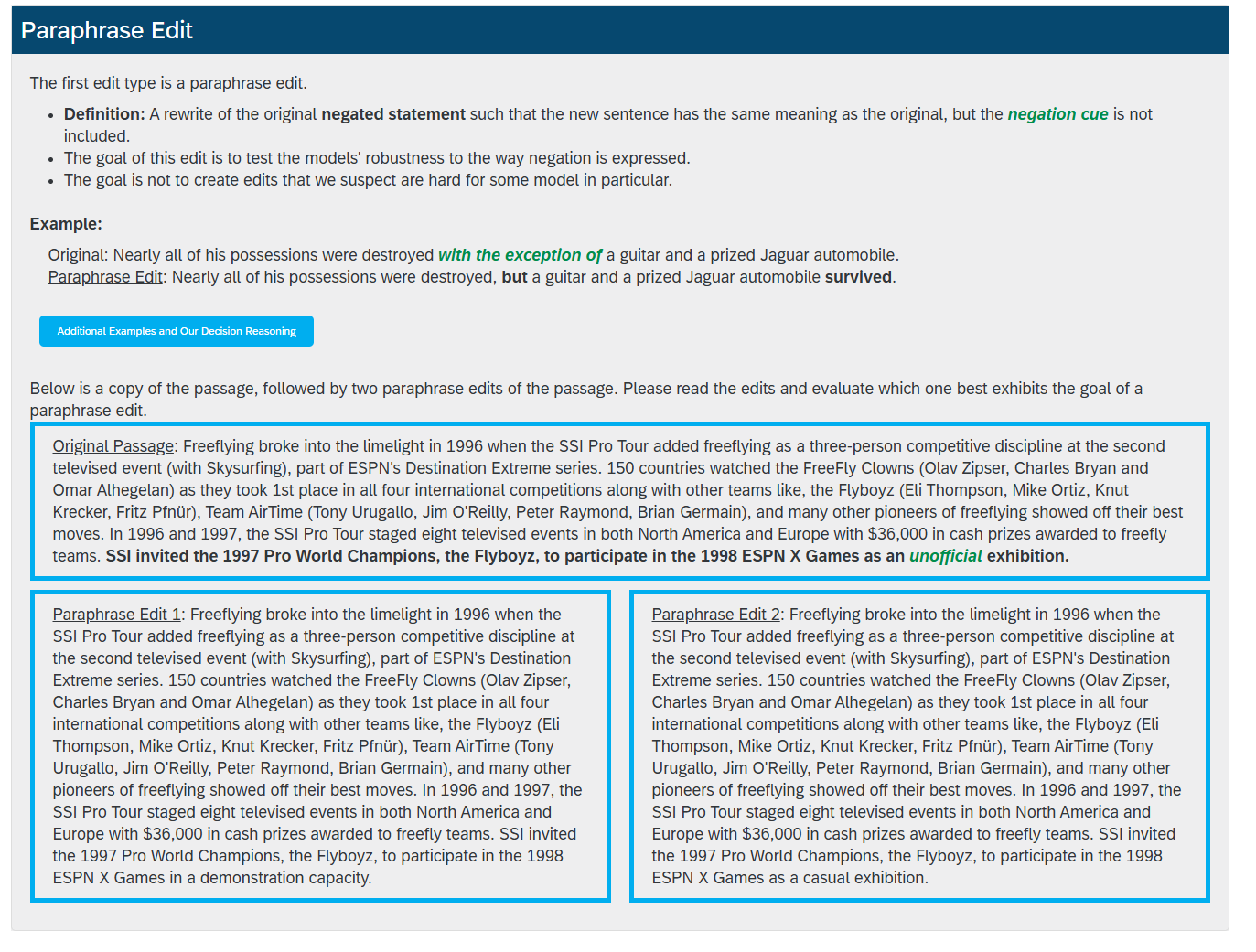}
    \label{fig:pref_study_pref_p}
\end{figure*}

\begin{figure*}[htbp]
    \centering
    \caption{The options annotators select for edits in the preference study (\sect{sec:pref_study}). Here shown for paraphrase edits.}
    \includegraphics[width=\textwidth]{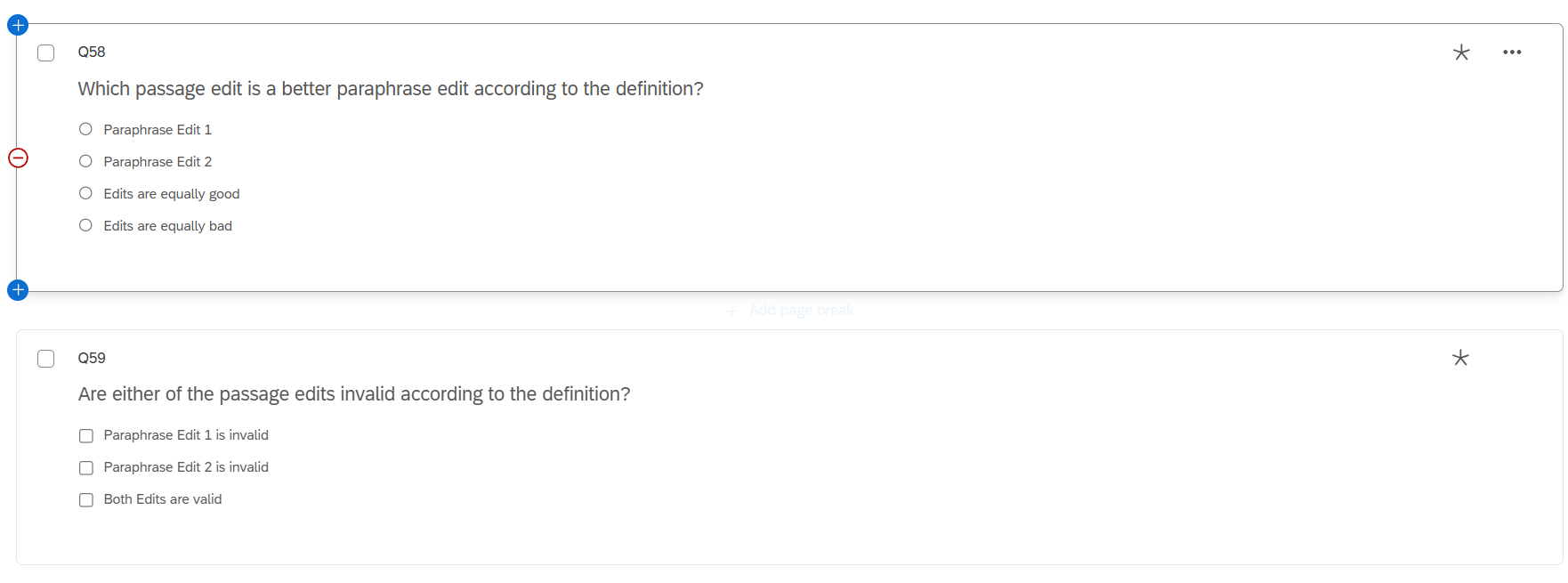}
    \label{fig:pref_study_edit_options}
\end{figure*}

\begin{figure*}[htbp]
    \centering
    \caption{After the paraphrase edit selection (Figures~\ref{fig:pref_study_pref_p}--\ref{fig:pref_study_edit_options}), annotators get the scope edit definition and a pair of two scope edits.}
    \includegraphics[width=\textwidth]{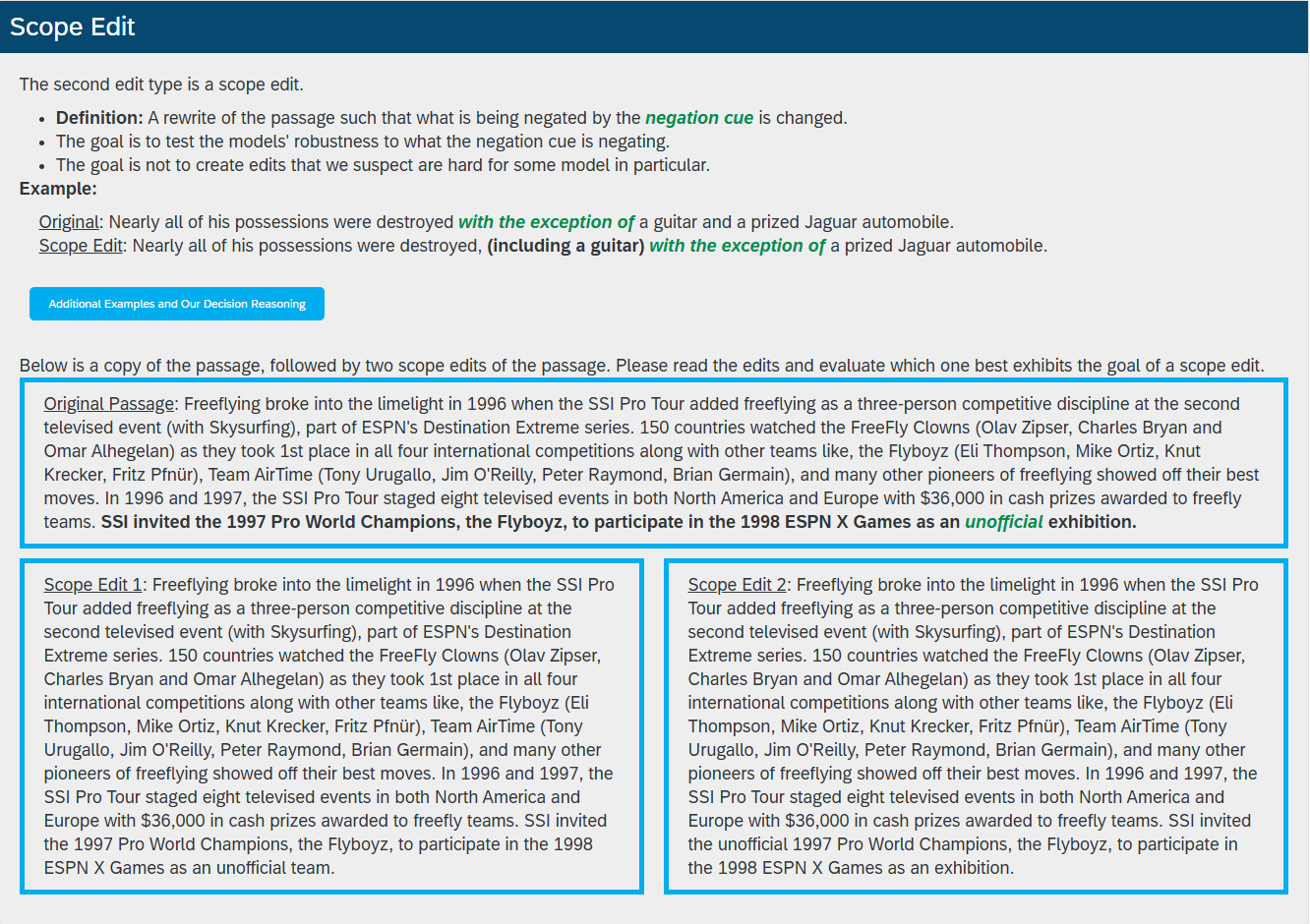}
    \label{fig:pref_study_scope}
\end{figure*}

\begin{figure*}[htbp]
    \centering
    \caption{After the scope edits (Fig.\ \ref{fig:pref_study_scope}), annotators get the affirmative edit definition and a pair of two affirmative edits.}
    \includegraphics[width=\textwidth]{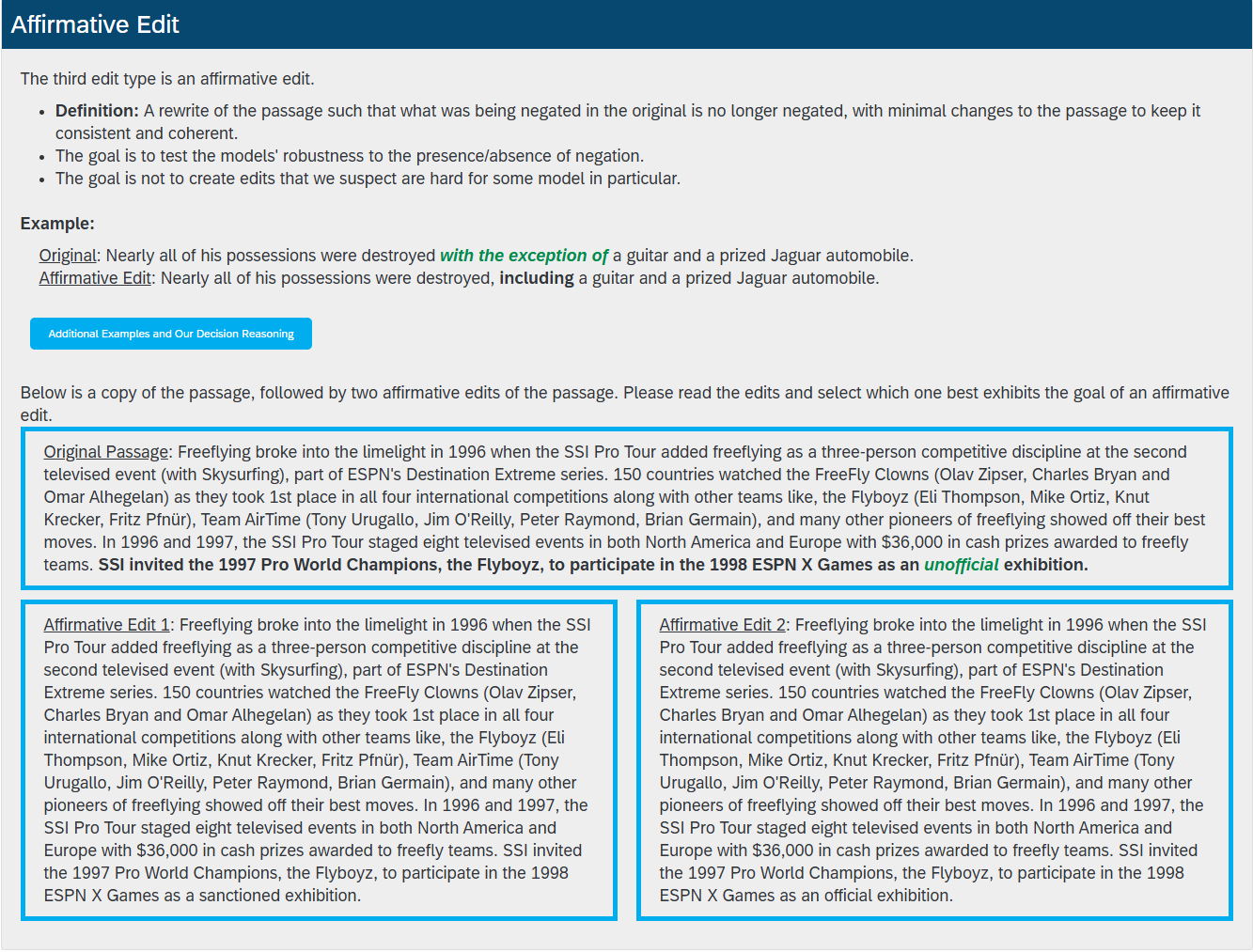}
    \label{fig:pref_study_aff}
\end{figure*}

\begin{figure*}[htbp]
    \centering
    \caption{After the affirmative edits (Fig.\ \ref{fig:pref_study_aff}), annotators get the CondaQA-style question definition and a pair of two questions.}
    \includegraphics[width=\textwidth]{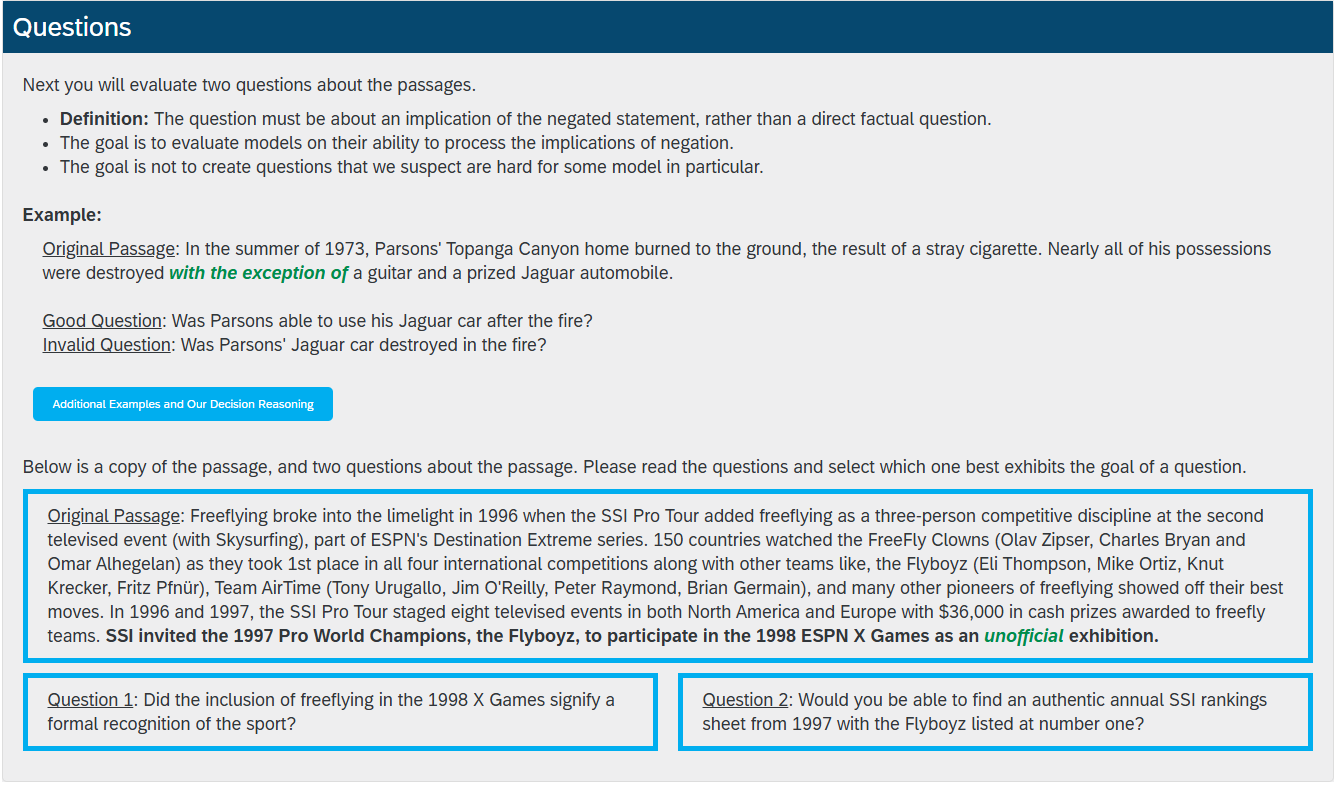}
    \label{fig:pref_study_questions}
\end{figure*}

\begin{figure*}[htbp]
    \centering
    \caption{The choices for stating a preference between two questions.}
    \includegraphics[width=0.8\textwidth]{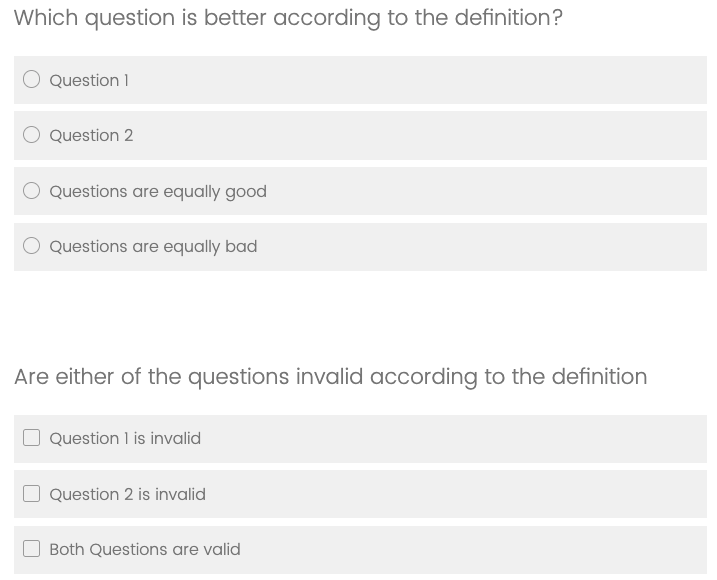}
    \label{fig:question_pref_choices}
\end{figure*}

\begin{figure*}[p]
  \centering
  \caption{The beginning of the instruction for the \textbf{DROP} preference study (\sect{sec:pref_study}).}
  \includegraphics[height=0.9\textheight,keepaspectratio]{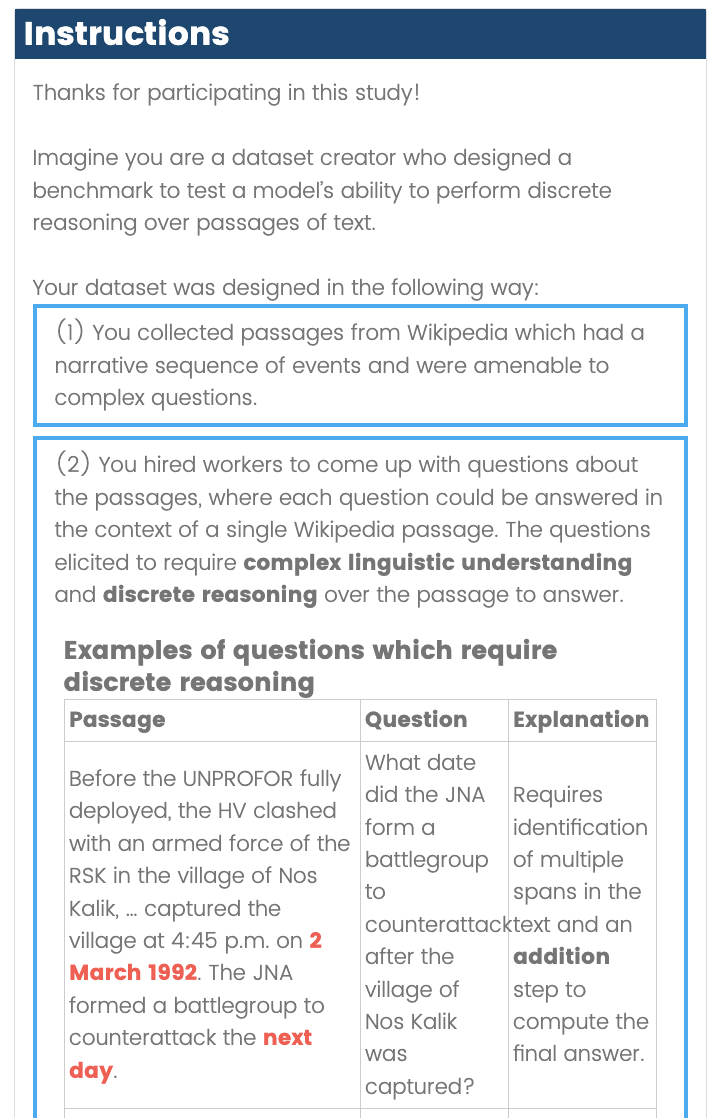}
  \label{fig:drop_pref_1}
\end{figure*}

\begin{figure*}[p]
  \centering
  \caption{The middle of the instruction for the \textbf{DROP} preference study.}
  \includegraphics[height=0.9\textheight,keepaspectratio]{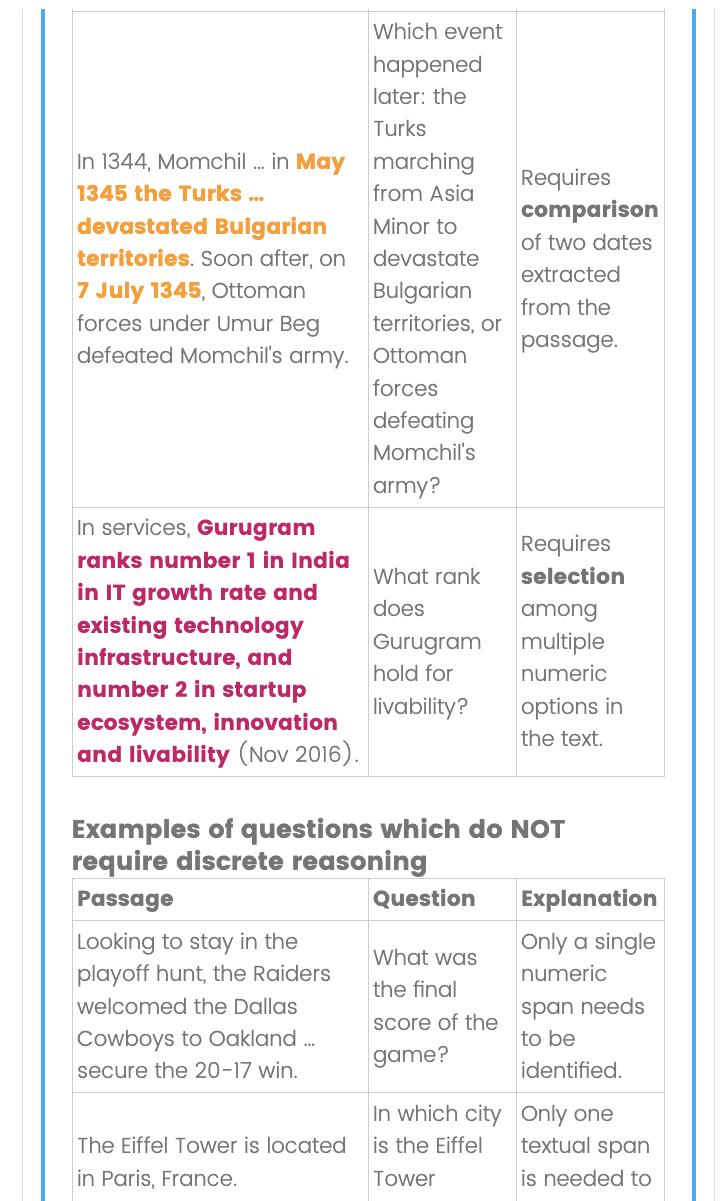}
  \label{fig:drop_pref_2}
\end{figure*}

\begin{figure*}[p]
  \centering
  \caption{The end of the instruction for the \textbf{DROP} preference study.}
  \includegraphics[height=0.8\textheight,keepaspectratio]{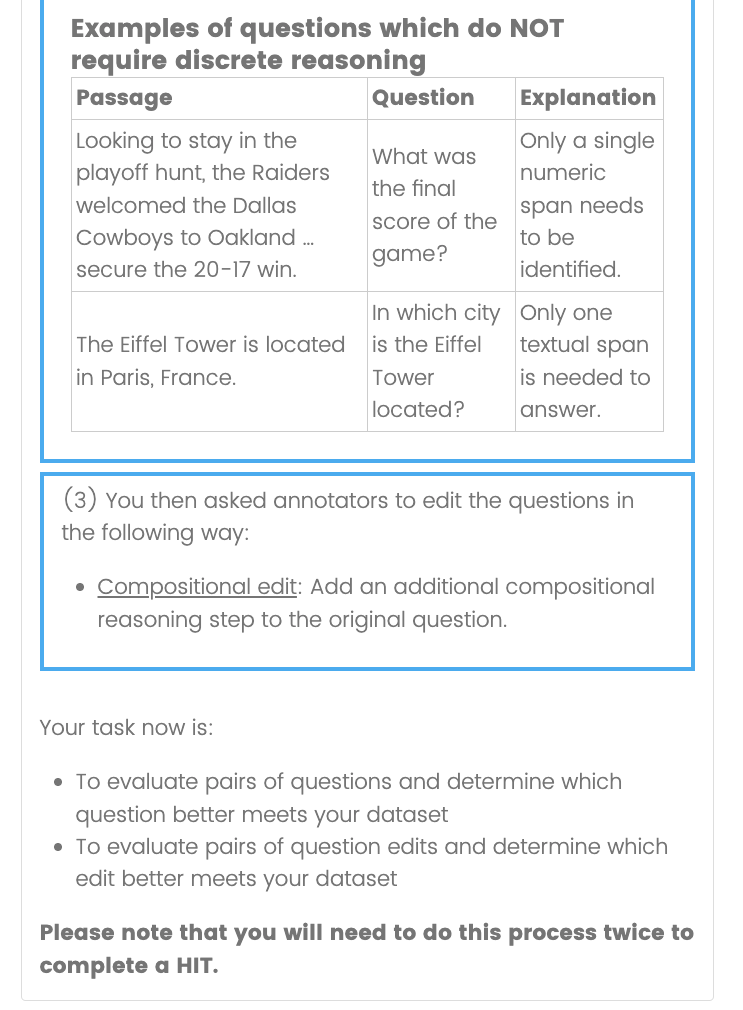}
  \label{fig:drop_pref_3}
\end{figure*}

\begin{figure*}[t]
  \centering
  \caption{After the general instructions, annotators are familiarized with DROP-like questions.}
  \includegraphics[height=0.8\textheight,keepaspectratio]{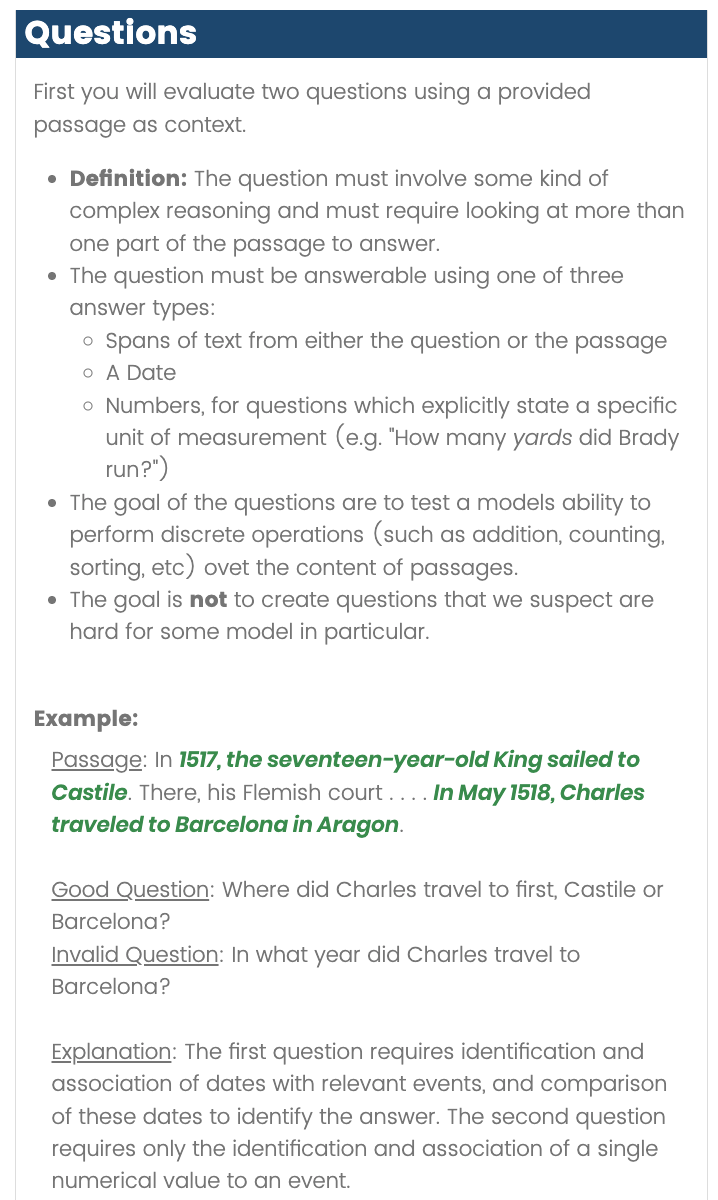}
  \label{fig:drop_pref_4}
\end{figure*}

\begin{figure*}[t]
  \centering
  \caption{The annotators are asked to give their preference between two DROP-like questions. One of the questions is generated, but that is not disclosed. The options are the same as for CondaQA; see Figure~\ref{fig:question_pref_choices}.}
  \includegraphics[height=0.8\textheight,keepaspectratio]{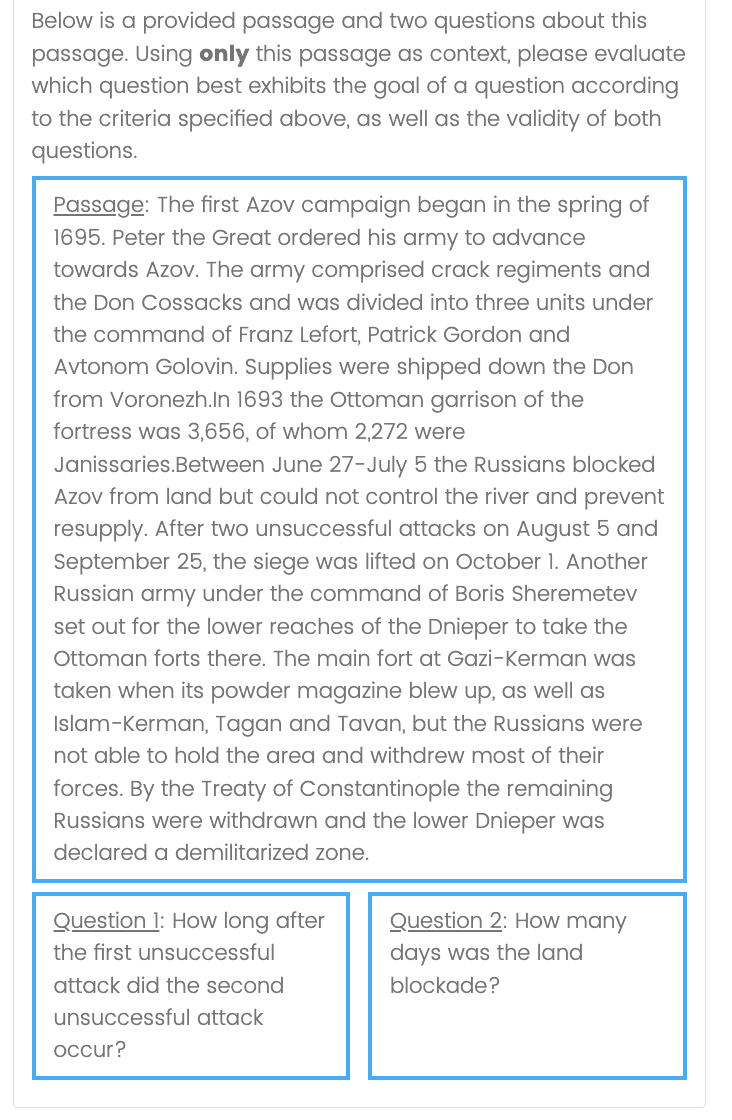}
  \label{fig:drop_pref_5}
\end{figure*}

\begin{figure*}[t]
  \centering
  \caption{After selecting a preferred question (if any), annotators are familiarized with DROP compositional edits. The instruction continues in the next figure.}
  \includegraphics[height=0.65\textheight,keepaspectratio]{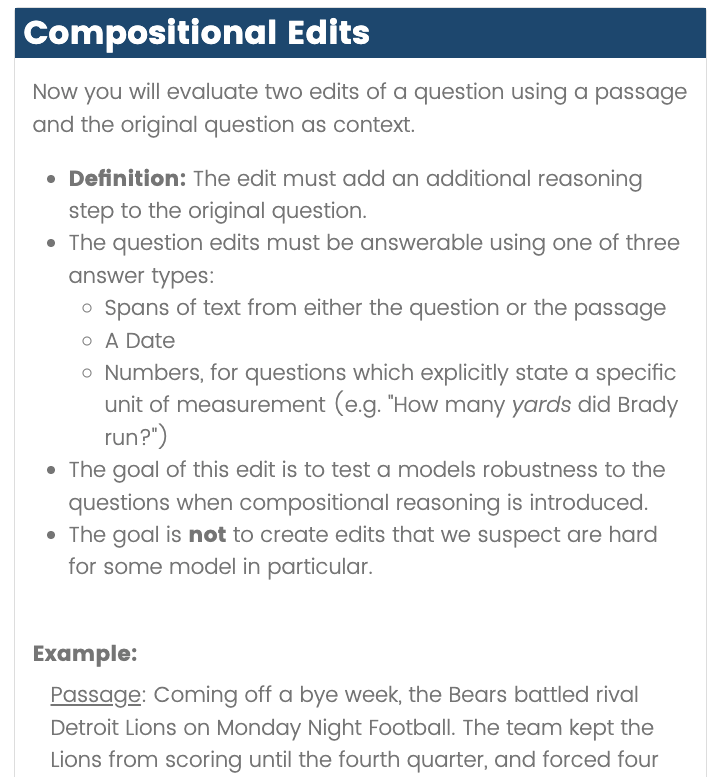}
  \label{fig:drop_pref_7}
\end{figure*}

\begin{figure*}[t]
  \centering
  \caption{The end of the instruction for the compositional edits.}
  \includegraphics[height=0.8\textheight,keepaspectratio]{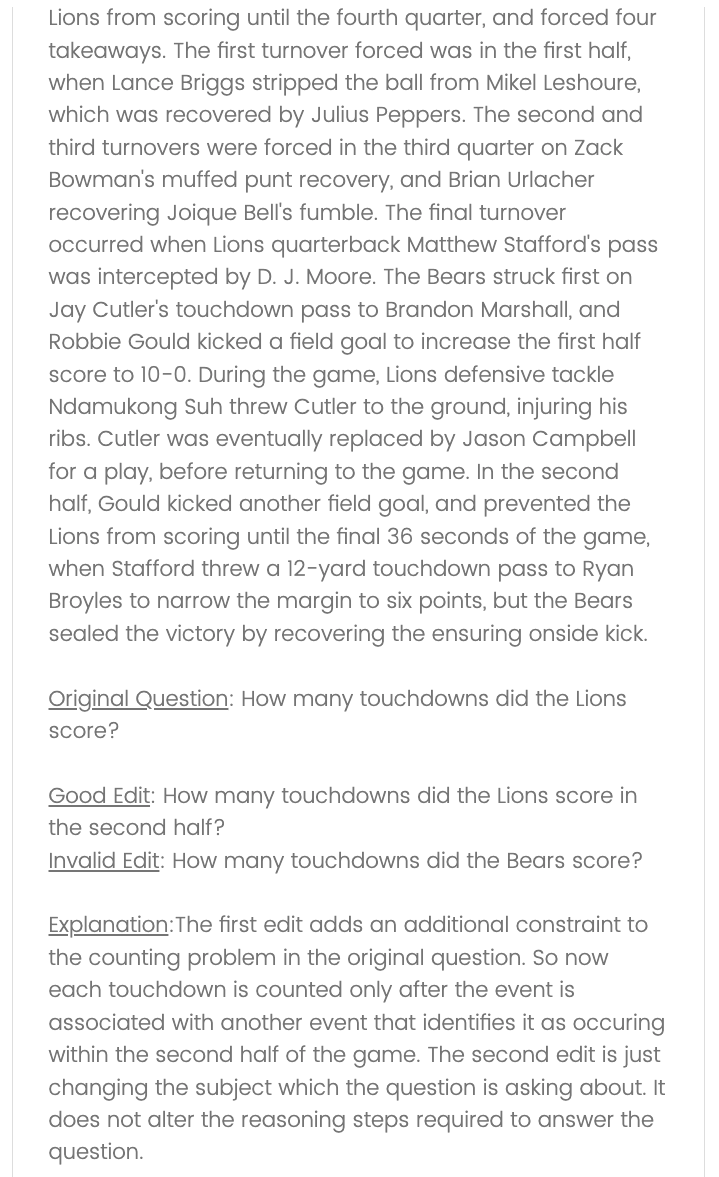}
  \label{fig:drop_pref_8}
\end{figure*}

\begin{figure*}[t]
  \centering
  \caption{The annotators are asked to give their preference between two DROP-like compositional edits. One of the edits is generated, but that is not disclosed. The options are the same as for CondaQA; see Figure~\ref{fig:pref_study_edit_options}.}
  \includegraphics[height=0.8\textheight,keepaspectratio]{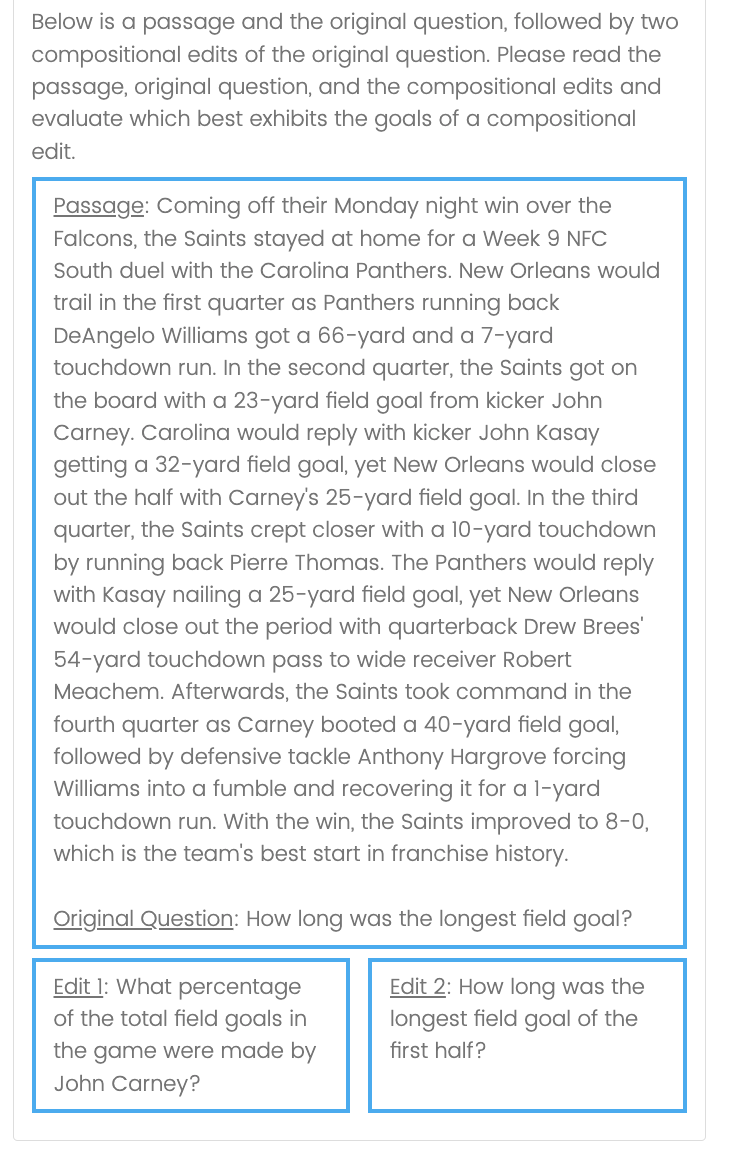}
  \label{fig:drop_pref_9}
\end{figure*}

Screenshots of the preference study are shown in Figures~\ref{fig:pref_study_instructions}--\ref{fig:question_pref_choices} for CondaQA and Figures~\ref{fig:drop_pref_1}--\ref{fig:drop_pref_9} for DROP. All participants in the study first provided informed consent.

In the DROP preference study, annotators first choose between two questions written for a given passage, and then choose between two edits. 
However, these edits are based on a different passage.
This mismatch arises because we reuse the generated questions from \sect{sec:generations}, but not the compositional edits. 
In \sect{sec:generations}, compositional edits are made on generated questions to simulate building a synthetic dataset from scratch. 
To enable a valid comparison between human and generated compositional edits in \sect{sec:pref_study}, we instead apply edits to human-authored questions that also have corresponding human-authored compositional edits.

For CondaQA, this complication does not occur. Annotators are shown pairs of questions and edits based on the same passage, and they express preferences directly over these pairs.

\section{DROP Sample Difficulty Sanity Checks}
\label{sec:dif-details}

We observe that the results on the sample of the original, human-authored DROP dataset in Table \ref{tab:chatdataset-difficulty}  are notably lower than reported DROP results. 
To verify our evaluation setup for DROP, we benchmark  \texttt{gpt-4o-2024-11-20} on a larger sample of 1,000 passage-question pairs from the original DROP test set. 
We obtain a token-F1 score of 83.1, which is in line with the reported token-F1 score for DROP for \texttt{gpt-4o-2024-11-20} (81.5)\footnote{\url{https://github.com/openai/simple-evals}}. 
This confirms that the low performance on the smaller sample is not due to a bug in our evaluation scripts.

We reiterate here that we did not choose the DROP sample for Table~\ref{tab:chatdataset-difficulty}, but use the sample of instances which have human compositional edits from the DROP contrastive set. 

One explanation for the difficulty of this sample could be the high ratio of questions with numerical answers as compared to the original DROP test set (77\% vs 61.6\%). Token-F1 for numerical answers reduces to the exact match to the gold label, which is stricter than the token-F1 evaluation for spans or dates. However, the ratio of numerical questions in the generated data is actually higher than the original sample (96.6\%), so we reject the hypothesis that the ratio of numerical questions alone is why the human-authored DROP benchmark in Table~\ref{tab:chatdataset-difficulty} is harder than the generated one. 

Moreover, for CondaQA, we do not observe the same discrepancy between performance on smaller and larger samples. Specifically, the full dev set results in Table~\ref{tab:chat-benchmark-results} for \texttt{o3-mini-2025-01-31} and \texttt{meta-llama/Llama-3.3-70B-Instruct}  line up with the results for the original data in Table~\ref{tab:chatdataset-difficulty}.

\section{Discussion and Practical Advice}
\label{sec:discussion}

\paragraph{Quality Tradeoffs of Automatic Evaluation.}  Using LLMs to generate datasets and validating them through sampling misses crucial quality metrics. Such assessments can miss out on other factors such as complexity and representativeness. Specifically, although generations may be valid, and may even appear preferable to human judges, we might not even realize what is missing: the nuances, complexity, and creativity of human-authored data. 

\paragraph{Cost of Constructing Evaluations.} For constructing benchmarks in this study, LLMs are orders of magnitude cheaper than crowdsourcing. We recommend synthetic evaluations for tasks where complexity and representativeness are not the primary criteria for measuring dataset quality. For example, models could be utilized to generate cheap and large-scale unit tests for models~\cite{naik-etal-2018-stress,ribeiro-etal-2020-beyond}, while human-constructed evaluations can be utilized complementarily for assessing real-world generalization, identifying corner cases, and validating model behavior on complex, nuanced scenarios.

\paragraph{Feedback Mechanisms.} Yet another aspect of constructing high-quality evaluations is the ability to give feedback that influences the annotation process. In the crowdsourcing scenario, dataset creators often engage in two rounds of iterations\,---\,an initial ``pilot study'' to refine instructions based on annotator data, and often a second stage (once instructions are finalized) where crowdworkers are given feedback on their annotations~\cite{nangia-etal-2021-ingredients}. Applying similar feedback loops to synthetic evaluations remains a challenge. While our prompt iterations resemble pilot studies, future work could explore feedback mechanisms\,---\,human or AI-driven\,---\,to help models refine their outputs.

\paragraph{On Human Effort while Seeking Difficulty.} Our findings suggest that generating valid instances through prompting is not the main challenge\,---\,it is finding prompts that generate hard content. To find a better prompt for difficulty, in theory, one could iteratively prompt an LLM to generate candidate instances, have one or more people validate them, recruit three or more people per instance to get reliable answers, benchmark model performance on the resulting set, and repeat this until the difficulty level matches that of a human-authored dataset. Basically, repeat what we did in this paper many times. But, in practice, executing such a procedure would require substantial human-in-the-loop effort. Moreover, this turns evaluation-data generation into a dataset construction with an adversary in the loop that prior work has raised concerns about.

\newpage
\begin{table}[t]
\centering
\resizebox{\columnwidth}{!}{%
\begin{tabular}{lrrr}
\toprule
\textbf{CondaQA} & \textbf{Human$^\dagger$} & \textbf{\texttt{o3-mini}}$^\triangle$ & \textbf{\texttt{Llama-3.3}}$^\circ$ \\
\midrule
Original-Data Accuracy & 91.9  & 71.5   & 77.78  \\
\arrayrulecolor{black!20}\midrule
Bundle Consistency & 81.6 & 40.3  & 36.2  \\
Paraphrase Consistency & 93.6  & 67.3 & 78.1  \\
Scope Consistency & 86.5 & 51.5  & 51.0  \\
Affirmative Consistency & 88.2 & 48.5 & 54.1  \\
\arrayrulecolor{black}\toprule
\textbf{DROP} & \textbf{Human$^\ddagger$} & \textbf{o3-mini}$^\triangle$ & \textbf{\texttt{Llama-3.3}}$^\circ$ \\
\midrule
Original-Data Token F1 & 96.4 & 84.3  & 70.9  \\
\arrayrulecolor{black!20}\midrule
Consistency & N/A  &  53.5  & 35.5  \\
\arrayrulecolor{black}\bottomrule
\end{tabular}%
}
\caption{State-of-the-art performance on the CondaQA development set and the DROP contrastive test set. The original accuracy is calculated on the data used to create contrastive instances. Consistency reflects the rate at which all minimally different examples within a bundle are answered correctly. \textbf{Here, we use separate chats for each passage-question pair.} \emph{These results show that these reasoning benchmarks remain challenging. We investigate whether they could be created using LLMs.} $^\dagger$Cf.\  \citet{ravichander-etal-2022-condaqa}. $^\ddagger$Cf.\ \citet{dua-etal-2019-drop}.
$^\triangle$\texttt{o3-mini-2025-01-31} 
$^\circ$\texttt{meta-llama/Llama-3.3-70B-Instruct}. %
}
\label{tab:benchmark-results}
\end{table}

\begin{table}[!h]
\centering
\resizebox{\columnwidth}{!}{
\begin{tabular}{p{4.2cm}p{4.2cm}}
\toprule
\textbf{Original} &
\textbf{Temporal Edit} \\
\midrule
\textbf{In the spring of 1625} the Spanish regained Bahia in Brazil and Breda in the Netherlands from the Dutch. \textbf{In the autumn} they repulsed the English at Cadiz.
&
\textbf{In the spring of 1625} the Spanish regained Bahia in Brazil and Breda in the Netherlands from the Dutch. \textbf{In winter the year earlier} they had repulsed the English at Cadiz.  \\
\bottomrule
\end{tabular}
}
\caption{Example of human-authored temporal DROP-paragraph edit. %
} 
\label{tab:temporal_example}
\end{table}

\begin{table*}[t]
\centering
\resizebox{\textwidth}{!}{
\begin{tabular}{llr}
\toprule
 & \textbf{Context} & \textbf{Count} \\
\midrule
\multirow{7}{*}{\rotatebox{90}{\textbf{CondaQA}}}
  & \texttt{is-valid}(Gen Question, Original Passage) $\wedge$ \texttt{is-valid}(Gen Question, Gen $x$-Edit Passage) & \multirow{2}{*}{44}\\
  & $\hookrightarrow \forall x \in \{\text{Paraphrase, Scope, Affirmative}\}$ \\
  \arrayrulecolor{lightgray}\cmidrule{2-3}
 &  \texttt{is-valid}(Gen Question, Original Passage) $\wedge$ \texttt{is-valid}(Gen Question, Gen Paraphrase-Edit Passage) & 56 \\
   \arrayrulecolor{lightgray}\cmidrule{2-3}
 &  \texttt{is-valid}(Gen Question, Original Passage) $\wedge$ \texttt{is-valid}(Gen Question, Gen Scope-Edit Passage) & 49 \\
   \arrayrulecolor{lightgray}\cmidrule{2-3}
 &  \texttt{is-valid}(Gen Question, Original Passage) $\wedge$ \texttt{is-valid}(Gen Question, Gen Affirmative-Edit Passage) & 52 \\
   \arrayrulecolor{lightgray}\cmidrule{2-3}
  & \texttt{is-valid}(Gen Question, Original Passage)              & 59 \\
\arrayrulecolor{black}\midrule
\multirow{4}{*}{\rotatebox{90}{\textbf{DROP}}}
  &  \texttt{is-valid}(Original Passage, Gen Question) $\wedge$ \texttt{is-valid}(Original Passage, Gen Compositional Edit of Gen Question) & 50 \\
    \arrayrulecolor{lightgray}\cmidrule{2-3}
  &\texttt{is-valid}(Original Passage, Gen Question)                     & 66 \\
    \arrayrulecolor{lightgray}\cmidrule{2-3}
  & \texttt{is-valid}(Original Passage, Gen Compositional Edit of Generated Question)                      & 52 \\
  \arrayrulecolor{black}\bottomrule
\end{tabular}
}
\caption{The count of generated instances that are unambiguous. The validity is determined by one of the authors. %
}
\label{tab:retained_subsets}
\end{table*}

\begin{table*}[t]
\centering
\footnotesize
\begin{tabular}{p{1.1cm}p{7cm}p{3cm}p{1.25cm}p{1.5cm}}
\toprule
{\bf Dataset} & {\bf Passage} (some parts shortened) & {\bf Question} & {\bf Answer} & {\bf Reasoning}\\
 \midrule
 DROP & In {\bf \color{red} 2008}, at another auction at Christies, Ulrich sold a 1982 {\bf \color{red} Basquiat piece}, Untitled (Boxer), for {\bf \color{red} US \$13,522,500} to an anonymous telephone bidder. {\bf \color{red} Another record price for a Basquiat painting was made in 2007}, when an untitled Basquiat work from 1981 {\bf \color{red} sold at Sothebys in New York for US\$14.6 million}. In 2012, [...] That year, [...] was sold [...] for \$16.3 million [...] A similar untitled piece [...] sold for \pounds12.92 million [...] In 2013, [...] sold for \$48.8 million [...] In 2016 an untitled piece sold at Christies for \$57.3 million [...] & How many more dollars was the Basquiat sold in 2007 than in 2008? \newline \hdashrule{3cm}{0.4pt}{2pt} \emph{Compositional}\newline\emph{Contrastive Question:}\newline How many more dollars was the Basquiat sold in 2016 than in 2008 and 2007 combined? & \$1,077,500 \newline\newline\newline \hdashrule{1.25cm}{0.4pt}{2pt} \emph{Contrast Answer:} \$29,177,500  & Discrete Reasoning \\ \midrule
CondaQA  & The school song, `Fair Reed,' is sung to the tune of the 1912 popular song `Believe Me, if All Those Endearing Young Charms.' It may be imitative of the Harvard anthem "Fair Harvard," which is also sung to the tune of `Believe Me, if All Those Endearing Young Charms.' {\bf \color{blue}{It was composed by former president William Trufant Foster shortly after Reed's founding, and is rarely heard today.}}  & Assuming that President Trufant Foster had been able and willing to copyright his composition, would his estate be receiving any noticeable amount of royalties from the song today? & No & Implications of Negation \\ 
 
 \bottomrule
\end{tabular}
\caption{Example questions and answers from DROP and CondaQA, showing examples of valid questions and the model capabilities they probe (discrete reasoning, and reasoning about the implications of negated statements).}
\label{tab:examples_drop_condaqa}
\end{table*}

\begin{table*}[t]
\centering
\resizebox{\textwidth}{!}{
\begin{tabular}{p{4.4cm}p{4.4cm}p{4.5cm}p{4.7cm}}
\toprule
\textbf{Original} &
\textbf{Paraphrase Edit} &
\textbf{Scope Edit} &
\textbf{Affirmative Edit} \\
\midrule
\underline{It} was composed by former president William Trufant Foster shortly after Reed's founding, and \underline{is} \textbf{rarely} \underline{heard today}. &
Shortly after Reed College was founded, former president William Trufant Foster composed \underline{the song}, which \underline{is} \textbf{seldom} \underline{heard today}. &
It was composed by former president William Trufant Foster shortly after Reed's founding, and despite its current popularity, \underline{it was} \textbf{rarely} \underline{heard in Foster's time}. &
\underline{It} was composed by former president William Trufant Foster shortly after Reed's founding, and \underline{was an immediate success}, \underline{remaining in use by the college} \underline{to this day}. \\
\bottomrule
\end{tabular}
}
\caption{Example of human-authored CondaQA edits. We omit the beginning of the paragraphs for space reasons.} 
\label{tab:condqa_edits}
\end{table*}

\begin{table*}[t]
\centering
\resizebox{\textwidth}{!}{%
\begin{tabular}{p{3.5cm}p{8.5cm}p{4.5cm}}
\toprule
\textbf{Prompting Strategy} & \textbf{Description} & \textbf{Data} \\
\midrule
Overgeneration & Generating more examples than will be retained in the final dataset & CondaQA Questions, CondaQA Edits, DROP Questions, DROP Edits \\
\midrule
Filters and Verifiers & Using LLMs or other automated methods to assess generated examples for specific quality criteria, and filter out examples that do not meet those criteria & CondaQA Questions, CondaQA Edits, DROP Questions, DROP Edits \\
\midrule
Reasoning Step Types & Prompt includes examples that capture the reasoning type in question, model is asked to generate an example that encapsulates that type & DROP Questions, DROP Edits \\
\midrule 
Human Guidelines & Use annotation instructions given to human annotators & CondaQA Questions, CondaQA Edits, DROP Questions, DROP Edits\\
\midrule
Decomposing & Breaking down the annotation task into sub-tasks an LLM can follow & CondaQA Questions, CondaQA Edits, DROP Questions, DROP Edits\\
\midrule
QA Generation & Help an LLM understand whether a piece of information changed by having it generate questions for which answers should change if the change was made & CondaQA scope edit \\
\bottomrule
\end{tabular}
}
\caption{An overview of prompting strategies used in the study (Prompting Strategy), descriptions of the prompting ideas (Description), and the datasets they were used in generating (Data).}
\label{tab:prompting-strategies}
\end{table*}

\begin{table*}[t]
\centering
\resizebox{\textwidth}{!}{%
\begin{tabular}{lp{12cm}}
\toprule
\textbf{Annotation Item} & \textbf{Validity Criteria} \\
\midrule
\textsc{CondaQA} & \\
\arrayrulecolor{black!20}\midrule
\emph{Question} & The question (i) targets the negated statement, rather than other information in the provided passage, and (ii) is about an implication of the negated statement. \\
\arrayrulecolor{black!20}\midrule
\emph{Paraphrase Edit} & The edit (i) yields a passage with the same meaning as the original passage, (ii) is coherent, and (iii) does not include the original negation cue. \\
\arrayrulecolor{black!20}\midrule
\emph{Scope Edit} & The edit (i) changes what is being negated by the negation cue, and (ii) is coherent. \\
\arrayrulecolor{black!20}\midrule
\emph{Affirmative Edit}& The edit (1) changes the passage such that what was being negated in the original passage is no longer negated, and (2) is coherent. \\
\arrayrulecolor{black}\midrule
\textsc{DROP} & \\
\arrayrulecolor{black!20}\midrule
\emph{Question} & The question (i) requires discrete reasoning over the passage and (ii) is not directly answerable from the passage. \\
\arrayrulecolor{black!20}\midrule
\emph{Compositional Edit}& The edit adds additional reasoning step(s) to the original question. \\
\arrayrulecolor{black!20}\midrule
\emph{Temporal Edit}& The edit modifies either the order in which events appear in the original passage or the dates associated with each event to swap their temporal order. \\
\arrayrulecolor{black}\bottomrule
\end{tabular}
}
\caption{Validity criteria for each annotation item.}
\label{tab:validity_criteria}
\end{table*}

\begin{table*}[t]
\centering
\resizebox{\textwidth}{!}{%
\begin{tabular}{llr}
\toprule
&  \textbf{Prompting Iteration Note} &\textbf{\% Valid} \\
\midrule
CondaQA & & \\
\quad \rotatebox[origin=c]{180}{$\Lsh$} Iteration 1 (16) & Full-bundle CondaQA instructions for people & 18.7 \\
\quad \rotatebox[origin=c]{180}{$\Lsh$} Iteration 2 (16) & Question-only CondaQA instructions for people & 25.0 \\
\quad \rotatebox[origin=c]{180}{$\Lsh$} Iteration 3 (16) & Started from scratch & 37.5\\
\quad \rotatebox[origin=c]{180}{$\Lsh$} Iteration 4 (16) & + Overgenerate \& Filter questions that don't target negation, aren't  & 68.7 \\
& \quad\quad\quad\quad\quad\quad\quad\quad about implication, \& have no valid answer type &  \\
\quad \rotatebox[origin=c]{180}{$\Lsh$} Iteration 5 (16) & + Filter questions that restate negation & 81.2 \\
\quad \rotatebox[origin=c]{180}{$\Lsh$} Iteration 6 (16) & + Add reasoning type examples to the prompt & 37.5 \\
\quad \rotatebox[origin=c]{180}{$\Lsh$} Final Evaluation $\Rightarrow$ Iteration 5
(114) & & 72.8 \\
\midrule
DROP & & \\
\quad \rotatebox[origin=c]{180}{$\Lsh$} Iteration 1 (16) & Question-only DROP instructions for people & 62.5 \\
\quad \rotatebox[origin=c]{180}{$\Lsh$} Iteration 2 (16) & Started from scratch & 25.0 \\
\quad \rotatebox[origin=c]{180}{$\Lsh$} Iteration 3 (16) & + Overgenerate \& Filter simple questions  & 31.3 \\
\quad \rotatebox[origin=c]{180}{$\Lsh$} Iteration 4 (16) & + Filter compound question and those with a wrong answer type  & 43.8 \\
\quad \rotatebox[origin=c]{180}{$\Lsh$} Iteration 5 (16) & + Add discrete reasoning question types to the prompt & 75.0 \\
\quad \rotatebox[origin=c]{180}{$\Lsh$} Iteration 6 (16) & + Filter question that are unanswerable without extra knowledge & 100.0 \\
\quad \rotatebox[origin=c]{180}{$\Lsh$} Final Evaluation $\Rightarrow$ Iteration 6 (115) & & 88.7 \\
\arrayrulecolor{black}\midrule
\end{tabular}
}
\caption{The percentage of valid questions in a sample of 16 for intermediate evaluations, and 100 or 114 for final evaluations. Questions are generated with \texttt{gpt-4-turbo-2024-04-09} and assessed by one author of this paper. %
}
\label{tab:validity_questions_intermediate}
\end{table*}
\clearpage

\begin{table*}[!ht]
\centering
\resizebox{\textwidth}{!}{%
\begin{tabular}{llr}
\toprule
\textbf{CondaQA} & \textbf{Prompting Iteration Note} &\textbf{\% Valid} \\
\midrule
\textbf{Paraphrase} & & \\ 
\quad \rotatebox[origin=c]{180}{$\Lsh$} Iteration 1 (16) & Full-bundle CondaQA instructions for people  & 75.0 \\
\quad \rotatebox[origin=c]{180}{$\Lsh$} Iteration 2 (16) & Paraphrase-only CondaQA instructions for people & 93.8 \\
\quad \rotatebox[origin=c]{180}{$\Lsh$} Iteration 3 (16) & Started from scratch & 100.0 \\
\quad \rotatebox[origin=c]{180}{$\Lsh$} Iteration 4 (16) & + Overgenerate \& Filter meaning-changing and incoherent edits & 100.0 \\
\quad \rotatebox[origin=c]{180}{$\Lsh$} Final Evaluation $\Rightarrow$ Iteration 4 (114) & & 96.5\\
\arrayrulecolor{black!20}\midrule
\textbf{Affirmative} & & \\ 
\quad \rotatebox[origin=c]{180}{$\Lsh$} Iteration 1 (16) & Full-bundle CondaQA instructions for people & 62.5 \\
\quad \rotatebox[origin=c]{180}{$\Lsh$} Iteration 2 (16) & Affirmative-only CondaQA instructions for people & 81.3\\
\quad \rotatebox[origin=c]{180}{$\Lsh$} Iteration 3 (16)& Started from scratch & 75.0 \\
\quad \rotatebox[origin=c]{180}{$\Lsh$} Iteration 4 (16) & + Overgenerate \& Filter paraphrasing and incoherent edits & 81.3 \\
\quad \rotatebox[origin=c]{180}{$\Lsh$} Iteration 5 (16)& + Identify then undo negation & 100.0 \\
\quad \rotatebox[origin=c]{180}{$\Lsh$} Final Evaluation $\Rightarrow$ Iteration 5 (114) & & 98.2 \\
\arrayrulecolor{black!20}\midrule
\textbf{Scope} & & \\ 
\quad \rotatebox[origin=c]{180}{$\Lsh$} Iteration 1 (16)& Full-bundle CondaQA instructions for people & 0.0\\
\quad \rotatebox[origin=c]{180}{$\Lsh$} Iteration 2 (16) & Scope-only CondaQA instructions for people & 25.0\\
\quad \rotatebox[origin=c]{180}{$\Lsh$} Iteration 3 (16)& Started from scratch & 50.0\\ 
\quad \rotatebox[origin=c]{180}{$\Lsh$} Iteration 4 (16)&  + Few-shot check incoherent passages & 62.5 \\
\quad \rotatebox[origin=c]{180}{$\Lsh$} Iteration 5 (16)& - Remove fewshot prompting in coherency check & 	62.5 \\
\quad \rotatebox[origin=c]{180}{$\Lsh$} Iteration 6 (16)& + Change scope via QA generation & 	75.0 \\
\quad \rotatebox[origin=c]{180}{$\Lsh$} Iteration 7 (16)& + Wording changes & 	75.0 \\
\quad \rotatebox[origin=c]{180}{$\Lsh$} Iteration 8 (16)& + Tweak task instructions & 	68.7 \\
\quad \rotatebox[origin=c]{180}{$\Lsh$} Iteration 9 (16) & - Remove coherency check & 	68.7 \\
\quad \rotatebox[origin=c]{180}{$\Lsh$} Iteration 10 (16) & + Overgenerate \& Filter failed scope change checked with QA  & 	87.5 \\
&\quad\quad\quad\quad\quad\quad\quad\hspace{1.1mm} Filter incoherent passages&\\
\quad \rotatebox[origin=c]{180}{$\Lsh$} Final Evaluation $\Rightarrow$ Iteration 10 (114) & & 	88.6 \\
\arrayrulecolor{black}\toprule
\textbf{DROP} & \textbf{Iteration Note} &\textbf{\% Valid} \\
\midrule
\textbf{Compositional} & & \\
\quad \rotatebox[origin=c]{180}{$\Lsh$} Iteration 1 (16)& Started from scratch &	12.5 \\
\quad \rotatebox[origin=c]{180}{$\Lsh$} Iteration 2 (16)& + Overgenerate \& Filter edits that do not add steps & 43.7 \\
& \quad\quad\quad\quad\quad\quad\quad\hspace{1mm} Filter compound or incoherent edit-questions  &	\\
\quad \rotatebox[origin=c]{180}{$\Lsh$} Iteration 3 (16)& + Wording changes &	43.7 \\
\quad \rotatebox[origin=c]{180}{$\Lsh$} Iteration 4 (16)& + Include reasoning step types in the prompt \&  &	81.2 \\
& \quad improve the number of reasoning steps filter & \\
\quad \rotatebox[origin=c]{180}{$\Lsh$} Final Evaluation $\Rightarrow$ Iteration 4 (100)	& & 65.0\\
\arrayrulecolor{black!20}\midrule
 \textbf{Temporal} & & \\
 \quad \rotatebox[origin=c]{180}{$\Lsh$} Iteration 1 (16) & Started from scratch & 87.5\\
 & + Overgenerate \& Filter incoherent edits + equal event order edits &  \\
 \quad \rotatebox[origin=c]{180}{$\Lsh$} Final Evaluation $\Rightarrow$ Iteration  (100) & & 86.0  \\
\arrayrulecolor{black}\bottomrule
\end{tabular}
}
\caption{The percentage of valid edits in a sample of 16 for intermediate evaluations, and 100 or 114 for final evaluations. 
Edits are generated with \texttt{gpt-4-turbo-2024-04-09} and assessed by one author of this paper. %
}
\label{tab:edit_validity_with_intermediate}
\end{table*}

\begin{table*}[htbp]
\centering
\resizebox{\textwidth}{!}{
\small
\begin{tabular}{p{1cm}p{10cm}p{2.5cm}p{2cm}}
\toprule
\textbf{Dataset} & \textbf{Passage} & \textbf{Human-Authored Question} & \textbf{LLM-Generated Question} \\
\midrule
DROP & \parbox[t]{10cm}{To open the regular season, the Lions hosted the St. Louis Rams. The Rams took an early lead when Greg Zuerlein kicked a 48-yard field goal, the only scoring drive of the first quarter. The Lions responded in the second quarter when Joique Bell ran in a touchdown from 1 yard out. St. Louis made it a 1-point game when Greg Zuerlein kicked a 29-yard field goal, and took the lead again when Cortland Finnegan intercepted a Kevin Smith pass and ran it back 31 yards for a touchdown. The final points of the half were scored when Detroit's Jason Hanson kicked a 41-yard field goal. The only points of the third quarter was a 45-yard field goal by Jason Hanson to tie the game. In the final quarter, the score continued to sea-saw. First the Rams' Brandon Gibson caught a 23-yard touchdown pass from Sam Bradford to take the lead back again, but the Lions soon tied it back up when Kevin Smith ran in a touchdown from 5 yards out. The Rams then went up by 3 points when kicked Greg Zuerlein his third field goal of the game, this one from 46 yards out. In the final 2 minutes, the Lions completed their third 80-yard scoring drive with a game-winning 5-yard touchdown catch by Kevin Smith, his second of the game. Greg Zuerlein's third field was the last of the game. The Lions were able to start 1-0 and capture their second straight regular season opening win.} & How many more yards was the last touchdown than the first? & How many touchdowns did Kevin Smith score in total during this game? \\
\midrule
DROP & \parbox[t]{10cm}{Hoping to rebound from their road loss to the Jaguars, the Chargers went home for a Week 12 duel with the Baltimore Ravens. After a scoreless first quarter, San Diego struck first with kicker Nate Kaeding getting a 27-yard field goal. Afterwards, the Ravens took the lead with RB Willis McGahee getting a 1-yard TD run. Fortunately, the Chargers regained the lead with QB Philip Rivers completing a 35-yard TD pass to TE Antonio Gates (with a failed PAT), Kaeding kicking a 46-yard field goal, Rivers completing a 5-yard TD pass to WR Chris Chambers, and Kaeding kicking a 41-yard field goal. In the third quarter, San Diego increased its lead with Rivers and Gates hooking up with each other again on a 25-yard TD pass. Baltimore would manage to one final score as Ravens QB Kyle Boller completed a 13-yard TD pass to FB Le'Ron McClain. Fortunately, in the fourth quarter, the 'Bolts sealed the win with Kaeding nailing a 41-yard field goal. RB LaDainian Tomlinson (24 carries for 77 yards) became the fourth-fastest player and the 23rd player in NFL history to get 10,000 career rushing yards.} & How many yards did Rivers complete? & What was the total distance of all field goals made by Nate Kaeding in the game? \\
\midrule
DROP & \parbox[t]{10cm}{Hoping to rebound from their road loss to the Jaguars, the Chargers went home for a Week 12 duel with the Baltimore Ravens. After a scoreless first quarter, San Diego struck first with kicker Nate Kaeding getting a 27-yard field goal. Afterwards, the Ravens took the lead with RB Willis McGahee getting a 1-yard TD run. Fortunately, the Chargers regained the lead with QB Philip Rivers completing a 35-yard TD pass to TE Antonio Gates (with a failed PAT), Kaeding kicking a 46-yard field goal, Rivers completing a 5-yard TD pass to WR Chris Chambers, and Kaeding kicking a 41-yard field goal. In the third quarter, San Diego increased its lead with Rivers and Gates hooking up with each other again on a 25-yard TD pass. Baltimore would manage to one final score as Ravens QB Kyle Boller completed a 13-yard TD pass to FB Le'Ron McClain. Fortunately, in the fourth quarter, the 'Bolts sealed the win with Kaeding nailing a 41-yard field goal. RB LaDainian Tomlinson (24 carries for 77 yards) became the fourth-fastest player and the 23rd player in NFL history to get 10,000 career rushing yards.} & How many yards was the longest field goal of the game? & How many more yards did Nate Kaeding's second field goal in the game exceed his first field goal by? \\
\bottomrule 
\end{tabular}
}
\caption{Examples of Human-Authored and LLM-Generated Questions for the DROP Dataset.}
\label{tab:dataset-questions_drop}
\end{table*}

\begin{table*}[htbp]
\centering
\resizebox{\textwidth}{!}{
\small
\begin{tabular}{p{8cm}p{4cm}p{2.75cm}}
\toprule
\textbf{Passage} & \textbf{Human-Authored Question} & \textbf{LLM-Generated Question} \\
\midrule
The south celestial pole is visible only from the Southern Hemisphere. It lies in the dim constellation Octans, the Octant. \begin{bf}Sigma Octantis is identified as the south pole star, more than one degree away from the pole, but with a magnitude of 5.5 it is {\color{blue}barely} visible on a clear night.\end{bf} & Would inexperienced sailors be able to easily use Sigma Octantis to determine their direction of travel within one compass degree of accuracy on a cloudless night? & Would Sigma Octantis be considered a reliable star for navigators to use under cloudy skies or in light-polluted areas? \\
\midrule
It is difficult to distinguish an extinct volcano from a dormant (inactive) one. Dormant volcanoes are those that have not erupted for thousands of years, but are likely to erupt again in the future. Volcanoes are often considered to be extinct if there are no written records of its activity.,. \begin{bf}Before its catastrophic eruption of 1991, Pinatubo was an {\color{blue}inconspicuous} volcano, unknown to most people in the surrounding areas.\end{bf} Two other examples are [...] & From the way Pinatubo is described in the passage, is it reasonable to assume that it was of fairly modest height, and/or lacking in the usual cone shape that is commonly associated with volcanoes? & Would Pinatubo likely have been a main point of interest on a geological tour before its eruption in 1991? \\
\midrule
 Much like elephants, manatees are polyphyodonts, and continuously replace their teeth from the back of the jaw. \begin{bf}Adults {\color{blue}lack} incisors, canines, and premolars, and instead have 8 to 10 cheek teeth in their mouth.\end{bf} Manatees have an infinite supply of teeth moving in from the back and shedding in the front [...] & If the skull of an unknown animal is discovered to have premolars, is it possible that the skull belongs to an adult manatee? & If someone discovered a canine tooth, could it belong to an adult manatee? \\
\bottomrule
\end{tabular}
}
\caption{Examples of human-authored and LLM-generated questions for the \textbf{CondaQA} Dataset. The negated statement is highlighted in {\bf bold} and the negation is highlighted in {\color{blue} blue}.}
\label{tab:dataset-questions_condaqa}
\end{table*}
\clearpage

\begin{table*}[htbp]
\centering
\resizebox{\textwidth}{!}{
\small
\begin{tabular}{p{1.75cm}p{7cm}p{2.5cm}p{2.5cm}p{4cm}}
\toprule
\textbf{Validity \newline Assessment} & \textbf{Passage} & \textbf{Original Question} & \textbf{LLM-Generated Compositional Edit} & \textbf{Author Note}\\
\midrule
Mutually Valid & \parbox[t]{7cm}{The U.S. finance industry comprised only 10\% of total non-farm business profits in 1947, but it grew to 50\% by 2010. Over the same period, finance industry income as a proportion of GDP rose from 2.5\% to 7.5\%, and the finance industrys proportion of all corporate income rose from 10\% to 20\%. The mean earnings per employee hour in finance relative to all other sectors has closely mirrored the share of total U.S. income earned by the top 1\% income earners since 1930. The mean salary in New York Citys finance industry rose from \$80,000 in 1981 to \$360,000 in 2011, while average New York City salaries rose from \$40,000 to \$70,000. In 1988, there were about 12,500 U.S. banks with less than \$300 million in deposits, and about 900 with more deposits, but by 2012, there were only 4,200 banks with less than \$300 million in deposits in the U.S., and over 1,800 with more.} & How many more dollars was the average New York salaries in 2011 compared to average New York salaries in 1981? & What is the difference between the increase in the mean salary of New York City's finance sector and the average salary increase in New York City from 1981 to 2011? & The edit adds a reasoning step of calculating the mean salary increase of the finance sector and subtracting the average salary increase which was the required reasoning step in the original question.
\\
\midrule
Mutually \newline Invalid & \parbox[t]{7cm}{The Steelers traveled to Kansas City to take on the Chiefs. In the first quarter, the Chiefs scored first when Cairo Santos kicked a 30-yard field goal for a 3-0 lead. The Steelers would tie it up later on as Chris Boswell nailed a field goal from 24 yards out to make the game 3-3. In the second quarter, it was all Chiefs when Santos nailed 2 more field goals: from 27 and 22 yards out moving ahead 6-3 and the eventual halftime score of 9-3. After the break, the Chiefs moved ahead by double digits when Chircandrick West ran for a 1-yard touchdown for a 16-3 game. Backup QB Landry Jones was able to find Martavis Bryant on a 19-yard pass to make the score 16-10 for the Steelers later on in the third quarter. In the fourth quarter, the Steelers came within 3 when Boswell nailed another field goal from 36-yards out for a 16-13 game. However, the Chiefs eventually sealed the win when Alex Smith found Chris Conley on a 6-yard pass for a final score of 23-13. With the loss, the Steelers fell to 4-3 and also 2-2 without Roethlisberger as the starter. The defense failed to force any takeaways, but got inside the pocket and sacked Smith twice.
} & How many field goals did Cairo Santos make?
& How many field goals did Cairo Santos make if the total number of field goals in the game was 5?
& The edit adds in extra information but it does not require an additional reasoning step to answer the question. The reasoning steps are the same as the original question. \\
\midrule

Author \newline Valid \begin{center}
    +
\end{center} Annotator \newline Invalid & \parbox[t]{7cm}{A 2016 survey conducted by the Razumkov Centre found that 70\% of the population declared themselves believers in any religion, while 6.3\% declared themselves non-believers, and 2.7\% declared to be atheists. Of the total Ukraian population, 81.9\% declared to be Christians, comprising a 63.4\% who declared to be Eastern Orthodox Church, 8.5\% Ukrainian Greek Catholic Church, 7.1\% simply christians, 1.9\% Protestantism, and 1.0\ Catholic Church. Islam comprises 1.1\% of the population, while Judaism and Hinduism were the religions of 0.2\% of the population each. A further 16.3\% of the population believed in some other religion not identifying in one of those listed hitherto; it may comprise Rodnovery and other faiths.
} & How many percent of the total Ukraian population were not declared Christians?
 & What is the sum of the percentage of the total Ukrainian population that were declared non-Christians and those who declared themselves non-believers or atheists?
& The edit, while adding reasoning steps, is no longer answerable. Since the percent of the population that are non-believers or atheists has some unknown overlap with the percent of the population that is non-christian, it is impossible to calculate with certainty the sum which the edit asks for. Annotators may have just noted the additional reasoning step and marked the edit as valid. \\
\bottomrule 
\end{tabular}
}
\caption{Examples of LLM-generated compositional edits for DROP with various validity assessments from annotators and one of the authors of this paper. Compositional edits must add an additional reasoning step to the original question. The answer to a valid compositional edit must be either a span of text from the question or passage, a date, or a number.%
}
\label{tab:compositional-agreements-drop}
\end{table*}

\begin{table*}[htbp]
\centering
\resizebox{\textwidth}{!}{
\small
\begin{tabular}{p{1.75cm}p{10cm}p{2.5cm}p{4cm}}
\toprule
\textbf{Validity \newline Assessment} & \textbf{Passage} & \textbf{LLM-Generated Question} & \textbf{Author Note}\\
\midrule
Mutually Valid & \parbox[t]{10cm}{Coming off their impressive road win over the 49ers, the Falcons went home for a Week 6 Sunday night duel with the Chicago Bears.  After a scoreless first quarter, Atlanta would trail early in the second quarter as Bears quarterback Jay Cutler found wide receiver Johnny Knox on a 23-yard touchdown pass.  Afterwards, the Falcons took the lead as quarterback Matt Ryan completed a 40-yard touchdown pass to wide receiver Roddy White and a 10-yard touchdown pass to tight end Tony Gonzalez. After a scoreless third quarter, Chicago would tie the game in the fourth quarter with Cutler hooking up with tight end Greg Olsen on a 2-yard touchdown.  Atlanta would regain the lead as running back Michael Turner got a 5-yard touchdown run.  Afterwards, the defense would fend off a last-second Bears drive to lock up the victory.
} & By how many yards was Matt Ryan's touchdown pass to Roddy White longer than his touchdown pass to Tony Gonzalez?
 & The question is answerable numerically, requires complex reasoning (subtraction), and requires identifying multiple parts of the passage to answer. Since it meets all the requirements distinctly, this question is easily categorized as valid.
\\
\midrule
Mutually Invalid & \parbox[t]{10cm}{Concerned by the Burmese consolidation of Shan states and Laotian states, China launched four invasions of Burma between 1765 and 1769. At first, the Qianlong Emperor envisaged an easy war, and sent in only the Green Standard troops stationed in Yunnan. The Qing invasion came as the majority of Burmese forces were deployed in their latest invasion of Siam. Nonetheless, battle-hardened Burmese troops defeated the first two invasions of 1765-1766 and 1766-1767 at the border. The regional conflict now escalated to a major war that involved military manoeuvres nationwide in both countries. The third invasion  led by the elite Manchu Bannermen nearly succeeded, penetrating deep into central Burma within a few days' march from Ava. But the Bannermen of northern China could not cope with unfamiliar tropical terrains and lethal endemic diseases, and were driven back with heavy losses. After the close-call, Hsinbyushin redeployed his armies from Siam to the Chinese front. The fourth and largest invasion got bogged down at the frontier. With the Qing forces completely encircled, a truce was reached between the field commanders of the two sides in December 1769.
} & How many different invasions of Burma did China launch between 1765 and 1769?
& This question is answered directly in the original passage in a single span \begin{it}
    "China launched four invasions of Burma between 1765 and 1769."
\end{it} so it does not meet the requirements of a DROP question and is invalid. \\
\midrule

Author \newline Invalid \begin{center}
    +
\end{center} Annotator \newline Valid & \parbox[t]{10cm}{Quarrels between Denmark and Sweden led to the Northern Seven Years' War in 1563, which ended in 1570 with the Treaty of Stettin. Primarily fought in western and southern Scandinavia, the war involved important naval battles fought in the Baltic. When Danish-held Varberg surrendered to Swedish forces in 1565, 150 Danish mercenaries escaped the subsequent massacre of the garrison by defecting to Sweden. Among them was Pontus de la Gardie, who thereafter became an important Swedish commander in the Livonian War. Livonia was also affected by the naval campaign of Danish admiral Peter or Per Munck, who bombarded Swedish Reval from sea in July 1569. The Treaty of Stettin made Denmark the supreme and dominating power in Northern Europe, yet failed to restore the Kalmar Union. Unfavourable conditions for Sweden led to a series of conflicts that only ended with the Great Northern War in 1720. Sweden agreed to turn over her possessions in Livonia in return for a payment by Holy Roman Emperor Maximilian II. Maximilian failed to pay the promised compensation, however, and thereby lost his influence on Baltic affairs. The terms of the treaty regarding Livonia were ignored, and thus the Livonian War continued. From Ivan's point of view, the treaty enabled the powers involved to form an alliance against him, now that they were no longer fighting each other.
} & How many years was it between the end of the Northern Seven Years' War and the beginning of the Great Northern War?
& The authors interpretation of the information  \begin{it}only ended with the Great Northern War in 1720\end{it} is that the beginning year of the Great Northern War is not explicitly mentioned, so the question is too ambiguous to be answered with information from the passage. Annotators likely interpreted the same quote as referring to the year the war began, in which case the question would be valid.\\
\midrule

Author \newline Valid \begin{center}
    +
\end{center} Annotator \newline Invalid & \parbox[t]{10cm}{Coming off their season-sweeping home win over the 49ers, the Seahawks stayed at home for a Week 11 duel against the Chicago Bears, in the rematch of last year's NFC Divisional game (previously in Chicago). In the first quarter, Seattle trailed early as Bears RB Cedric Benson got a 43-yard TD run, along with kicker Robbie Gould getting a 31-yard field goal.  The Seahawks would get on the board with QB Matt Hasselbeck completing a 19-yard TD pass to WR D. J. Hackett.  In the second quarter, the Seahawks took the lead with RB Maurice Morris getting a 19-yard TD run.  However, Chicago regained the lead with RB Adrian Peterson getting a 5-yard TD run.  Seattle would tie the game kicker Josh Brown getting a 40-yard field goal. In the third quarter, the Seahwaks retook the lead as Hasselbeck completed a 4-yard TD pass to WR Nate Burleson for the only score of the period.  In the fourth quarter, the Bears tried to retaliate as Gould kicked a 47-yard field goal.  Afterwards, Seattle pulled away with Brown kicking a 23-yard and a 46-yard field goal.  Chicago's final response would be Gould nailing a 48-yard field goal.
} & How many yards were the field goals that Josh Brown made in the fourth quarter?

& An author of this paper marked the question as valid as it can be answered by summing up the two field goals from Brown. Annotators likely marked it as invalid as the question does not explicitly ask for the total yards of the field goals, so the question could be interpreted as asking for the yards of each field goal individually.\\
\bottomrule 
\end{tabular}
}
\caption{Examples of LLM-Generated Questions for DROP with various validity assessments. Questions must involve complex reasoning and looking at more than one part of the passage to answer. The answer to a valid DROP question must be either a span of text from the question or passage, a date, or a number.}
\label{tab:question-agreements-drop}
\end{table*}

\begin{table*}[htbp]
\centering
\resizebox{\textwidth}{!}{
\small
\begin{tabular}{p{1.75cm}p{7cm}p{2.5cm}p{5cm}}
\toprule
\textbf{Validity \newline Assessment} & \textbf{Passage}  & \textbf{LLM-Generated Question} & \textbf{Author Note}\\
\midrule
Mutually Valid & \parbox[t]{7cm}{The south celestial pole is visible only from the Southern Hemisphere. It lies in the dim constellation Octans, the Octant. \begin{bf}Sigma Octantis is identified as the south pole star, more than one degree away from the pole, but with a magnitude of 5.5 it is {\color{blue}barely} visible on a clear night\end{bf}.
} & Would Sigma Octantis be considered a reliable star for navigators to use under cloudy skies or in light-polluted areas?
 & The question targets the negation that the star is barely visible, and asks about an implication of the star being barely visible.
\\
\midrule
Mutually \newline Invalid & \parbox[t]{7cm}{During the 1980s and early 1990s, Murdoch's publications were generally supportive of Britain's Prime Minister Margaret Thatcher. At the end of the Thatcher/Major era, Murdoch switched his support to the Labour Party and its leader, Tony Blair. The closeness of his relationship with Blair and their secret meetings to discuss national policies was to become a political issue in Britain. This later changed, with "The Sun", in its English editions, publicly renouncing the ruling Labour government and lending its support to David Cameron's Conservative Party, which soon afterwards formed a coalition government. \begin{bf}In Scotland, where the Conservatives had suffered a complete annihilation in 1997, the paper began to endorse the Scottish National Party (though {\color{blue}not} yet its flagship policy of independence), which soon after came to form the first ever outright majority in the proportionally elected Scottish Parliament.\end{bf} Former Prime Minister Gordon Brown's official spokesman said in November 2009 that Brown and Murdoch "were in regular communication" and that "there is nothing unusual in the prime minister talking to Rupert Murdoch".} & Did "The Sun" initially support the SNP\'s stance on independence when they first endorsed the party?
& The question targets the negation, but not an implication of the negation. It asks for a direct factual interpretation of the negation, so the question is invalid.\\
\midrule

Author \newline Invalid \begin{center}
    +
\end{center} Annotator \newline Valid & \parbox[t]{7cm}{A Dominican friar, Orsini focused on his religious responsibilities as bishop rather than on papal administration. \begin{bf}Orsini's lack of political expertise led him to increasingly rely on an {\color{blue}unscrupulous} secretary (Cardinal Niccolò Coscia) whose financial abuses ruined the papal treasury, causing great damage to the Church in Rome.\end{bf}} & Did Orsini’s choice of a close associate contribute positively to his leadership effectiveness?
 & One author concluded that this question did not target the implications of the negated statement of Niccolò Coscia being unscrupulous directly. His being unscrupulous is not directly attributable to Orsini's leadership effectiveness. Orsini could have been extremely effective despite Niccolò's unscrupulous behavior. Annotators likely felt that the direct connection did exist, and therefore rated the question as valid.\\
 \midrule

Author \newline Valid \begin{center}
    +
\end{center} Annotator \newline Invalid & \parbox[t]{7cm}{The walls and sanctum's door frame of the both levels are intricately carved. These show themes such as "makaras" and other mythical creatures, apsaras, elephants in different stages of activity, females in waving or welcoming gesture. The upper level of Cave 6 is significant in that it shows a devotee in a kneeling posture at the Buddha's feet, an indication of devotional worship practices by the 5th century. The colossal Buddha of the shrine has an elaborate throne back, but was hastily finished in 477/478 CE, when king Harisena died. \begin{bf}The shrine antechamber of the cave features an {\color{blue}unfinished} sculptural group of the Six Buddhas of the Past, of which only five statues were carved.\end{bf} This idea may have been influenced from those in Bagh Caves of Madhya Pradesh.
} & Is there more than one fully completed sculptural group in the antechamber of the cave?

& An author concluded that this question was valid as the negation implies what the answer to this question would be. The answer should be "don't know" here because despite the finished or unfinished state of the sculptural group, there could potentially be more sculptural groups in the antechamber. Annotators marked the question invalid, either due to assuming that an answer of "don't know" made the question invalid, or deeming the connection between the implication of the negation and the question too weak.\\
\bottomrule 
\end{tabular}
}
\caption{Examples of LLM-Generated Questions for CondaQA with various validity assessments. Questions must be about an implication of the negated statement. The negated statement is highlighted in {\bf bold} and the negation is highlighted in {\color{blue} blue}.}
\label{tab:question-agreements-condaqa}
\end{table*}

\begin{table*}[htbp]
\centering
\resizebox{\textwidth}{!}{
\small
\begin{tabular}{p{1.75cm}p{7cm}p{7cm}p{4cm}}
\toprule
\textbf{Validity \newline Assessment} & \textbf{Passage} & \textbf{LLM-Generated Paraphrase Edit} & \textbf{Author Note}\\
\midrule
Mutually Valid & \parbox[t]{7cm}{The school song, "Fair Reed," is sung to the tune of the 1912 popular song "Believe Me, if All Those Endearing Young Charms." It may be imitative of the Harvard anthem "Fair Harvard," which is also sung to the tune of "Believe Me, if All Those Endearing Young Charms." \begin{bf}It was composed by former president William Trufant Foster shortly after Reed's founding, and is {\color{blue}rarely} heard today.\end{bf}
} & \parbox[t]{7cm}{The school song, "Fair Reed," is sung to the tune of the 1912 popular song "Believe Me, if All Those Endearing Young Charms." It may be imitative of the Harvard anthem "Fair Harvard," which is also sung to the tune of "Believe Me, if All Those Endearing Young Charms." \begin{bf}Former president William Trufant Foster composed it shortly after Reed's founding, and it is seldom heard today.\end{bf}}
 & The edited passage has the same meaning as the original, and the negation rarely is no longer present, so the edit is valid.
\\
\midrule
Mutually \newline Invalid & \parbox[t]{7cm}{Most of what is known about Anthony comes from the "Life of Anthony". \begin{bf}Written in Greek around 360 by Athanasius of Alexandria, it depicts Anthony as an {\color{blue}illiterate} and holy man who through his existence in a primordial landscape has an absolute connection to the divine truth, which always is in harmony with that of Athanasius as the biographer.
\end{bf}
} & \parbox[t]{7cm}{Most of what is known about Anthony comes from the "Life of Anthony". \begin{bf}Written in Greek around 360 by Athanasius of Alexandria, it portrays Anthony as an uneducated yet holy man who, through his life in a primordial landscape, achieves a total connection to the divine truth, which is invariably aligned with that presented by Athanasius as the biographer.\end{bf}} & Although the negation is no longer present, the edited passage does not retain the same meaning as the original. \begin{it}Uneducated\end{it} does not imply the same meaning as \begin{it} illiterate \end{it}. Therefore the edit is invalid.
 \\
\midrule

Author \newline Invalid \begin{center}
    +
\end{center} Annotator \newline Valid & \parbox[t]{7cm}{In 2008, Narconon executive Marie Cecchini published, with Richard Lennox, a paper that claimed to show that the Narconon educational program reduced drug use among youths. \begin{bf}However, the study was funded by Narconon's parent organisation, ABLE, and subsequent correspondence in the same journal asserted that the study's conclusions were contradicted by its own data: that the control group "were more likely to {\color{blue}resist} pressures to take drugs" than the Narconon group.\end{bf}
} & \parbox[t]{7cm}{In 2008, Narconon executive Marie Cecchini published, with Richard Lennox, a paper that claimed to show that the Narconon educational program reduced drug use among youths. \begin{bf}However, the study was funded by Narconon's parent organisation, ABLE, and subsequent correspondence in the same journal asserted that the study's conclusions were contradicted by its own data: the control group "were more likely to yield to pressures to take drugs" less than the Narconon group.\end{bf}
}
 & While the edited passage does remove the negation, the resulting passage is incoherent. \begin{it}"were more likely to yield to pressure to take drugs" less than the Narconon group \end{it} doesn't make any sense, since the presence of "more" implies that there is another group the control group is more likely than to yield to pressure to take drugs. If the edit had been \begin{it}"were less likely to yield to pressure to take drugs"\end{it}, then the edit would be valid. Annotators likely just saw that the double negation preserved the meaning of the original passage, and interpreted the sentence as coherent.\\
\midrule

Author \newline Valid \begin{center}
    +
\end{center} Annotator \newline Invalid & \parbox[t]{7cm}{In the early 20th century and into the Mexican Revolution, the production of coffee was particularly important but labor-intensive. This would lead to a practice called "enganche" (hook), where recruiters would lure workers with advanced pay and other incentives such as alcohol and then trap them with debts for travel and other items to be worked off. \begin{bf}This practice would lead to a kind of indentured servitude and uprisings in areas of the state, although they {\color{blue}never} led to large rebel armies as in other parts of Mexico.\end{bf}
} & \parbox[t]{7cm}{ In the early 20th century and into the Mexican Revolution, the production of coffee was particularly important but labor-intensive. This would lead to a practice called "enganche" (hook), where recruiters would lure workers with advanced pay and other incentives such as alcohol and then trap them with debts for travel and other items to be worked off. \begin{bf}This practice would lead to a kind of indentured servitude and uprisings in areas of the state, albeit without resulting in the formation of large rebel armies as seen in other parts of Mexico.\end{bf}}
 & An author marked this edit as valid as it retained the meaning of the original passage (that the practice did not lead to the formation of large rebel armies). Annotators marked it as invalid, possibly because the exact meaning of the original passage is slightly changed in the paraphrase edit.
 \\
\bottomrule 
\end{tabular}
}
\caption{Examples of LLM-Generated Paraphrase Edits for CondaQA with various validity assessments. Paraphrase edits must be a rewrite of the original negated statement such that the new passage has the same meaning as the original, but the negation is no longer included. The negated statement is highlighted in {\bf bold} and the negation is highlighted in {\color{blue} blue}.}
\label{tab:paraphrase-agreements-condaqa}
\end{table*}

\begin{table*}[htbp]
\centering
\resizebox{\textwidth}{!}{
\small
\begin{tabular}{p{1.75cm}p{7cm}p{7cm}p{4cm}}
\toprule
\textbf{Validity \newline Assessment} & \textbf{Passage} & \textbf{LLM-Generated Scope Edit} & \textbf{Author Note}\\
\midrule
Mutual Valid & \parbox[t]{7cm}{Much like elephants, manatees are polyphyodonts, and continuously replace their teeth from the back of the jaw. \begin{bf}Adults {\color{blue}lack} incisors, canines, and premolars, and instead have 8 to 10 cheek teeth in their mouth.\end{bf} Manatees have an infinite supply of teeth moving in from the back and shedding in the front, which are continuously formed by a dental capsule behind the tooth-row. These teeth are constantly worn down by the abrasive vascular plants they forage, particularly aquatic grasses. Unlike in manatees, the dugong's teeth do not continually grow back via horizontal tooth replacement. The dugong has two tusks which emerge in males during puberty, and sometime later in life for females after reaching the base of the premaxilla. The number of growth layer groups in a tusk indicates the age of a dugong.
} & \parbox[t]{7cm}{Much like elephants, manatees are polyphyodonts, and continuously replace their teeth from the back of the jaw. \begin{bf}Adults {\color{blue}lack} molars and instead have 8 to 10 incisors, canines, and premolars in their mouth.\end{bf} Manatees have an infinite supply of teeth moving in from the back and shedding in the front, which are continuously formed by a dental capsule behind the tooth-row. These teeth are constantly worn down by the abrasive vascular plants they forage, particularly aquatic grasses. Unlike in manatees, the dugong's teeth do not continually grow back via horizontal tooth replacement. The dugong has two tusks which emerge in males during puberty, and sometime later in life for females after reaching the base of the premaxilla. The number of growth layer groups in a tusk indicates the age of a dugong.}
 & The edit changes what is being negated by lack from the incisors, canines and premolars to the molars, so the scope edit is valid.
\\
\midrule
Mutually \newline Invalid & \parbox[t]{7cm}{It is difficult to distinguish an extinct volcano from a dormant (inactive) one. Dormant volcanoes are those that have not erupted for thousands of years, but are likely to erupt again in the future. Volcanoes are often considered to be extinct if there are no written records of its activity. Nevertheless, volcanoes may remain dormant for a long period of time. For example, Yellowstone has a repose/recharge period of around 700,000 years, and Toba of around 380,000 years. Vesuvius was described by Roman writers as having been covered with gardens and vineyards before its eruption of 79 CE, which destroyed the towns of Herculaneum and Pompeii. \begin{bf}Before its catastrophic eruption of 1991, Pinatubo was an {\color{blue}inconspicuous} volcano, unknown to most people in the surrounding areas. \end{bf}Two other examples are the long-dormant Soufrière Hills volcano on the island of Montserrat, thought to be extinct before activity resumed in 1995, and Fourpeaked Mountain in Alaska, which, before its September 2006 eruption, had not erupted since before 8000 BCE and had long been thought to be extinct.
} & \parbox[t]{7cm}{It is difficult to distinguish an extinct volcano from a dormant (inactive) one. Dormant volcanoes are those that have not erupted for thousands of years, but are likely to erupt again in the future. Volcanoes are often considered to be extinct if there are no written records of its activity. Nevertheless, volcanoes may remain dormant for a long period of time. For example, Yellowstone has a repose/recharge period of around 700,000 years, and Toba of around 380,000 years. Vesuvius was described by Roman writers as having been covered with gardens and vineyards before its eruption of 79 CE, which destroyed the towns of Herculaneum and Pompeii. \begin{bf}Before its catastrophic eruption of 1991, Pinatubo was an {\color{blue}inconspicuous} volcano, except to a few local geologists who had studied its potential for activity. \end{bf}Two other examples are the long-dormant Soufrière Hills volcano on the island of Montserrat, thought to be extinct before activity resumed in 1995, and Fourpeaked Mountain in Alaska, which, before its September 2006 eruption, had not erupted since before 8000 BCE and had long been thought to be extinct.}
& The negation in the edit is still negating the Pinatubo volcano. As the subject of the negation has not changed, the scope edit is therefore invalid.
 \\
\midrule

Author \newline Invalid \begin{center}
    +
\end{center} Annotator \newline Valid & \parbox[t]{7cm}{Hearts is an "evasion-type" trick-taking playing card game for four players, although most variations can accommodate between three and six players. It was first recorded in America in the 1880s and has many variants, some of which are also referred to as "Hearts"; especially the games of Black Lady and Black Maria. \begin{bf}The game is a member of the Whist group of trick-taking games (which also includes Bridge and Spades), but is {\color{blue}unusual} among Whist variants in that it is a trick-avoidance game; players avoid winning certain penalty cards in tricks, usually by avoiding winning tricks altogether.\end{bf} The original game of Hearts is still current, but has been overtaken in popularity by Black Lady in the United States and Black Maria in Great Britain.
} & \parbox[t]{7cm}{ Hearts is an "evasion-type" trick-taking playing card game for four players, although most variations can accommodate between three and six players. It was first recorded in America in the 1880s and has many variants, some of which are also referred to as "Hearts"; especially the games of Black Lady and Black Maria. \begin{bf}The game is a member of the Whist group of trick-taking games (which also includes Bridge and Spades), but is {\color{blue}unusual} among Whist variants in that it involves using a unique scoring system where points are deducted for each trick won, encouraging players to win as few tricks as possible.\end{bf} The original game of Hearts is still current, but has been overtaken in popularity by Black Lady in the United States and Black Maria in Great Britain.
}
 & An author interpreted the subject of negation in both the original and edited passages as Hearts being different from other Whist variants. The scope edit changes why Hearts is different from other Whist variants, but not the subject of what is unusual, which remains that Hearts is different from other Whist variants. An annotator likely interpreted the change in why Hearts is different as a sufficient change in scope of the negation.\\
\midrule

Author \newline Valid \begin{center}
    +
\end{center} Annotator \newline Invalid & \parbox[t]{7cm}{Gwen Stacy is recast as Paige Embry in "The Refrigerator Monologues". She is the unofficial leader of the Hell Hath Club, a group of women in the afterlife trying to cope with the brutal termination of their plot lines, and provides connecting narration for each of their stories. \begin{bf}She loved wearing her lab coat and feels {\color{blue}unhappy} about being stuck in the clothes she was buried in.
\end{bf}
} & \parbox[t]{7cm}{ Gwen Stacy is recast as Paige Embry in "The Refrigerator Monologues". She is the unofficial leader of the Hell Hath Club, a group of women in the afterlife trying to cope with the brutal termination of their plot lines, and provides connecting narration for each of their stories. \begin{bf}She loved wearing her lab coat and feels {\color{blue}unhappy} about not being able to wear it anymore, as she is stuck in the clothes she was buried in.\end{bf}
}
 & The difference here is subtle, but the subject of the negation unhappy is on being stuck in the clothes she was wearing in the original passage, and is on not being able to wear her lab coat anymore in the scope edit. An author deemed this change in subject as sufficient. An annotator likely did not agree that the change in scope was different enough for the edit to be valid.
 \\
\bottomrule 
\end{tabular}
}
\caption{Examples of LLM-Generated Scope Edits for CondaQA with various validity assessments. Scope edits must be a rewrite of the original passage such that what is being negated by the negation is changed. The negated statement is highlighted in {\bf bold} and the negation is highlighted in {\color{blue} blue}.}
\label{tab:scope-agreements-condaqa}
\end{table*}

\begin{table*}[htbp]
\centering
\resizebox{\textwidth}{!}{
\small
\begin{tabular}{p{1.75cm}p{7cm}p{7cm}p{4cm}}
\toprule
\textbf{Validity \newline Assessment} & \textbf{Passage} & \textbf{LLM-Generated Affirmative Edit} & \textbf{Author Note}\\
\midrule
Mutual Valid & \parbox[t]{7cm}{Johann Spurzheim was Gall's most important collaborator. \begin{bf}He worked as Gall's anatomist until 1813 when for {\color{blue}unknown} reasons they had a permanent falling out.\end{bf} Publishing under his own name Spurzheim successfully disseminated phrenology throughout the United Kingdom during his lecture tours through 1814 and 1815 and the United States in 1832 where he would eventually die.
} & \parbox[t]{7cm}{Johann Spurzheim was Gall's most important collaborator. \begin{bf}He worked as Gall's anatomist until 1813 when, due to professional disagreements over the methods and applications of phrenology, they had a permanent falling out.\end{bf} Publishing under his own name, Spurzheim successfully disseminated phrenology throughout the United Kingdom during his lecture tours through 1814 and 1815 and the United States in 1832, where he would eventually die.}
 & This edit is valid since the reasons for the falling out are no longer unknown in the Affirmative edit, and the new passage is coherent and consistent.
\\
\midrule

Author \newline Invalid \begin{center}
    +
\end{center} Annotator \newline Valid & \parbox[t]{7cm}{In 2008, Narconon executive Marie Cecchini published, with Richard Lennox, a paper that claimed to show that the Narconon educational program reduced drug use among youths. \begin{bf}However, the study was funded by Narconon's parent organisation, ABLE, and subsequent correspondence in the same journal asserted that the study's conclusions were contradicted by its own data: that the control group "were more likely to {\color{blue}resist} pressures to take drugs" than the Narconon group.\end{bf}
} & \parbox[t]{7cm}{In 2008, Narconon executive Marie Cecchini published, with Richard Lennox, a paper that claimed to show that the Narconon educational program reduced drug use among youths. \begin{bf}However, the study was funded by Narconon's parent organisation, ABLE, and subsequent correspondence in the same journal asserted that the study's conclusions were contradicted by its own data: that the control group "were more likely to succumb to pressures to take drugs" than the Narconon group.\end{bf}}
 & While the affirmative edit does undo the negation, it also contradicts an earlier portion of the passage. The passage states that the studies conclusions contradict its own data, but the undone negation actually supports the studies conclusions. Annotators likely noted the reversal of the negation but failed to notice that it makes the passage as a whole inconsistent.\\
\midrule

Author \newline Valid \begin{center}
    +
\end{center} Annotator \newline InValid & \parbox[t]{7cm}{Cro-tatting combines needle tatting with crochet. The cro-tatting tool is a tatting needle with a crochet hook at the end. One can also cro-tat with a bullion crochet hook or a very straight crochet hook. In the 19th century, "crochet tatting" patterns were published which simply called for a crochet hook. One of the earliest patterns is for a crocheted afghan with tatted rings forming a raised design. Patterns are available in English and are equally divided between yarn and thread. In its most basic form, the rings are tatted with a length of plain thread between them, as in single-shuttle tatting. In modern patterns, beginning in the early 20th century, the rings are tatted and the arches or chains are crocheted. Many people consider cro-tatting more difficult than crochet or needle tatting. Some tatting instructors recommend using a tatting needle and a crochet hook to work cro-tatting patterns. \begin{bf}Stitches of cro-tatting (and needle tatting before a ring is closed) unravel easily, {\color{blue}unlike} tatting made with a shuttle.\end{bf}
} &
\parbox[t]{7cm}{Cro-tatting combines needle tatting with crochet. The cro-tatting tool is a tatting needle with a crochet hook at the end. One can also cro-tat with a bullion crochet hook or a very straight crochet hook. In the 19th century, "crochet tatting" patterns were published which simply called for a crochet hook. One of the earliest patterns is for a crocheted afghan with tatted rings forming a raised design. Patterns are available in English and are equally divided between yarn and thread. In its most basic form, the rings are tatted with a length of plain thread between them, as in single-shuttle tatting. In modern patterns, beginning in the early 20th century, the rings are tatted and the arches or chains are crocheted. Many people consider cro-tatting more difficult than crochet or needle tatting. Some tatting instructors recommend using a tatting needle and a crochet hook to work cro-tatting patterns. \begin{bf}Stitches of cro-tatting (and needle tatting before a ring is closed) do not unravel easily, similar to tatting made with a shuttle.\end{bf}}
 & An author of this paper assessed the edit as valid, as it undoes the negation of unlike and changes it to similar. Although it changes another part of the passage, the negation of unlike compared to stitches of cro-tatting is removed, and the passage is coherent and consistent. Annotators likely noticed that the passage is still saying that tatting made with a shuttle does not unravel easily as in the original passage, and took that to mean the affirmative edit was not successful.
 \\
\bottomrule 
\end{tabular}
}
\caption{Examples of LLM-Generated Affirmative Edits for CondaQA with various validity assessments. Affirmative edits must be a rewrite of the original passage such that was being negated is no longer negated, with minimal changes to the passage to keep it consistent and coherent. The negated statement is highlighted in {\bf bold} and the negation is highlighted in {\color{blue} blue}.}
\label{tab:affirmative-agreements-condaqa}
\end{table*}

\begin{thebibliography}{61}
\providecommand{\natexlab}[1]{#1}

\bibitem[{Amin et~al.(2025)Amin, Babakniya, Bie, Kong, Syed, and Vassilvitskii}]{amin2025escapingcollapsestrengthweak}
Kareem Amin, Sara Babakniya, Alex Bie, Weiwei Kong, Umar Syed, and Sergei Vassilvitskii. 2025.
\newblock \href {https://arxiv.org/abs/2502.08924} {Escaping collapse: The strength of weak data for large language model training}.
\newblock \emph{Preprint}, arXiv:2502.08924.

\bibitem[{Bai et~al.(2022)Bai, Kadavath, Kundu, Askell, Kernion, Jones, Chen, Goldie, Mirhoseini, McKinnon, Chen, Olsson, Olah, Hernandez, Drain, Ganguli, Li, Tran-Johnson, Perez, Kerr, Mueller, Ladish, Landau, Ndousse, Lukosuite, Lovitt, Sellitto, Elhage, Schiefer, Mercado, DasSarma, Lasenby, Larson, Ringer, Johnston, Kravec, Showk, Fort, Lanham, Telleen-Lawton, Conerly, Henighan, Hume, Bowman, Hatfield-Dodds, Mann, Amodei, Joseph, McCandlish, Brown, and Kaplan}]{DBLP:journals/corr/abs-2212-08073}
Yuntao Bai, Saurav Kadavath, Sandipan Kundu, Amanda Askell, Jackson Kernion, Andy Jones, Anna Chen, Anna Goldie, Azalia Mirhoseini, Cameron McKinnon, Carol Chen, Catherine Olsson, Christopher Olah, Danny Hernandez, Dawn Drain, Deep Ganguli, Dustin Li, Eli Tran-Johnson, Ethan Perez, Jamie Kerr, Jared Mueller, Jeffrey Ladish, Joshua Landau, Kamal Ndousse, Kamile Lukosuite, Liane Lovitt, Michael Sellitto, Nelson Elhage, Nicholas Schiefer, Noemi Mercado, Nova DasSarma, Robert Lasenby, Robin Larson, Sam Ringer, Scott Johnston, Shauna Kravec, Sheer~El Showk, Stanislav Fort, Tamera Lanham, Timothy Telleen-Lawton, Tom Conerly, Tom Henighan, Tristan Hume, Samuel~R. Bowman, Zac Hatfield-Dodds, Ben Mann, Dario Amodei, Nicholas Joseph, Sam McCandlish, Tom Brown, and Jared Kaplan. 2022.
\newblock \href {https://arxiv.org/abs/2212.08073} {Constitutional ai: Harmlessness from ai feedback}.
\newblock \emph{Preprint}, arXiv:2212.08073.

\bibitem[{Balepur et~al.(2025)Balepur, Gu, Ravichander, Feng, Boyd-Graber, and Rudinger}]{DBLP:journals/corr/abs-2410-15512}
Nishant Balepur, Feng Gu, Abhilasha Ravichander, Shi Feng, Jordan~Lee Boyd-Graber, and Rachel Rudinger. 2025.
\newblock \href {https://aclanthology.org/2025.naacl-short.5/} {Reverse question answering: Can an {LLM} write a question so hard (or bad) that it can`t answer?}
\newblock In \emph{Proceedings of the 2025 Conference of the Nations of the Americas Chapter of the Association for Computational Linguistics: Human Language Technologies (Volume 2: Short Papers)}, pages 44--64, Albuquerque, New Mexico. Association for Computational Linguistics.

\bibitem[{Bowman and Dahl(2021)}]{bowman-dahl-2021-will}
Samuel~R. Bowman and George Dahl. 2021.
\newblock \href {https://doi.org/10.18653/v1/2021.naacl-main.385} {What will it take to fix benchmarking in natural language understanding?}
\newblock In \emph{Proceedings of the 2021 Conference of the North American Chapter of the Association for Computational Linguistics: Human Language Technologies}, pages 4843--4855, Online. Association for Computational Linguistics.

\bibitem[{Cegin et~al.(2023)Cegin, Simko, and Brusilovsky}]{cegin-etal-2023-chatgpt}
Jan Cegin, Jakub Simko, and Peter Brusilovsky. 2023.
\newblock \href {https://doi.org/10.18653/v1/2023.emnlp-main.117} {{C}hat{GPT} to replace crowdsourcing of paraphrases for intent classification: Higher diversity and comparable model robustness}.
\newblock In \emph{Proceedings of the 2023 Conference on Empirical Methods in Natural Language Processing}, pages 1889--1905, Singapore. Association for Computational Linguistics.

\bibitem[{Chen et~al.(2023)Chen, Gao, Bosselut, Sabharwal, and Richardson}]{chen-etal-2023-disco}
Zeming Chen, Qiyue Gao, Antoine Bosselut, Ashish Sabharwal, and Kyle Richardson. 2023.
\newblock \href {https://doi.org/10.18653/v1/2023.acl-long.302} {{DISCO}: Distilling counterfactuals with large language models}.
\newblock In \emph{Proceedings of the 61st Annual Meeting of the Association for Computational Linguistics (Volume 1: Long Papers)}, pages 5514--5528, Toronto, Canada. Association for Computational Linguistics.

\bibitem[{Das et~al.(2024)Das, Langis, Martin-Boyle, Kim, Lee, Kim, Hayati, Owan, Hu, Parkar, Koo, Park, Tyagi, Ferland, Roy, Liu, and Kang}]{das2024surfacetrackingartifactualityllmgenerated}
Debarati Das, Karin~De Langis, Anna Martin-Boyle, Jaehyung Kim, Minhwa Lee, Zae~Myung Kim, Shirley~Anugrah Hayati, Risako Owan, Bin Hu, Ritik Parkar, Ryan Koo, Jonginn Park, Aahan Tyagi, Libby Ferland, Sanjali Roy, Vincent Liu, and Dongyeop Kang. 2024.
\newblock \href {https://arxiv.org/abs/2401.14698} {Under the surface: Tracking the artifactuality of llm-generated data}.
\newblock \emph{Preprint}, arXiv:2401.14698.

\bibitem[{Dixit et~al.(2022)Dixit, Paranjape, Hajishirzi, and Zettlemoyer}]{dixit-etal-2022-core}
Tanay Dixit, Bhargavi Paranjape, Hannaneh Hajishirzi, and Luke Zettlemoyer. 2022.
\newblock \href {https://aclanthology.org/2022.findings-emnlp.216} {{CORE}: A retrieve-then-edit framework for counterfactual data generation}.
\newblock In \emph{Findings of the Association for Computational Linguistics: EMNLP 2022}, pages 2964--2984, Abu Dhabi, United Arab Emirates. Association for Computational Linguistics.

\bibitem[{Dua et~al.(2019)Dua, Wang, Dasigi, Stanovsky, Singh, and Gardner}]{dua-etal-2019-drop}
Dheeru Dua, Yizhong Wang, Pradeep Dasigi, Gabriel Stanovsky, Sameer Singh, and Matt Gardner. 2019.
\newblock \href {https://doi.org/10.18653/v1/N19-1246} {{DROP}: A reading comprehension benchmark requiring discrete reasoning over paragraphs}.
\newblock In \emph{Proceedings of the 2019 Conference of the North {A}merican Chapter of the Association for Computational Linguistics: Human Language Technologies, Volume 1 (Long and Short Papers)}, pages 2368--2378, Minneapolis, Minnesota. Association for Computational Linguistics.

\bibitem[{Gardner et~al.(2020)Gardner, Artzi, Basmov, Berant, Bogin, Chen, Dasigi, Dua, Elazar, Gottumukkala, Gupta, Hajishirzi, Ilharco, Khashabi, Lin, Liu, Liu, Mulcaire, Ning, Singh, Smith, Subramanian, Tsarfaty, Wallace, Zhang, and Zhou}]{gardner-etal-2020-evaluating}
Matt Gardner, Yoav Artzi, Victoria Basmov, Jonathan Berant, Ben Bogin, Sihao Chen, Pradeep Dasigi, Dheeru Dua, Yanai Elazar, Ananth Gottumukkala, Nitish Gupta, Hannaneh Hajishirzi, Gabriel Ilharco, Daniel Khashabi, Kevin Lin, Jiangming Liu, Nelson~F. Liu, Phoebe Mulcaire, Qiang Ning, Sameer Singh, Noah~A. Smith, Sanjay Subramanian, Reut Tsarfaty, Eric Wallace, Ally Zhang, and Ben Zhou. 2020.
\newblock \href {https://doi.org/10.18653/v1/2020.findings-emnlp.117} {Evaluating models{'} local decision boundaries via contrast sets}.
\newblock In \emph{Findings of the Association for Computational Linguistics: EMNLP 2020}, pages 1307--1323, Online. Association for Computational Linguistics.

\bibitem[{Gilardi et~al.(2023)Gilardi, Alizadeh, and Kubli}]{Gilardi_2023}
Fabrizio Gilardi, Meysam Alizadeh, and Maël Kubli. 2023.
\newblock \href {https://doi.org/10.1073/pnas.2305016120} {{ChatGPT} outperforms crowd workers for text-annotation tasks}.
\newblock \emph{Proceedings of the National Academy of Sciences}, 120(30).

\bibitem[{Guo et~al.(2024)Guo, Liao, Li, and Chua}]{guo2024surveyneuralquestiongeneration}
Shasha Guo, Lizi Liao, Cuiping Li, and Tat{-}Seng Chua. 2024.
\newblock \href {https://www.ijcai.org/proceedings/2024/889} {A survey on neural question generation: Methods, applications, and prospects}.
\newblock In \emph{Proceedings of the Thirty-Third International Joint Conference on Artificial Intelligence {(IJCAI)}}, pages 8038--8047, Jeju, South Korea.

\bibitem[{Gururangan et~al.(2018)Gururangan, Swayamdipta, Levy, Schwartz, Bowman, and Smith}]{gururangan-etal-2018-annotation}
Suchin Gururangan, Swabha Swayamdipta, Omer Levy, Roy Schwartz, Samuel Bowman, and Noah~A. Smith. 2018.
\newblock \href {https://doi.org/10.18653/v1/N18-2017} {Annotation artifacts in natural language inference data}.
\newblock In \emph{Proceedings of the 2018 Conference of the North {A}merican Chapter of the Association for Computational Linguistics: Human Language Technologies, Volume 2 (Short Papers)}, pages 107--112, New Orleans, Louisiana. Association for Computational Linguistics.

\bibitem[{Harsha et~al.(2025)Harsha, Phogat, Dasaratha, Puranam, and Ramakrishna}]{harsha-etal-2025-synthetic}
Chetan Harsha, Karmvir~Singh Phogat, Sridhar Dasaratha, Sai~Akhil Puranam, and Shashishekar Ramakrishna. 2025.
\newblock \href {https://aclanthology.org/2025.finnlp-1.7/} {Synthetic data generation using large language models for financial question answering}.
\newblock In \emph{Proceedings of the Joint Workshop of the 9th Financial Technology and Natural Language Processing (FinNLP), the 6th Financial Narrative Processing (FNP), and the 1st Workshop on Large Language Models for Finance and Legal (LLMFinLegal)}, pages 76--95, Abu Dhabi, UAE. Association for Computational Linguistics.

\bibitem[{Hosking and Lapata(2021)}]{hosking-lapata-2021-factorising}
Tom Hosking and Mirella Lapata. 2021.
\newblock \href {https://doi.org/10.18653/v1/2021.acl-long.112} {Factorising meaning and form for intent-preserving paraphrasing}.
\newblock In \emph{Proceedings of the 59th Annual Meeting of the Association for Computational Linguistics and the 11th International Joint Conference on Natural Language Processing (Volume 1: Long Papers)}, pages 1405--1418, Online. Association for Computational Linguistics.

\bibitem[{Hosseini et~al.(2024)Hosseini, Petrov, Fabrikant, and Louis}]{hosseini-etal-2024-synthetic}
Mohammad~Javad Hosseini, Andrey Petrov, Alex Fabrikant, and Annie Louis. 2024.
\newblock \href {https://doi.org/10.18653/v1/2024.acl-long.120} {A synthetic data approach for domain generalization of {NLI} models}.
\newblock In \emph{Proceedings of the 62nd Annual Meeting of the Association for Computational Linguistics (Volume 1: Long Papers)}, pages 2212--2226, Bangkok, Thailand. Association for Computational Linguistics.

\bibitem[{Kaushik et~al.(2020)Kaushik, Hovy, and Lipton}]{DBLP:conf/iclr/KaushikHL20}
Divyansh Kaushik, Eduard~H. Hovy, and Zachary~Chase Lipton. 2020.
\newblock \href {https://openreview.net/forum?id=Sklgs0NFvr} {Learning the difference that makes {A} difference with counterfactually-augmented data}.
\newblock In \emph{Proceedings of the 8th International Conference on Learning Representations, ({ICLR})}.

\bibitem[{Ko{\v{c}}isk{\'y} et~al.(2018)Ko{\v{c}}isk{\'y}, Schwarz, Blunsom, Dyer, Hermann, Melis, and Grefenstette}]{kocisky-etal-2018-narrativeqa}
Tom{\'a}{\v{s}} Ko{\v{c}}isk{\'y}, Jonathan Schwarz, Phil Blunsom, Chris Dyer, Karl~Moritz Hermann, G{\'a}bor Melis, and Edward Grefenstette. 2018.
\newblock \href {https://doi.org/10.1162/tacl_a_00023} {The {N}arrative{QA} reading comprehension challenge}.
\newblock \emph{Transactions of the Association for Computational Linguistics}, 6:317--328.

\bibitem[{Krishna et~al.(2023)Krishna, Song, Karpinska, Wieting, and Iyyer}]{DBLP:conf/nips/KrishnaSKWI23}
Kalpesh Krishna, Yixiao Song, Marzena Karpinska, John Wieting, and Mohit Iyyer. 2023.
\newblock \href {http://papers.nips.cc/paper\_files/paper/2023/hash/575c450013d0e99e4b0ecf82bd1afaa4-Abstract-Conference.html} {Paraphrasing evades detectors of ai-generated text, but retrieval is an effective defense}.
\newblock In \emph{Advances in Neural Information Processing Systems 36: Annual Conference on Neural Information Processing Systems 2023, NeurIPS 2023, New Orleans, LA, USA, December 10 - 16, 2023}.

\bibitem[{Kulshreshtha and Rumshisky(2023)}]{kulshreshtha-rumshisky-2023-reasoning}
Saurabh Kulshreshtha and Anna Rumshisky. 2023.
\newblock \href {https://doi.org/10.18653/v1/2023.nlrse-1.6} {Reasoning circuits: Few-shot multi-hop question generation with structured rationales}.
\newblock In \emph{Proceedings of the 1st Workshop on Natural Language Reasoning and Structured Explanations (NLRSE)}, pages 59--77, Toronto, Canada. Association for Computational Linguistics.

\bibitem[{Li et~al.(2020)Li, Shengshuo, Liu, Wu, Zhou, and Steinert-Threlkeld}]{li-etal-2020-linguistically}
Chuanrong Li, Lin Shengshuo, Zeyu Liu, Xinyi Wu, Xuhui Zhou, and Shane Steinert-Threlkeld. 2020.
\newblock \href {https://doi.org/10.18653/v1/2020.blackboxnlp-1.12} {Linguistically-informed transformations ({LIT}): A method for automatically generating contrast sets}.
\newblock In \emph{Proceedings of the Third BlackboxNLP Workshop on Analyzing and Interpreting Neural Networks for NLP}, pages 126--135, Online. Association for Computational Linguistics.

\bibitem[{Liu et~al.(2022)Liu, Swayamdipta, Smith, and Choi}]{liu-etal-2022-wanli}
Alisa Liu, Swabha Swayamdipta, Noah~A. Smith, and Yejin Choi. 2022.
\newblock \href {https://aclanthology.org/2022.findings-emnlp.508} {{WANLI}: Worker and {AI} collaboration for natural language inference dataset creation}.
\newblock In \emph{Findings of the Association for Computational Linguistics: EMNLP 2022}, pages 6826--6847, Abu Dhabi, United Arab Emirates. Association for Computational Linguistics.

\bibitem[{Long et~al.(2024)Long, Wang, Xiao, Zhao, Ding, Chen, and Wang}]{long-etal-2024-llms}
Lin Long, Rui Wang, Ruixuan Xiao, Junbo Zhao, Xiao Ding, Gang Chen, and Haobo Wang. 2024.
\newblock \href {https://doi.org/10.18653/v1/2024.findings-acl.658} {On {LLM}s-driven synthetic data generation, curation, and evaluation: A survey}.
\newblock In \emph{Findings of the Association for Computational Linguistics: ACL 2024}, pages 11065--11082, Bangkok, Thailand. Association for Computational Linguistics.

\bibitem[{Maheshwari et~al.(2024)Maheshwari, Ivanov, and Haddad}]{maheshwari2024efficacysyntheticdatabenchmark}
Gaurav Maheshwari, Dmitry Ivanov, and Kevin~El Haddad. 2024.
\newblock \href {https://arxiv.org/abs/2409.11968} {Efficacy of synthetic data as a benchmark}.
\newblock \emph{Preprint}, arXiv:2409.11968.

\bibitem[{McKeown(1983)}]{mckeown-1983-paraphrasing}
Kathleen~R. McKeown. 1983.
\newblock \href {https://aclanthology.org/J83-1001} {Paraphrasing questions using given and new information}.
\newblock \emph{American Journal of Computational Linguistics}, 9(1):1--10.

\bibitem[{M{\o}ller et~al.(2024)M{\o}ller, Pera, Dalsgaard, and Aiello}]{moller-etal-2024-parrot}
Anders~Giovanni M{\o}ller, Arianna Pera, Jacob Dalsgaard, and Luca Aiello. 2024.
\newblock \href {https://aclanthology.org/2024.eacl-short.17} {The parrot dilemma: Human-labeled vs. {LLM}-augmented data in classification tasks}.
\newblock In \emph{Proceedings of the 18th Conference of the European Chapter of the Association for Computational Linguistics (Volume 2: Short Papers)}, pages 179--192, St. Julian{'}s, Malta. Association for Computational Linguistics.

\bibitem[{Naik et~al.(2018)Naik, Ravichander, Sadeh, Rose, and Neubig}]{naik-etal-2018-stress}
Aakanksha Naik, Abhilasha Ravichander, Norman Sadeh, Carolyn Rose, and Graham Neubig. 2018.
\newblock \href {https://aclanthology.org/C18-1198} {Stress test evaluation for natural language inference}.
\newblock In \emph{Proceedings of the 27th International Conference on Computational Linguistics}, pages 2340--2353, Santa Fe, New Mexico, USA. Association for Computational Linguistics.

\bibitem[{Nangia et~al.(2021)Nangia, Sugawara, Trivedi, Warstadt, Vania, and Bowman}]{nangia-etal-2021-ingredients}
Nikita Nangia, Saku Sugawara, Harsh Trivedi, Alex Warstadt, Clara Vania, and Samuel~R. Bowman. 2021.
\newblock \href {https://doi.org/10.18653/v1/2021.acl-long.98} {What ingredients make for an effective crowdsourcing protocol for difficult {NLU} data collection tasks?}
\newblock In \emph{Proceedings of the 59th Annual Meeting of the Association for Computational Linguistics and the 11th International Joint Conference on Natural Language Processing (Volume 1: Long Papers)}, pages 1221--1235, Online. Association for Computational Linguistics.

\bibitem[{Nguyen et~al.(2024)Nguyen, Seifert, and Schl{\"o}tterer}]{nguyen-etal-2024-ceval-benchmark}
Van~Bach Nguyen, Christin Seifert, and J{\"o}rg Schl{\"o}tterer. 2024.
\newblock \href {https://aclanthology.org/2024.inlg-main.6} {{CE}val: A benchmark for evaluating counterfactual text generation}.
\newblock In \emph{Proceedings of the 17th International Natural Language Generation Conference}, pages 55--69, Tokyo, Japan. Association for Computational Linguistics.

\bibitem[{Pan et~al.(2019)Pan, Lei, Chua, and Kan}]{DBLP:journals/corr/abs-1905-08949}
Liangming Pan, Wenqiang Lei, Tat-Seng Chua, and Min-Yen Kan. 2019.
\newblock \href {https://arxiv.org/abs/1905.08949} {Recent advances in neural question generation}.
\newblock \emph{Preprint}, arXiv:1905.08949.

\bibitem[{Patwa et~al.(2024)Patwa, Filice, Chen, Castellucci, Rokhlenko, and Malmasi}]{patwa-etal-2024-enhancing}
Parth Patwa, Simone Filice, Zhiyu Chen, Giuseppe Castellucci, Oleg Rokhlenko, and Shervin Malmasi. 2024.
\newblock \href {https://aclanthology.org/2024.lrec-main.533/} {Enhancing low-resource {LLM}s classification with {PEFT} and synthetic data}.
\newblock In \emph{Proceedings of the 2024 Joint International Conference on Computational Linguistics, Language Resources and Evaluation (LREC-COLING 2024)}, pages 6017--6023, Torino, Italia. ELRA and ICCL.

\bibitem[{Pehlivanoglu et~al.(2024)Pehlivanoglu, Gobosho, Syakura, Shanmuganathan, and de{-}la{-}Fuente{-}Valent{\'{\i}}n}]{DBLP:journals/es/PehlivanogluGSSd24}
Meltem~Kurt Pehlivanoglu, Robera~Tadesse Gobosho, Muhammad~Abdan Syakura, Vimal Shanmuganathan, and Luis de{-}la{-}Fuente{-}Valent{\'{\i}}n. 2024.
\newblock \href {https://doi.org/10.1111/EXSY.13699} {Comparative analysis of paraphrasing performance of chatgpt, gpt-3, and {T5} language models using a new chatgpt generated dataset: Paragpt}.
\newblock \emph{Expert Syst. J. Knowl. Eng.}, 41(11).

\bibitem[{Qi et~al.(2025)Qi, Ma, Li, Du, Hui, Wu, Laili, and He}]{qi2025large}
Chengwen Qi, Ren Ma, Bowen Li, He~Du, Binyuan Hui, Jinwang Wu, Yuanjun Laili, and Conghui He. 2025.
\newblock \href {https://openreview.net/forum?id=C25SgeXWjE} {Large language models meet symbolic provers for logical reasoning evaluation}.
\newblock In \emph{Proceedings of The Thirteenth International Conference on Learning Representations (ICLR)}.

\bibitem[{Rajpurkar et~al.(2016)Rajpurkar, Zhang, Lopyrev, and Liang}]{rajpurkar-etal-2016-squad}
Pranav Rajpurkar, Jian Zhang, Konstantin Lopyrev, and Percy Liang. 2016.
\newblock \href {https://doi.org/10.18653/v1/D16-1264} {{SQ}u{AD}: 100,000+ questions for machine comprehension of text}.
\newblock In \emph{Proceedings of the 2016 Conference on Empirical Methods in Natural Language Processing}, pages 2383--2392, Austin, Texas. Association for Computational Linguistics.

\bibitem[{Ravichander et~al.(2022)Ravichander, Gardner, and Marasovic}]{ravichander-etal-2022-condaqa}
Abhilasha Ravichander, Matt Gardner, and Ana Marasovic. 2022.
\newblock \href {https://aclanthology.org/2022.emnlp-main.598} {{CONDAQA}: A contrastive reading comprehension dataset for reasoning about negation}.
\newblock In \emph{Proceedings of the 2022 Conference on Empirical Methods in Natural Language Processing}, pages 8729--8755, Abu Dhabi, United Arab Emirates. Association for Computational Linguistics.

\bibitem[{Ribeiro et~al.(2020)Ribeiro, Wu, Guestrin, and Singh}]{ribeiro-etal-2020-beyond}
Marco~Tulio Ribeiro, Tongshuang Wu, Carlos Guestrin, and Sameer Singh. 2020.
\newblock \href {https://doi.org/10.18653/v1/2020.acl-main.442} {Beyond accuracy: Behavioral testing of {NLP} models with {C}heck{L}ist}.
\newblock In \emph{Proceedings of the 58th Annual Meeting of the Association for Computational Linguistics}, pages 4902--4912, Online. Association for Computational Linguistics.

\bibitem[{Roemmele and Gordon(2024)}]{roemmele-gordon-2024-test}
Melissa Roemmele and Andrew Gordon. 2024.
\newblock \href {https://doi.org/10.18653/v1/2024.findings-emnlp.299} {From test-taking to test-making: Examining {LLM} authoring of commonsense assessment items}.
\newblock In \emph{Findings of the Association for Computational Linguistics: EMNLP 2024}, pages 5193--5203, Miami, Florida, USA. Association for Computational Linguistics.

\bibitem[{Rogers et~al.(2023)Rogers, Gardner, and Augenstein}]{DBLP:journals/csur/RogersGA23}
Anna Rogers, Matt Gardner, and Isabelle Augenstein. 2023.
\newblock \href {https://doi.org/10.1145/3560260} {{QA} dataset explosion: {A} taxonomy of {NLP} resources for question answering and reading comprehension}.
\newblock \emph{{ACM} Comput. Surv.}, 55(10):197:1--197:45.

\bibitem[{Ross et~al.(2021)Ross, Marasovi{\'c}, and Peters}]{ross-etal-2021-explaining}
Alexis Ross, Ana Marasovi{\'c}, and Matthew Peters. 2021.
\newblock \href {https://doi.org/10.18653/v1/2021.findings-acl.336} {Explaining {NLP} models via minimal contrastive editing ({M}i{CE})}.
\newblock In \emph{Findings of the Association for Computational Linguistics: ACL-IJCNLP 2021}, pages 3840--3852, Online. Association for Computational Linguistics.

\bibitem[{Ross et~al.(2022)Ross, Wu, Peng, Peters, and Gardner}]{ross-etal-2022-tailor}
Alexis Ross, Tongshuang Wu, Hao Peng, Matthew Peters, and Matt Gardner. 2022.
\newblock \href {https://doi.org/10.18653/v1/2022.acl-long.228} {Tailor: Generating and perturbing text with semantic controls}.
\newblock In \emph{Proceedings of the 60th Annual Meeting of the Association for Computational Linguistics (Volume 1: Long Papers)}, pages 3194--3213, Dublin, Ireland. Association for Computational Linguistics.

\bibitem[{Ruggeri et~al.(2024)Ruggeri, Misino, Muti, Korre, Torroni, and Barrón-Cedeño}]{ruggeri2024letguidelinesguideyou}
Federico Ruggeri, Eleonora Misino, Arianna Muti, Katerina Korre, Paolo Torroni, and Alberto Barrón-Cedeño. 2024.
\newblock \href {https://arxiv.org/abs/2406.14099} {Let guidelines guide you: A prescriptive guideline-centered data annotation methodology}.
\newblock \emph{Preprint}, arXiv:2406.14099.

\bibitem[{Rus et~al.(2010)Rus, Wyse, Piwek, Lintean, Stoyanchev, and Moldovan}]{rus-etal-2010-first}
Vasile Rus, Brendan Wyse, Paul Piwek, Mihai Lintean, Svetlana Stoyanchev, and Christian Moldovan. 2010.
\newblock \href {https://aclanthology.org/W10-4234} {The first question generation shared task evaluation challenge}.
\newblock In \emph{Proceedings of the 6th International Natural Language Generation Conference}. Association for Computational Linguistics.

\bibitem[{Samuel et~al.(2024)Samuel, Aynaou, Chowdhury, Venkat~Ramanan, and Chadha}]{samuel-etal-2024-llms}
Vinay Samuel, Houda Aynaou, Arijit Chowdhury, Karthik Venkat~Ramanan, and Aman Chadha. 2024.
\newblock \href {https://doi.org/10.18653/v1/2024.acl-srw.36} {Can {LLM}s augment low-resource reading comprehension datasets? opportunities and challenges}.
\newblock In \emph{Proceedings of the 62nd Annual Meeting of the Association for Computational Linguistics (Volume 4: Student Research Workshop)}, pages 307--317, Bangkok, Thailand. Association for Computational Linguistics.

\bibitem[{Seo et~al.(2017)Seo, Kembhavi, Farhadi, and Hajishirzi}]{seo2016bidirectional}
Min~Joon Seo, Aniruddha Kembhavi, Ali Farhadi, and Hannaneh Hajishirzi. 2017.
\newblock \href {https://openreview.net/forum?id=HJ0UKP9ge} {Bidirectional attention flow for machine comprehension}.
\newblock In \emph{Proceedings of the 5th International Conference on Learning Representations, ({ICLR})}.

\bibitem[{Shaikh et~al.(2022)Shaikh, Ferreira, and Stent}]{-2022-international}
Samira Shaikh, Thiago Ferreira, and Amanda Stent, editors. 2022.
\newblock \href {https://aclanthology.org/2022.inlg-genchal.0} {\emph{Proceedings of the 15th International Conference on Natural Language Generation: Generation Challenges}}. Association for Computational Linguistics, Waterville, Maine, USA and virtual meeting.

\bibitem[{Stepin et~al.(2021)Stepin, Alonso, Catal{\'{a}}, and Pereira{-}Fari{\~{n}}a}]{DBLP:journals/access/StepinACP21}
Ilia Stepin, Jos{\'{e}}~Maria Alonso, Alejandro Catal{\'{a}}, and Martin Pereira{-}Fari{\~{n}}a. 2021.
\newblock \href {https://doi.org/10.1109/ACCESS.2021.3051315} {A survey of contrastive and counterfactual explanation generation methods for explainable artificial intelligence}.
\newblock \emph{{IEEE} Access}, 9:11974--12001.

\bibitem[{Team(2024)}]{qwen2.5}
Qwen Team. 2024.
\newblock \href {https://qwenlm.github.io/blog/qwen2.5/} {Qwen2.5: A party of foundation models}.

\bibitem[{Tonga et~al.(2025)Tonga, Clement, and Oudeyer}]{pmlr-v264-tonga25a}
Junior~Cedric Tonga, Benjamin Clement, and Pierre-Yves Oudeyer. 2025.
\newblock \href {https://proceedings.mlr.press/v264/tonga25a.html} {Automatic generation of question hints for mathematics problems using large language models in educational technology}.
\newblock In \emph{Proceedings of Large Foundation Models for Educational Assessment}, volume 264 of \emph{Proceedings of Machine Learning Research}, pages 61--102. PMLR.

\bibitem[{Ushio et~al.(2022)Ushio, Alva-Manchego, and Camacho-Collados}]{ushio-etal-2022-generative}
Asahi Ushio, Fernando Alva-Manchego, and Jose Camacho-Collados. 2022.
\newblock \href {https://aclanthology.org/2022.emnlp-main.42} {Generative language models for paragraph-level question generation}.
\newblock In \emph{Proceedings of the 2022 Conference on Empirical Methods in Natural Language Processing}, pages 670--688, Abu Dhabi, United Arab Emirates. Association for Computational Linguistics.

\bibitem[{Wang et~al.(2023)Wang, Kordi, Mishra, Liu, Smith, Khashabi, and Hajishirzi}]{wang-etal-2023-self-instruct}
Yizhong Wang, Yeganeh Kordi, Swaroop Mishra, Alisa Liu, Noah~A. Smith, Daniel Khashabi, and Hannaneh Hajishirzi. 2023.
\newblock \href {https://doi.org/10.18653/v1/2023.acl-long.754} {Self-instruct: Aligning language models with self-generated instructions}.
\newblock In \emph{Proceedings of the 61st Annual Meeting of the Association for Computational Linguistics (Volume 1: Long Papers)}, pages 13484--13508, Toronto, Canada. Association for Computational Linguistics.

\bibitem[{Wang et~al.(2024)Wang, Qiu, Yue, Guo, Zeng, Feng, and Shen}]{wang2024surveynaturallanguagecounterfactual}
Yongjie Wang, Xiaoqi Qiu, Yu~Yue, Xu~Guo, Zhiwei Zeng, Yuhong Feng, and Zhiqi Shen. 2024.
\newblock \href {https://doi.org/10.18653/v1/2024.findings-emnlp.276} {A survey on natural language counterfactual generation}.
\newblock In \emph{Findings of the Association for Computational Linguistics: EMNLP 2024}, pages 4798--4818, Miami, Florida, USA. Association for Computational Linguistics.

\bibitem[{Weissenborn et~al.(2017)Weissenborn, Wiese, and Seiffe}]{weissenborn-etal-2017-making}
Dirk Weissenborn, Georg Wiese, and Laura Seiffe. 2017.
\newblock \href {https://doi.org/10.18653/v1/K17-1028} {Making neural {QA} as simple as possible but not simpler}.
\newblock In \emph{Proceedings of the 21st Conference on Computational Natural Language Learning ({C}o{NLL} 2017)}, pages 271--280, Vancouver, Canada. Association for Computational Linguistics.

\bibitem[{West et~al.(2024)West, Lu, Dziri, Brahman, Li, Hwang, Jiang, Fisher, Ravichander, Chandu, Newman, Koh, Ettinger, and Choi}]{west2024the}
Peter West, Ximing Lu, Nouha Dziri, Faeze Brahman, Linjie Li, Jena~D. Hwang, Liwei Jiang, Jillian Fisher, Abhilasha Ravichander, Khyathi Chandu, Benjamin Newman, Pang~Wei Koh, Allyson Ettinger, and Yejin Choi. 2024.
\newblock \href {https://openreview.net/forum?id=CF8H8MS5P8} {The generative {AI} paradox: {\textquotedblleft}what it can create, it may not understand{\textquotedblright}}.
\newblock In \emph{Proceedings of The Twelfth International Conference on Learning Representations (ICLR)}.

\bibitem[{Witteveen and Andrews(2019)}]{witteveen-andrews-2019-paraphrasing}
Sam Witteveen and Martin Andrews. 2019.
\newblock \href {https://doi.org/10.18653/v1/D19-5623} {Paraphrasing with large language models}.
\newblock In \emph{Proceedings of the 3rd Workshop on Neural Generation and Translation}, pages 215--220, Hong Kong. Association for Computational Linguistics.

\bibitem[{Wu et~al.(2021)Wu, Wang, Li, Zhang, Xiao, Wu, Zhang, and Wang}]{wu-etal-2021-data}
Kun Wu, Lijie Wang, Zhenghua Li, Ao~Zhang, Xinyan Xiao, Hua Wu, Min Zhang, and Haifeng Wang. 2021.
\newblock \href {https://doi.org/10.18653/v1/2021.emnlp-main.707} {Data augmentation with hierarchical {SQL}-to-question generation for cross-domain text-to-{SQL} parsing}.
\newblock In \emph{Proceedings of the 2021 Conference on Empirical Methods in Natural Language Processing}, pages 8974--8983, Online and Punta Cana, Dominican Republic. Association for Computational Linguistics.

\bibitem[{Wu et~al.(2025)Wu, Zhu, Albayrak, Axon, Bertsch, Deng, Ding, Guo, Gururaja, Kuo, Liang, Liu, Mandal, Milbauer, Ni, Padmanabhan, Ramkumar, Sudjianto, Taylor, Tseng, Vaidos, Wu, Wu, and Yang}]{wu2023llmsworkershumancomputationalalgorithms}
Tongshuang Wu, Haiyi Zhu, Maya Albayrak, Alexis Axon, Amanda Bertsch, Wenxing Deng, Ziqi Ding, Boyuan Guo, Sireesh Gururaja, Tzu{-}Sheng Kuo, Jenny~T. Liang, Ryan Liu, Ihita Mandal, Jeremiah Milbauer, Xiaolin Ni, Namrata Padmanabhan, Subhashini Ramkumar, Alexis Sudjianto, Jordan Taylor, Ying{-}Jui Tseng, Patricia Vaidos, Zhijin Wu, Wei Wu, and Chenyang Yang. 2025.
\newblock \href {https://doi.org/10.1145/3706599.3706690} {Llms as workers in human-computational algorithms? replicating crowdsourcing pipelines with llms}.
\newblock In \emph{Proceedings of the Extended Abstracts of the {CHI} Conference on Human Factors in Computing Systems, {CHI} {EA} 2025, Yokohama, Japan, 26 April 2025- 1 May 2025}, pages 684:1--684:10. {ACM}.

\bibitem[{Yang et~al.(2018)Yang, Qi, Zhang, Bengio, Cohen, Salakhutdinov, and Manning}]{yang-etal-2018-hotpotqa}
Zhilin Yang, Peng Qi, Saizheng Zhang, Yoshua Bengio, William Cohen, Ruslan Salakhutdinov, and Christopher~D. Manning. 2018.
\newblock \href {https://doi.org/10.18653/v1/D18-1259} {{H}otpot{QA}: A dataset for diverse, explainable multi-hop question answering}.
\newblock In \emph{Proceedings of the 2018 Conference on Empirical Methods in Natural Language Processing}, pages 2369--2380, Brussels, Belgium. Association for Computational Linguistics.

\bibitem[{Ye et~al.(2022)Ye, Gao, Li, Xu, Feng, Wu, Yu, and Kong}]{ye-etal-2022-zerogen}
Jiacheng Ye, Jiahui Gao, Qintong Li, Hang Xu, Jiangtao Feng, Zhiyong Wu, Tao Yu, and Lingpeng Kong. 2022.
\newblock \href {https://aclanthology.org/2022.emnlp-main.801} {{Z}ero{G}en: Efficient zero-shot learning via dataset generation}.
\newblock In \emph{Proceedings of the 2022 Conference on Empirical Methods in Natural Language Processing}, pages 11653--11669, Abu Dhabi, United Arab Emirates. Association for Computational Linguistics.

\bibitem[{Yehudai et~al.(2024)Yehudai, Carmeli, Mass, Arviv, Mills, Toledo, Shnarch, and Choshen}]{yehudai-et-al-2024-genie}
Asaf Yehudai, Boaz Carmeli, Yosi Mass, Ofir Arviv, Nathaniel Mills, Assaf Toledo, Eyal Shnarch, and Leshem Choshen. 2024.
\newblock \href {https://arxiv.org/abs/2401.14367} {Genie: Achieving human parity in content-grounded datasets generation}.
\newblock In \emph{Proceedings of The Twelfth International Conference on Learning Representations (ICLR)}.

\bibitem[{Yu et~al.(2024)Yu, Jiang, Shi, Yu, Liu, Zhang, Kwok, Li, Weller, and Liu}]{DBLP:conf/iclr/YuJSYLZKLWL24}
Longhui Yu, Weisen Jiang, Han Shi, Jincheng Yu, Zhengying Liu, Yu~Zhang, James~T. Kwok, Zhenguo Li, Adrian Weller, and Weiyang Liu. 2024.
\newblock \href {https://openreview.net/forum?id=N8N0hgNDRt} {Metamath: Bootstrap your own mathematical questions for large language models}.
\newblock In \emph{Proceedings of The Twelfth International Conference on Learning Representations ({ICLR})}.

\bibitem[{Yu et~al.(2020)Yu, Jiang, Dong, and Feng}]{DBLP:conf/iclr/YuJDF20}
Weihao Yu, Zihang Jiang, Yanfei Dong, and Jiashi Feng. 2020.
\newblock \href {https://openreview.net/forum?id=HJgJtT4tvB} {{ReClor:} {A} reading comprehension dataset requiring logical reasoning}.
\newblock In \emph{Proceedings of the 8th International Conference on Learning Representations ({ICLR})}.

\end{thebibliography}
\end{document}